\documentclass{article}
\pdfoutput=1

\usepackage{arxiv}

\usepackage[utf8]{inputenc} 
\usepackage[T1]{fontenc}    
\usepackage{hyperref}       
\usepackage{url}            
\usepackage{booktabs}       
\usepackage{amsfonts}       
\usepackage{amsmath}
\usepackage{amssymb}
\usepackage{algpseudocode}  
\usepackage{algorithm}      
\usepackage{siunitx}
\usepackage{placeins}
\usepackage[inline]{enumitem}
\usepackage{graphicx}
\usepackage{nicefrac}       
\usepackage{microtype}      
\usepackage{multirow}
\usepackage{xcolor}
\usepackage{subcaption}
\usepackage{footnote}
\usepackage{scrextend}
\usepackage[page,header]{appendix}
\usepackage{titletoc}
\usepackage{blindtext}

\newcommand{\R}{\mathbb{R}}

\renewcommand{\vec}[1]{\boldsymbol{#1}}

\newcommand{\citep}{\cite}

\makesavenoteenv{tabular}
\makesavenoteenv{table}

\newcommand*\samethanks[1][\value{footnote}]{\footnotemark[#1]}

\MakeRobust{\Call}

\title{Solving Rubik’s Cube with a Robot Hand}

\author{
  OpenAI \\
  Ilge Akkaya\thanks{Authors are listed alphabetically. We include a detailed contribution statement at the end of this manuscript. Please cite as OpenAI et al., and use the following bibtex for citation: \url{https://openai.com/bibtex/openai2019rubiks.bib}},
  Marcin Andrychowicz\samethanks,
  Maciek Chociej\samethanks,
  Mateusz Litwin\samethanks,
  Bob McGrew\samethanks,
  Arthur Petron\samethanks, \\
  Alex Paino\samethanks,
  Matthias Plappert\samethanks,
  Glenn Powell\samethanks,
  Raphael Ribas\samethanks,
  Jonas Schneider\samethanks,
  Nikolas Tezak\samethanks,\\
  Jerry Tworek\samethanks,
  Peter Welinder\samethanks,
  Lilian Weng\samethanks,
  Qiming Yuan\samethanks,
  Wojciech Zaremba\samethanks,
  Lei Zhang\samethanks
}

\begin{document}

\maketitle

\begin{figure}[h!]
\centering
\includegraphics[width=1.0\textwidth]{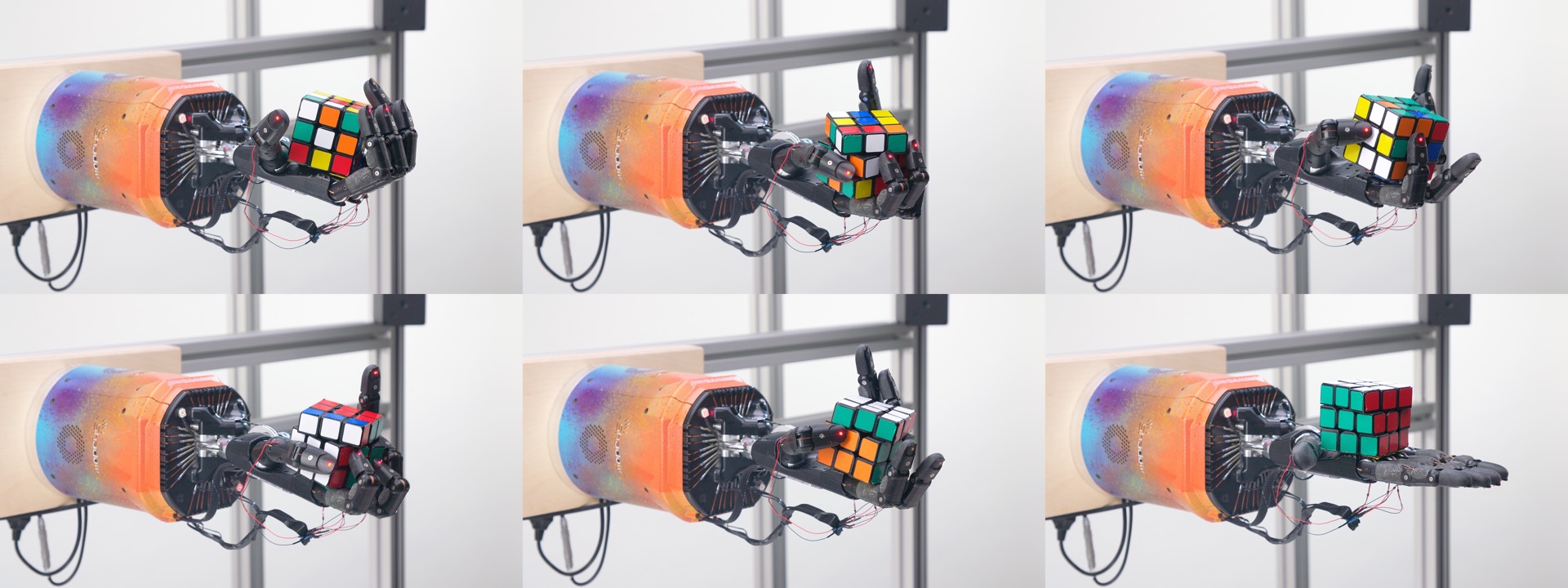} \\ \vspace{0.1cm}
\caption{A five-fingered humanoid hand trained with reinforcement learning and automatic domain randomization solving a Rubik's cube.}
\end{figure}

\begin{abstract}
We demonstrate that models trained only in simulation can be used to solve a manipulation problem of unprecedented complexity on a real robot. This is made possible by two key components: a novel algorithm, which we call automatic domain randomization (ADR) and a robot platform built for machine learning. ADR automatically generates a distribution over randomized environments of ever-increasing difficulty. Control policies and vision state estimators trained with ADR exhibit vastly improved sim2real transfer. For control policies, memory-augmented models trained on an ADR-generated distribution of environments show clear signs of emergent meta-learning at test time. The combination of ADR with our custom robot platform allows us to solve a Rubik's cube with a humanoid robot hand, which involves both control and state estimation problems. Videos summarizing our results are available: \url{https://openai.com/blog/solving-rubiks-cube/}
\end{abstract}

\section{Introduction}
\label{sec:intro}
\begin{figure}
    \centering
    \includegraphics[width=0.95\textwidth]{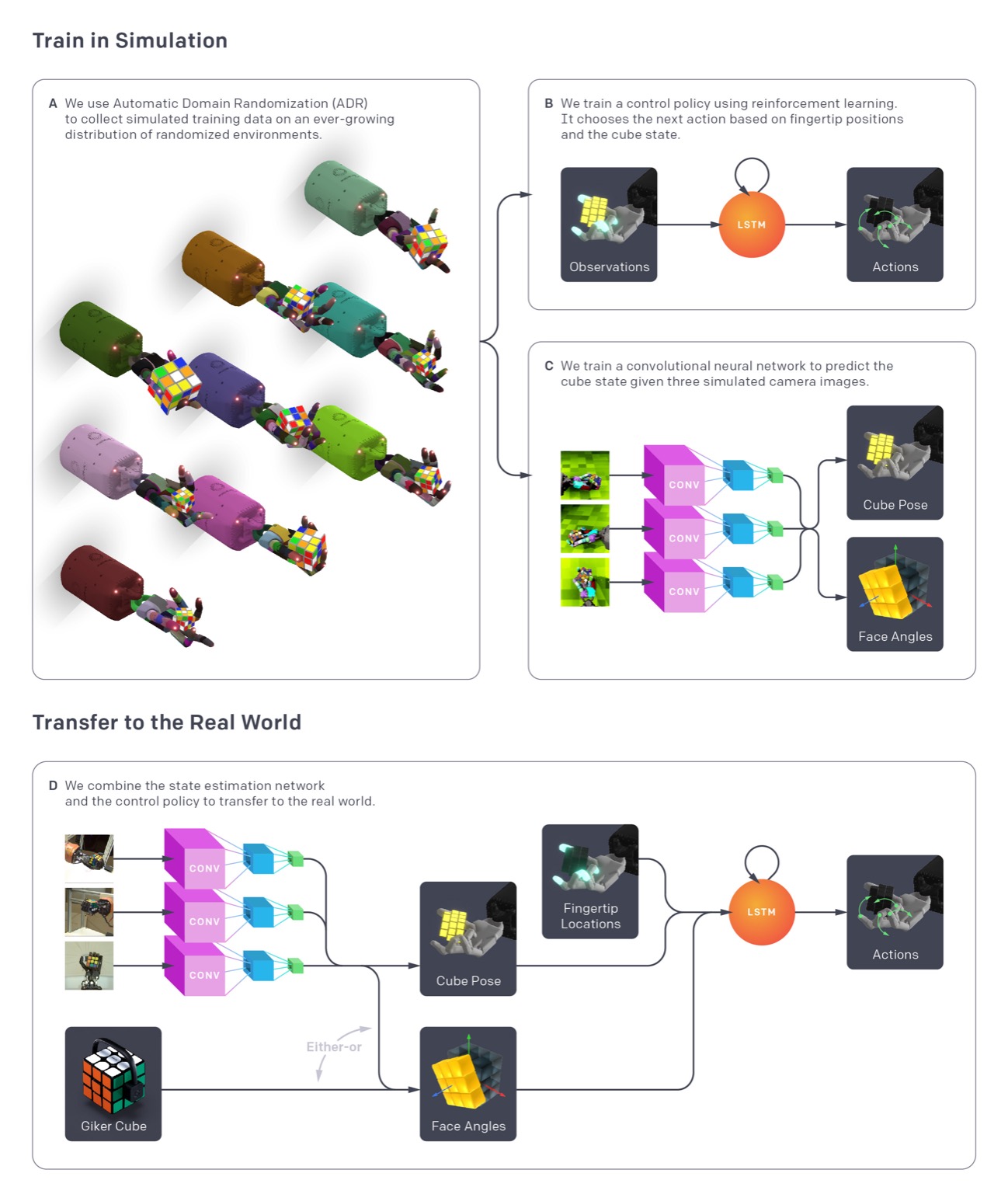}
    \caption{
        System Overview. (a) We use automatic domain randomization (ADR) to generate a growing distribution of simulations with randomized parameters and appearances. We use this data for both the control policy and vision-based state estimator. (b) The control policy receives observed robot states and rewards from the randomized simulations and learns to solve them using a recurrent neural network and reinforcement learning. (c) The vision-based state estimator uses rendered scenes collected from the randomized simulations and learns to predict the pose as well as face angles of the Rubik's cube using a convolutional neural network (CNN), trained separately from the control policy. (d) To transfer to the real world, we predict the Rubik's cube's pose from 3 real camera feeds with the CNN and measure the robot fingertip locations using a 3D motion capture system. The face angles that describe the internal rotational state of the Rubik's cube are provided by either the same vision state estimator \emph{or} the Giiker cube,  a custom cube with embedded sensors and feed it into the policy network.
    }
    \label{fig:overview}
\end{figure}

Building robots that are as versatile as humans remains a grand challenge of robotics. While humanoid robotics systems exist~\cite{bostondynamicsatlas,shadow-robot,toyotathr3,sakagami2002intelligent,asfour2013armar}, using them in the real world for complex tasks remains a daunting challenge. Machine learning has the potential to change this by \emph{learning} how to use sensor information to control the robot system appropriately instead of hand-programming the robot using expert knowledge.

However, learning requires vast amount of training data, which is hard and expensive to acquire on a physical system.  Collecting all data in simulation is therefore appealing. However, the simulation does not capture the environment or the robot accurately in every detail and therefore we also need to solve the resulting sim2real transfer problem.
Domain randomization techniques~\cite{tobin2017domain, peng2017sim} have shown great potential and have demonstrated that models trained only in simulation can transfer to the real robot system.

In prior work, we have demonstrated that we can perform complex in-hand manipulation of a block~\cite{openai2018learning}. This time, we aim to solve the manipulation and state estimation problems required to solve a Rubik's cube with the Shadow Dexterous Hand~\cite{shadow-robot} using only simulated data. This problem is much more difficult since it requires significantly more dexterity and precision for manipulating the Rubik's cube. The state estimation problem is also much harder as we need to know with high accuracy what the pose and internal state of the Rubik's cube are. We achieve this by introducing a novel method for automatically generating a distribution over randomized environments for training reinforcement learning policies and vision state estimators. We call this algorithm \emph{automatic domain randomization} (ADR). We also built a robot platform for solving a Rubik's cube in the real world in a way that complements our machine learning approach. \autoref{fig:overview} shows an overview of our system.

We investigate why policies trained with ADR transfer so well from simulation to the real robot. We find clear signs of emergent learning that happens at \emph{test time} within the recurrent internal state of our policy. We believe that this is a direct result of us training on an ever-growing distribution over randomized environments with a memory-augmented policy. In other words, training an LSTM over an ADR distribution is implicit meta-learning. We also systematically study and quantify this observation in our work.

The remainder of this manuscript is structured as follows. \autoref{sec:task} introduces two manipulation tasks we consider here. \autoref{sec:physical-setup} describes our physical setup and \autoref{sec:sim} describes how our setup is modeled in simulation. We introduce a new algorithm called automatic domain randomization (ADR), in \autoref{sec:adr}. In \autoref{sec:policy} and \autoref{sec:vision} we describe how we train control policies and vision state estimators, respectively. We present our key quantitative and qualitative results on the two tasks in \autoref{sec:exp-adr}. In \autoref{sec:exp-meta} we systematically analyze our policy for signs of emergent meta-learning. \autoref{sec:related} reviews related work and we conclude with \autoref{sec:conclusion}.

If you are mostly interested in the machine learning aspects of this manuscript, \autoref{sec:adr},  \autoref{sec:policy},  \autoref{sec:vision}, \autoref{sec:exp-adr}, and \autoref{sec:exp-meta} are especially relevant. If you are interested in the robotics aspects, \autoref{sec:physical-setup}, \autoref{sec:sim}, and \autoref{sec:result-rubik} are especially relevant.

\section{Tasks}
\label{sec:task}
\begin{figure}[h]
    \centering
    \begin{subfigure}[b]{0.48\textwidth}
        \includegraphics[width=\textwidth]{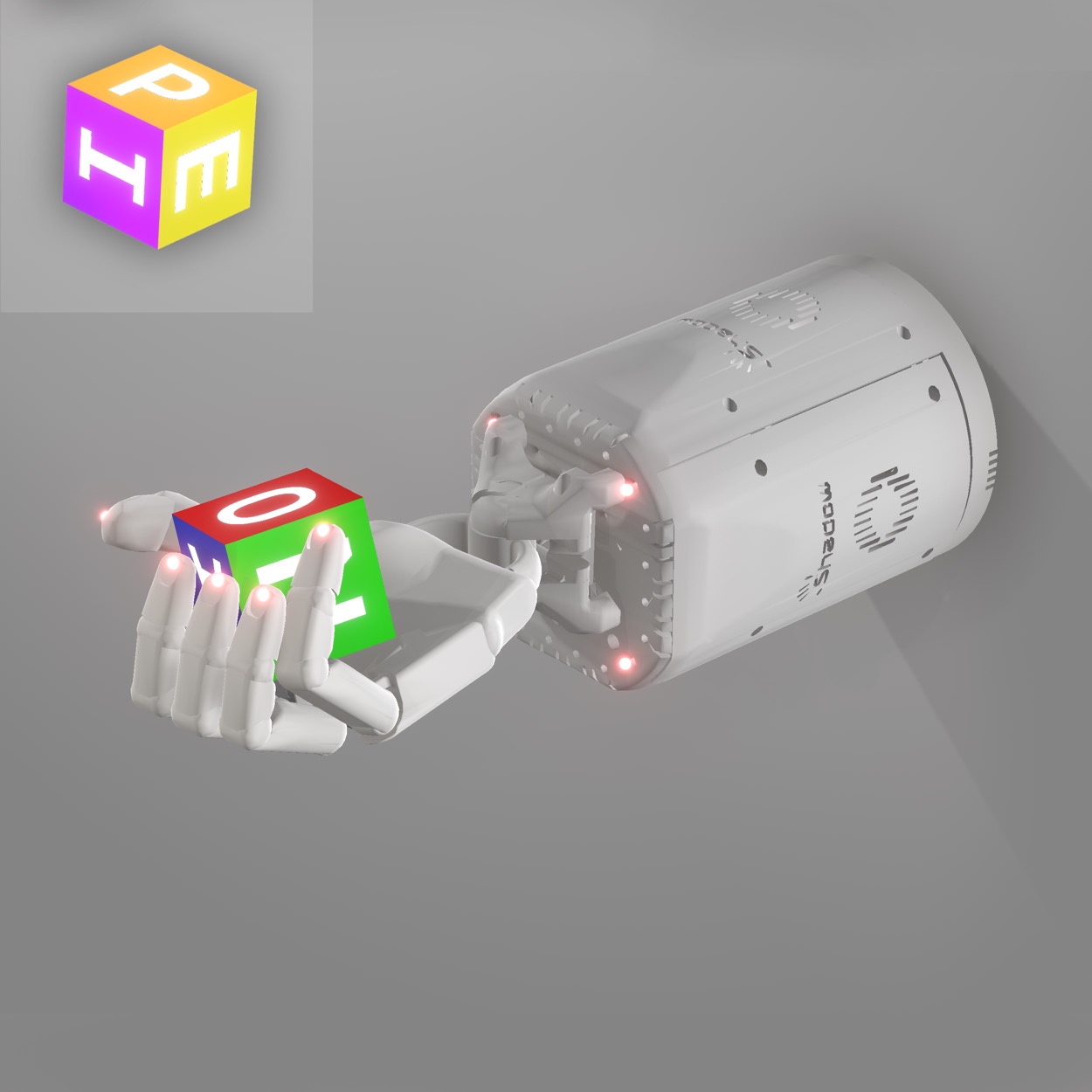}
        \caption{Block reorientation}
        \label{fig:task-block}
    \end{subfigure}
    \hfill
    \begin{subfigure}[b]{0.48\textwidth}
        \includegraphics[width=\textwidth]{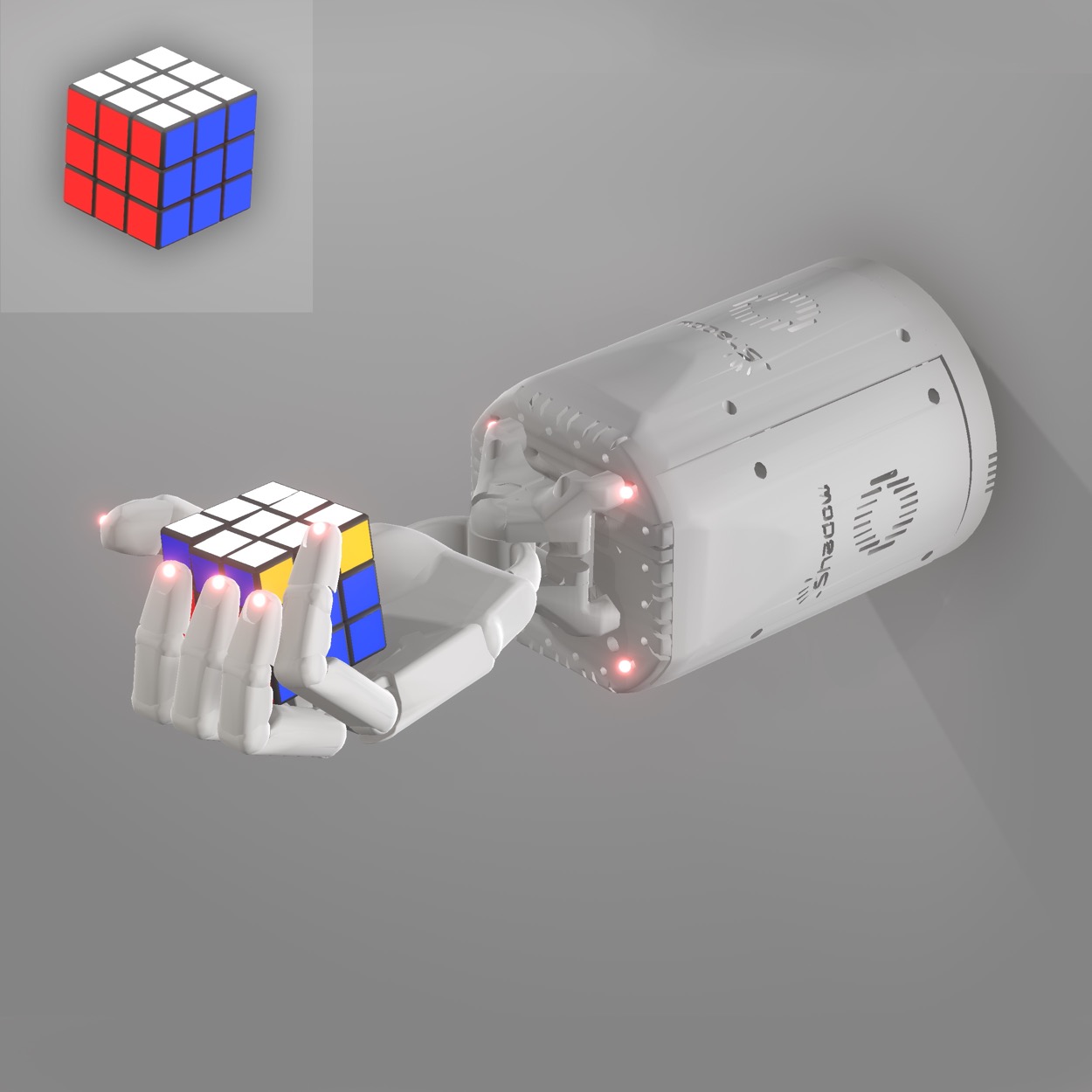}
        \caption{Rubik's cube}
        \label{fig:task-cube}
    \end{subfigure}
    \caption{Visualization of the block reorientation task (left) and the Rubik's cube task (right). In both cases, we use a single Shadow Dexterous Hand to solve the task. We also depict the goal that the policy is asked to achieve in the upper left corner.}
    \label{fig:task}
\end{figure}

In this work, we consider two different tasks that both use the Shadow Dexterous Hand~\cite{shadow-robot}: the block reorientation task from our previous work~\cite{openai2018learning, plappert2018multi} and the task of solving a Rubik's cube. Both tasks are visualized in \autoref{fig:task}. We briefly describe the details of each task in this section.

\subsection{Block Reorientation}
The block reorientation task was previously proposed in~\cite{plappert2018multi} and solved on a physical robot hand in~\cite{openai2018learning}. We briefly review it here; please refer to the aforementioned citations for additional details.

The goal of the block reorientation task is to rotate a block into a desired goal orientation. For example, in \autoref{fig:task-block}, the desired orientation is shown next to the hand with the red face facing up, the blue face facing to the left and the green face facing forward. A goal is considered achieved if the block's rotation matches the goal rotation within $0.4$~radians. After a goal is achieved, a new random goal is generated.

\subsection{Rubik's Cube}
We introduce a new and significantly more difficult problem in this work: solving a Rubik's cube\footnote{\url{https://en.wikipedia.org/wiki/Rubik's_Cube}} with the same Shadow Dexterous Hand. In brief, a Rubik's cube is a puzzle with 6 internal degrees of freedom. It consists of $26$~\emph{cubelets} that are connected via a system of joints and springs. Each of the 6 \emph{faces} of the cube can be rotated, allowing the Rubik's cube to be \emph{scrambled}. A Rubik's cube is considered solved if all 6 faces have been returned to a single color each. \autoref{fig:task-cube} depicts a Rubik's cube that is a single $90$~degree rotation of the top face away from being solved.

We consider two types of \emph{subgoals}: A \emph{rotation} corresponds to rotating a single face of the Rubik's cube by $90$~degrees in the clockwise or counter-clockwise direction. A \emph{flip} corresponds to moving a different face of the Rubik's cube to the top. We found rotating the top face to be far simpler than rotating other faces. Thus, instead of rotating arbitrary faces, we combine together a flip and a top face rotation in order to perform the desired operation. These subgoals can then be performed sequentially to eventually solve the Rubik's cube.

The difficulty of solving a Rubik's cube obviously depends on how much it has been scrambled before. We use the official scrambling method used by the World Cube Association\footnote{\url{https://www.worldcubeassociation.org/regulations/scrambles/}} to obtain what they refer to as a \emph{fair scramble}. A fair scramble typically consists of around $20$ moves that are applied to a solved Rubik's cube to scramble it.

When it comes to solving the Rubik's cube, computing a solution sequence can easily be done with existing software libraries like the Kociemba solver~\cite{kociemba}. We use this solver to produce a solution sequence of subgoals for the hand to perform. In this work, the key problem is thus about sensing and control, \emph{not} finding the solution sequence. More concretely, we need to obtain the state of the Rubik's cube (i.e. its pose as well as its 6 face angles) and use that information to control the robot hand such that each subgoal is successfully achieved.

\section{Physical Setup}
\label{sec:physical-setup}
Having described the task, we next describe the physical setup that we use to solve the block and the Rubik's cube in the real world. We focus on the differences that made it possible to solve the Rubik's cube since~\cite{openai2018learning} has already described our physical setup for solving the block reorientation task.

\subsection{Robot Platform}

\begin{figure}[h]
    \centering
    \begin{subfigure}[b]{0.48\textwidth}
        \includegraphics[width=\textwidth]{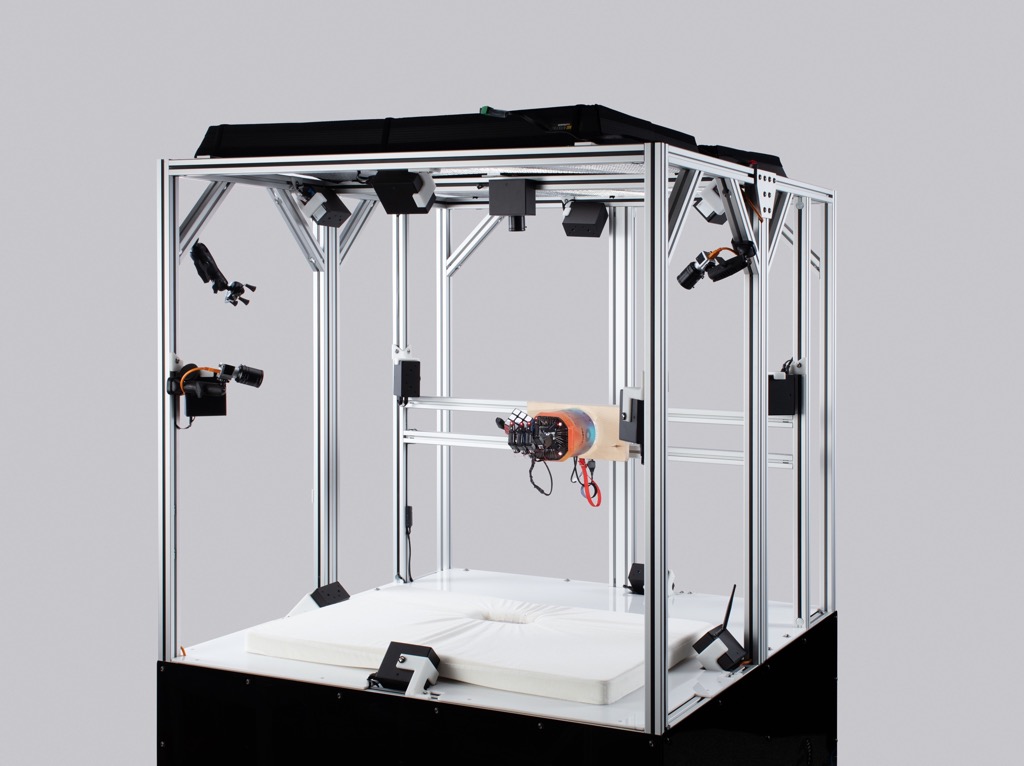}
        \caption{The cage.}
        \label{fig:setup-cage}
    \end{subfigure}
    \hfill
    \begin{subfigure}[b]{0.48\textwidth}
        \includegraphics[width=\textwidth]{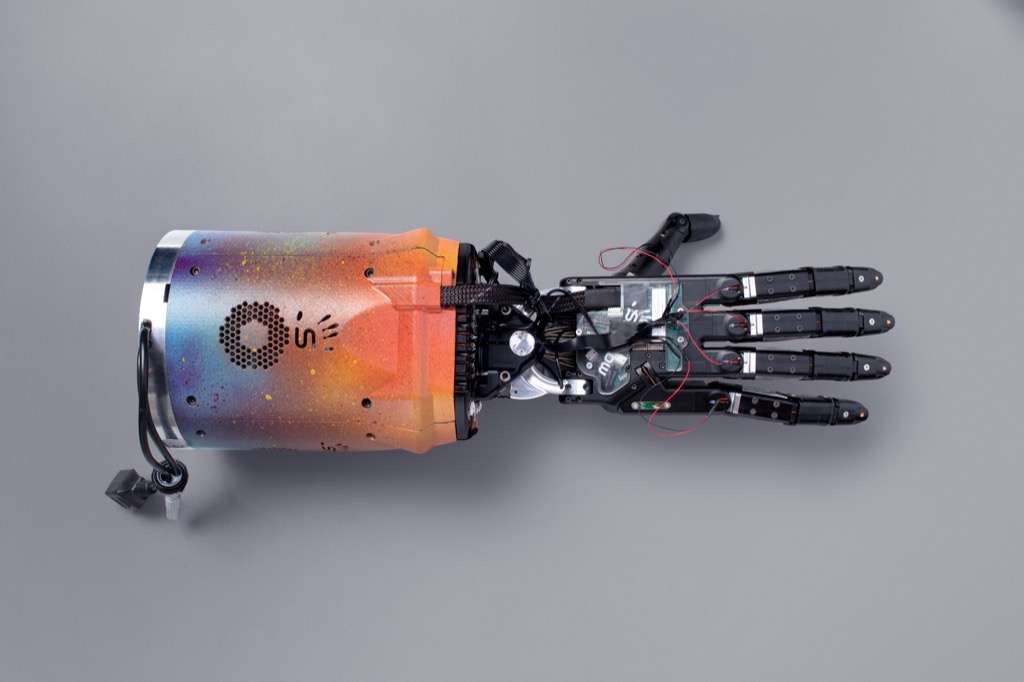}
        \caption{The Shadow Dexterous Hand.}
        \label{fig:setup-hand}
    \end{subfigure}
    \caption{The latest version of our cage (left) that houses the Shadow Dexterous Hand, RGB cameras, and the PhaseSpace motion capture system. We made some modifications to the Shadow Dexterous Hand (right) to improve reliability for our setup by moving the PhaseSpace LEDs and cables inside the fingers and by adding rubber to the fingertips.}
    \label{fig:setup-overview}
\end{figure}

Our robot platform is based on the configuration described in~\cite{openai2018learning}.  We still use the Shadow Dexterous E Series Hand (E3M5R) \cite{shadow-robot} as a humanoid robot hand and the PhaseSpace motion capture system to track the Cartesian coordinates of all five fingertips. We use the same 3 RGB Basler cameras for vision pose estimation.

However, a number of improvements have been made since our previous publication. \autoref{fig:setup-cage} depicts the latest iteration of our robot cage. The cage is now fully contained, i.e. all computers are housed within the system. The cage is also on coasters and can therefore be moved more easily. The larger dimensions of the new cage make calibration of the PhaseSpace motion capture system easier and help prevent disturbing calibration when taking the hand in and out of the cage.

We have made a number of customizations to the E3M5R since our last publication (see also \autoref{fig:setup-hand}). We moved routing of the cables that connect the PhaseSpace LEDs on each fingertip to the PhaseSpace micro-driver within the hand, thus reducing the wear and tear on those cables. We worked with The Shadow Robot Company\footnote{\url{https://www.shadowrobot.com/}} to improve the robustness and reliability of some components for which we noticed breakages over time. We also modified the distal part of the fingers to extend the rubber area to cover a larger span to increase the grip of the hand when it interacts with an object. We increased the diameter of the wrist flexion/extension pulley in order to reduce tendon stress which has extended the life of the tendon to more than three times its typical mean time before failure (MTBF). Finally, the tendon tensioners in the hand have been upgraded and this has improved the MTBF of the finger tendons by approximately five to ten times. 

We also made improvements to our software stack that interfaces with the E3M5R. For example, we found that manual tuning of the maximum torque that each motor can exercise was superior to our automated methods in avoiding physical breakage and ensuring consistent policy performance. More concretely, torque limits were minimized such that the hand can reliably achieve a series of commanded positions.

We also invested in real-time system monitoring so that issues with the physical setup could be identified and resolved more quickly. We describe our monitoring system in greater detail in~\autoref{app:hardware-monitoring}.

\subsection{Giiker Cube}

\begin{figure}[h]
    \centering
    \begin{subfigure}[b]{0.48\textwidth}
        \includegraphics[width=\textwidth]{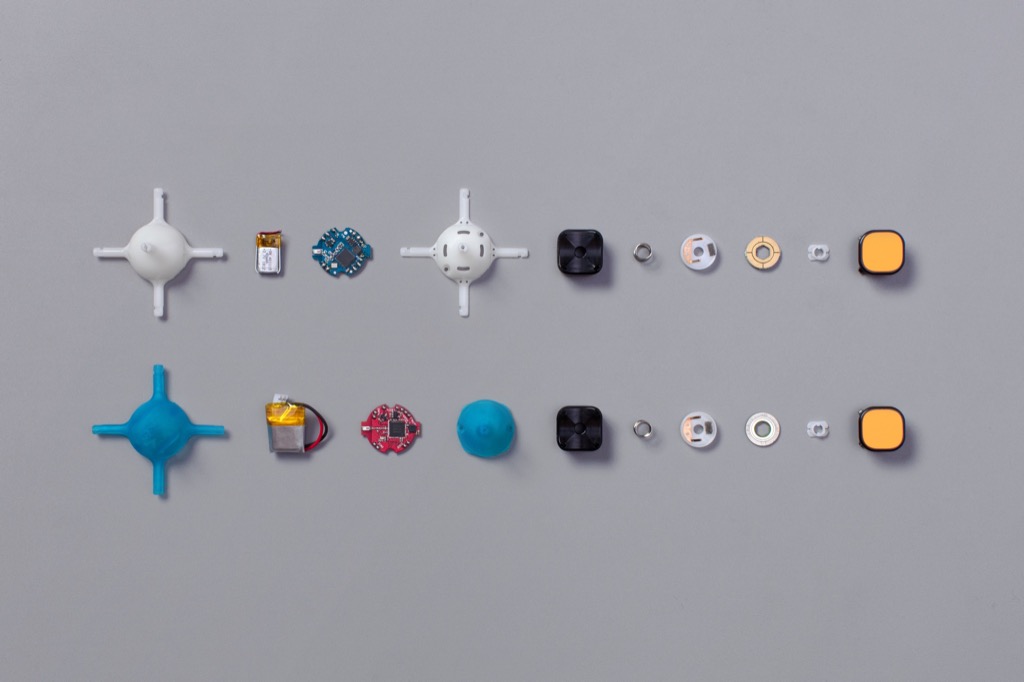}
        \caption{The components of the Giiker cube.}
        \label{fig:setup-giiker}
    \end{subfigure}
    \hfill
    \begin{subfigure}[b]{0.48\textwidth}
        \includegraphics[width=\textwidth]{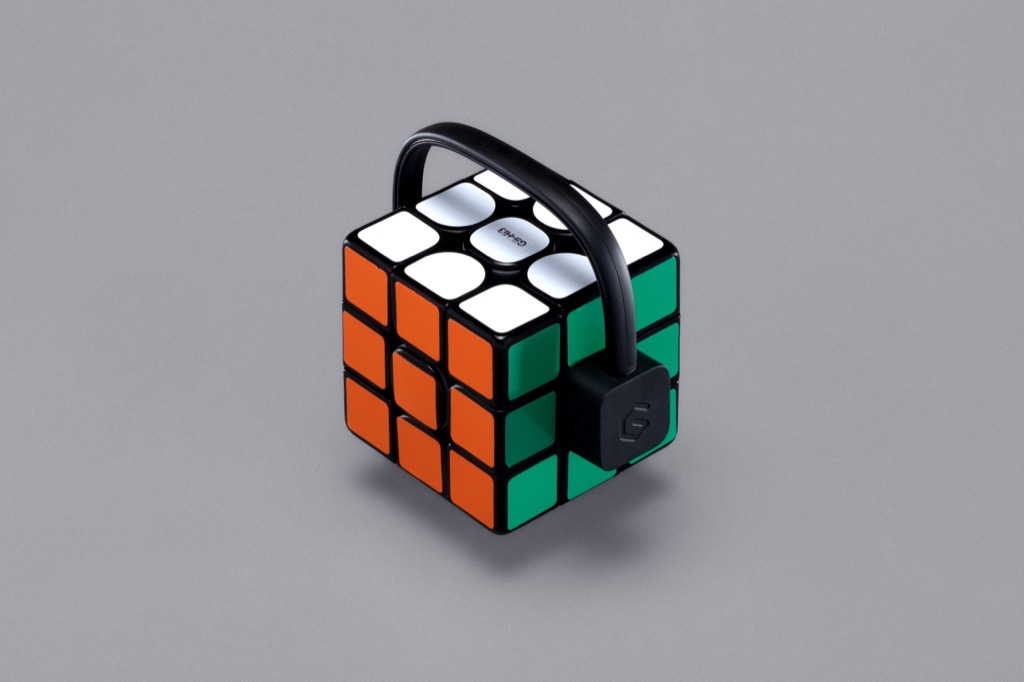}
        \caption{An assembled Giiker cube while charging.}
        \label{fig:setup-giiker2}
    \end{subfigure}
    \caption{We use an off-the-shelf Giiker cube but modify its internals (subfigure a, right) to provider higher resolution for the 6 face angles. The components from left to right are (i) bottom center enclosure, (ii) lithium polymer battery, (iii) main PCBa with BLE, (iv) top center enclosure, (v) cubelet bottom, (vi) compression spring, (vii) contact brushes, (viii) absolute resistive rotary encoder, (ix) locking cap, (x) cubelet top. Once assembled, the Giiker cube can be charged with its ``headphones on'' (right).}
    \label{fig:setup-giiker-overview}
\end{figure}

Sensing the state of a Rubik's cube from vision only is a challenging task. We therefore use a ``smart'' Rubik's cube with built-in sensors and a Bluetooth module as a stepping stone: We used this cube while face angle predictions from vision were not yet ready in order to continue work on the control policy. We also used the Giiker cube for some of our experiments to test the control policy without compounding errors made by the vision model's face angle predictions (we always use the vision model for pose estimation).

Our hardware is based on the Xiaomi Giiker cube.\footnote{\url{https://www.xiaomitoday.com/xiaomi-giiker-m3-intelligent-rubik-cube-review/}} This cube is equipped with a Bluetooth module and allows us to sense the state of the Rubik's cube. However, it only has a face angle resolution of $90\degree$, which is not sufficient for state tracking purposes on the robot setup. We therefore replace some of the components of the original Giiker cube with custom ones in order to achieve a tracking accuracy of approximately $5\degree$. \autoref{fig:setup-giiker} shows the components of the unmodified Giiker cube and our custom replacements side by side, as well as the assembled modified Giiker cube. Since we only use our modified version, we henceforth refer to it as only ``Giiker cube''.

\subsubsection{Design}

\begin{figure}[h]
    \centering
    \begin{subfigure}[b]{0.48\textwidth}
        \centering
        \includegraphics[width=0.5\textwidth]{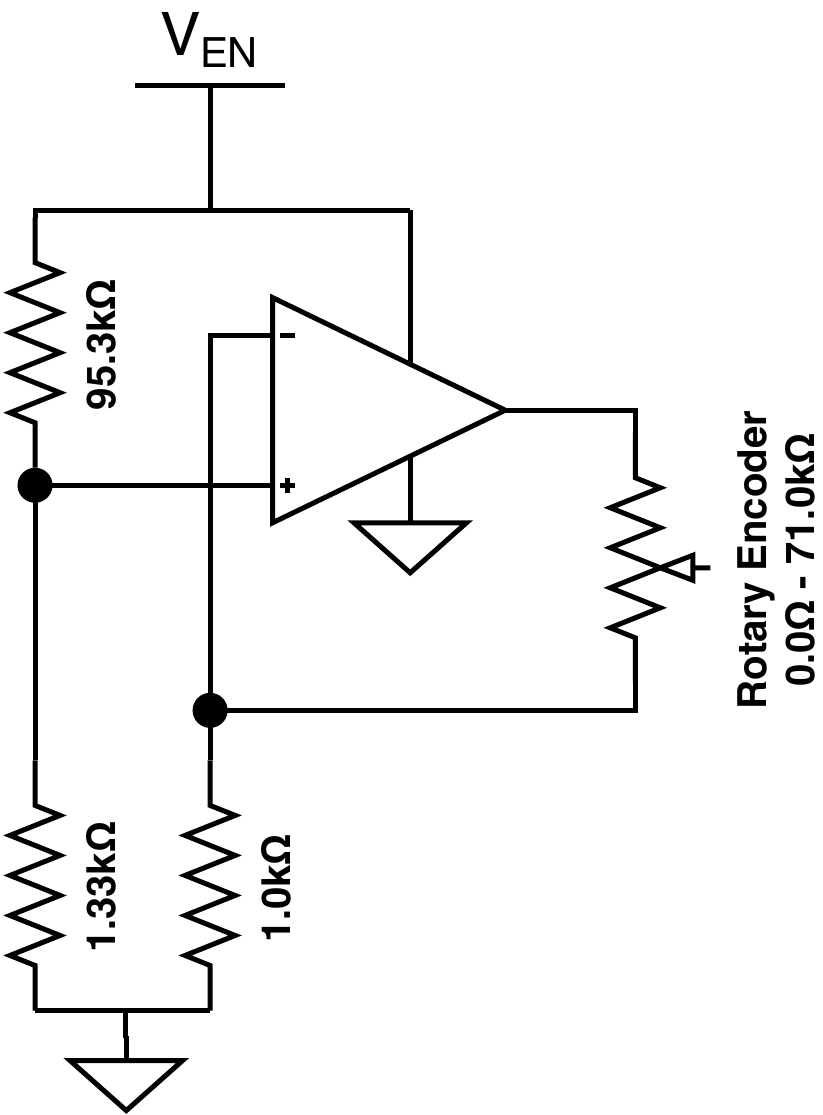}
        \caption{The linearizing circuit used to read the position of the faces.}
        \label{fig:linearizer}
    \end{subfigure}
    \hfill
    \begin{subfigure}[b]{0.48\textwidth}
        \includegraphics[width=\textwidth]{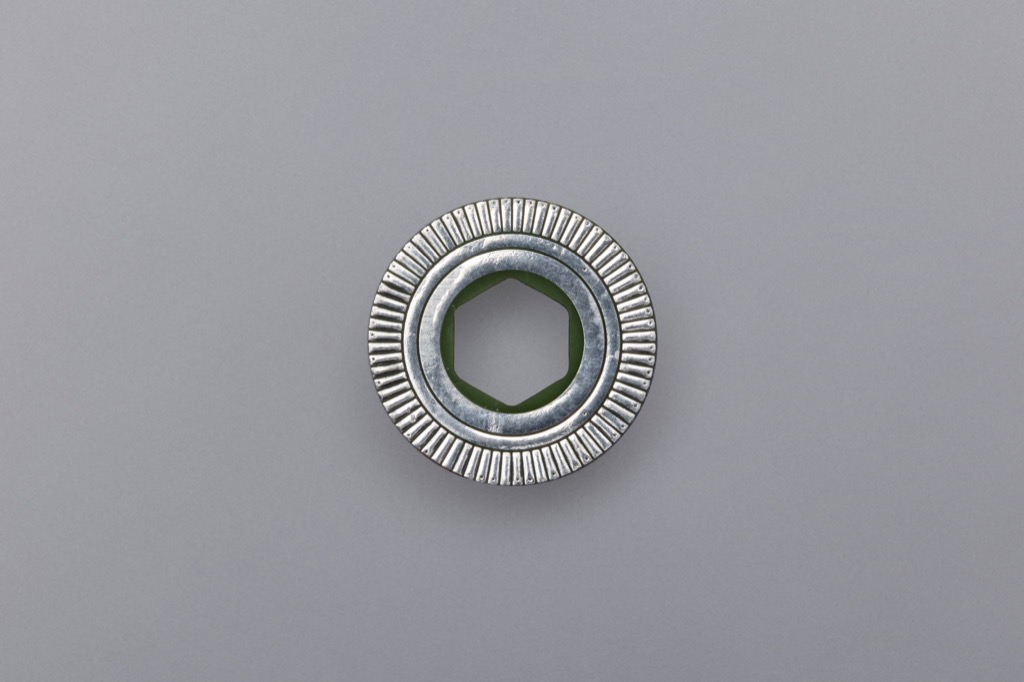}
        \caption{The absolute resistive encoders used to read the position of the faces.}
        \label{fig:encoder}
    \end{subfigure}
\end{figure}

We have redesigned all parts of the Giiker cube but the exterior cubelet elements. The central support was redesigned to move the parting line off of the central line of symmetry to facilitate a more friendly development platform because the off-the-shelf design would have required de-soldering in order to program the microcontroller. The main Bluetooth and signal processing board is based on the NRF52 integrated circuit \cite{nordic}. Six separately printed circuit boards (Figure \ref{fig:encoder}) were designed to improve the resolution from $90\degree$ to $5 \degree$ using an absolute resistive encoder layout. The position is read with a linearizing circuit shown in Figure \ref{fig:linearizer}. The linearized, analog signal is then read by an ADC pin on the microcontroller and sent as a face angle over the Bluetooth Low Energy (BLE) connection to the host.

The custom firmware implements a protocol that is based on the Nordic UART service~(NUS) to emulate a serial port over BLE~\cite{nordic}. We then use a Node.js\footnote{\url{https://nodejs.org/en/}} based client application to periodically request angle readings from the UART module and to send calibration requests to reset angle references when needed. Starting from a solved Rubik's cube, the client is able to track face rotations performed on the cube in real time and thus is able to reconstruct the Rubik's cube state given periodic angle readings. 

\subsubsection{Data Accuracy and Timing}
In order to ensure reliability of physical experiments, we performed regular accuracy tracking tests on integrated Giiker cubes. To assess accuracy, we considered all four right angle rotations as reference points on each cube face and estimated sensor accuracy based on measurements collected at each reference point. Across two custom cubes, the resistive encoders were subject to an absolute mean tracking error of $5.90\degree$ and the standard deviation of reference point readings was $7.61\degree$. 

During our experiments, we used a $12.5$~Hz update frequency\footnote{We run the control policy at this frequency.} for the angle readings, which was sufficient to provide low-latency observations to the robot policy. 

\subsubsection{Calibration}
We perform a combination of firmware and software-side calibration of the sensors to ensure zero-positions can be dynamically set for each face angle sensor. On connecting to a cube for the first time, we record ADC offsets for each sensor in the firmware via a reset request. Furthermore, we add a software-side reset of the angle readings before starting each physical trial on the robot to ensure sensor errors do not accumulate across trials. 

In order to track any physical degradation in the sensor accuracy of the fully custom hardware, we created a calibration procedure which instructs an operator to rotate each face a full $360 \degree$, stopping at each $90 \degree$ alignment of the cube. We then record the expected and actual angles to measure the accuracy over time.

\section{Simulation}
\label{sec:sim}
The simulation setup is similar to~\cite{openai2018learning}: we simulate the physical system with the MuJoCo physics engine~\cite{mujoco}, and we use ORRB~\cite{chociej2019orrb}, a remote rendering backend built on top of Unity3D,\footnote{Unity is a cross-platform game engine. See \url{https://www.unity.com} for more information.} to render synthetic images for training the vision based pose estimator.

While the simulation cannot perfectly match reality, we still found it beneficial to help bridge the gap by modeling our physical setup accurately. Our MuJoCo model of the Shadow Dexterous Hand has thus been further improved since~\cite{openai2018learning} to better match the physical system via new dynamics calibration and modeling of a subset of tendons existing in the physical hand and we developed an accurate model of the Rubik's cube.

\subsection{Hand Dynamics Calibration}

We measured joint positions for the same time series of actions for the real and simulated hands in an environment where the hand can move freely and made two observations:

\begin{figure}[h]
    \centering
    \includegraphics[width=\textwidth]{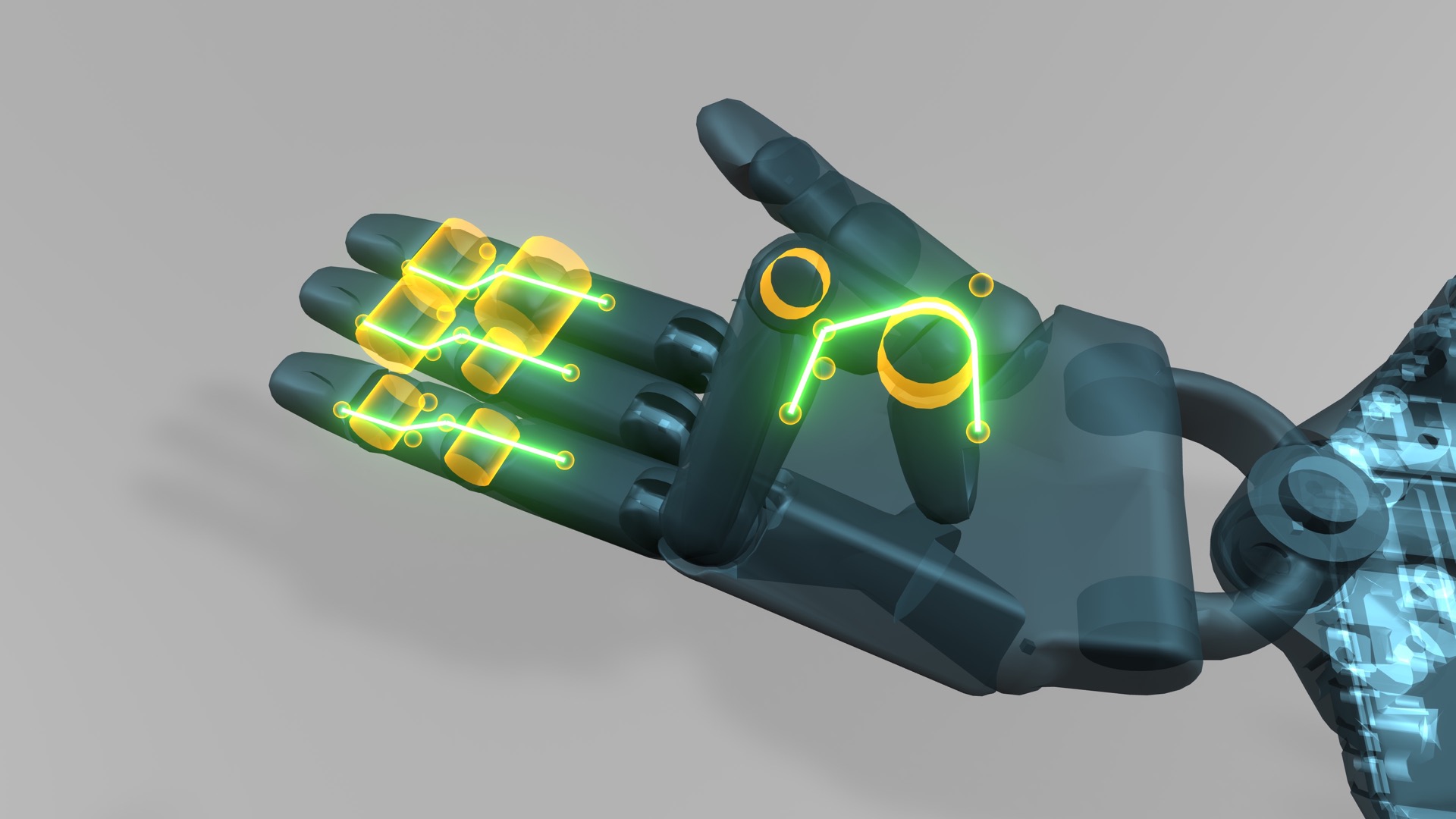}
    \caption{Transparent view of the hand in the new simulation. One spatial tendon (green lines) and two cylindrical geometries acting as pulleys (yellow cylinders) have been added for each non-thumb finger in order to achieve coupled joints dynamics similar to the physical robot.}
    \label{fig:simulation-coupling}
\end{figure}

\begin{figure}[h]
    \centering
    \begin{subfigure}[b]{0.48\textwidth}
        \centering
        \includegraphics[width=\textwidth]{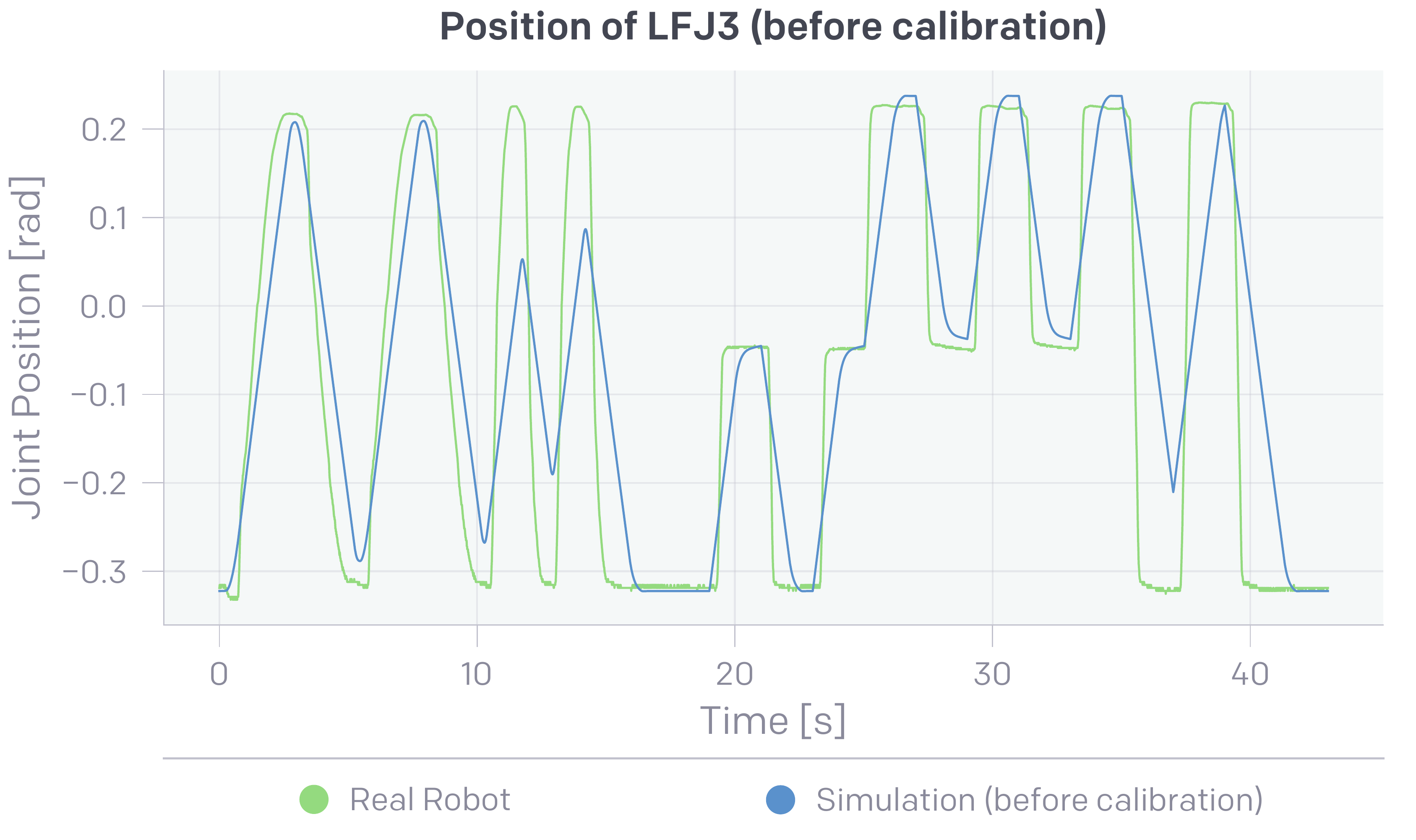}
        \caption{Comparison against original simulation.}
        \label{fig:simulation-before-calibration}
    \end{subfigure}
    \hfill
    \begin{subfigure}[b]{0.48\textwidth}
        \centering
        \includegraphics[width=\textwidth]{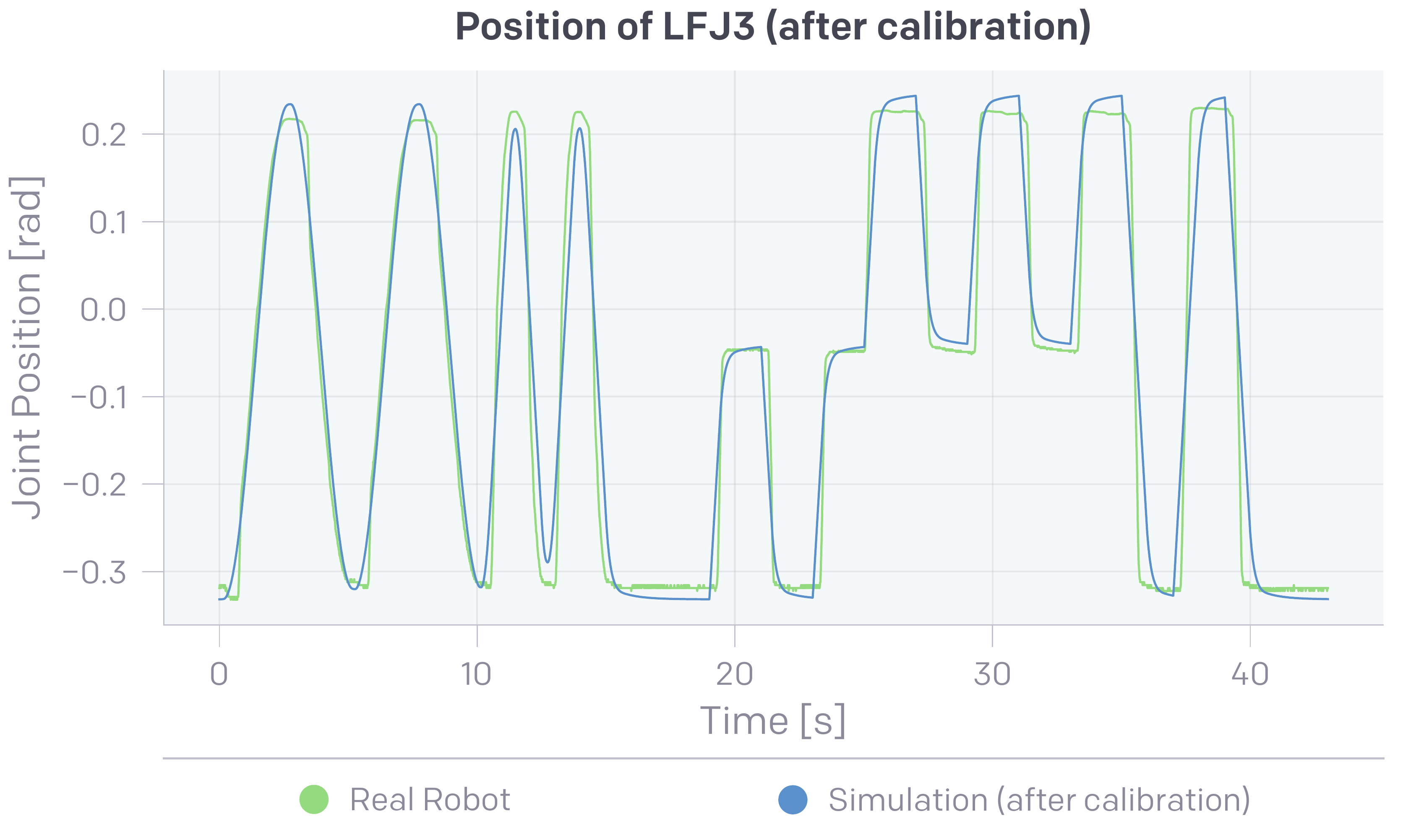}
        \caption{Comparison against new simulation.}
        \label{fig:simulation-after-calibration}
    \end{subfigure}
    \caption{Comparison of positions of the LFJ3 joint on a real and simulated robot hand for the same control sequence, for the original simulation (a) and for the new simulation (b)}
    \label{fig:simulation-calibration}
\end{figure}

\begin{enumerate}
  \item The joint positions recorded on a physical robot and in simulation were visibly different (see \autoref{fig:simulation-before-calibration}).
  \item The dynamics of \emph{coupled joints} (i.e. distal two joints of non-thumb fingers, see~\cite[Appendix~B.1]{openai2018learning}) were different on a physical robot and in simulation. In the original simulation used in \cite{openai2018learning}, movement of coupled joints was modeled with two fixed tendons which resulted in both joints traveling roughly the same distance for each action. However, on the physical robot, movement of coupled joints depends on the current position of each joint. For instance, like in the human hand, the proximal segment of a finger bends before the distal segment when bending a finger.
\end{enumerate}

To address the dynamics of coupled joints, we added a non-actuated spatial tendon and pulleys to the simulated non-thumb fingers (see \autoref{fig:simulation-coupling}), analogous to the non-actuated tendon present in the physical robot.
Parameters relevant to the joint movement in the new MuJoCo model were then calibrated to minimize root mean square error between reference joint positions recorded on a physical robot and joint positions recorded in simulation for the same time series of actions. We observe that better modeling of coupling and dynamics calibration improves performance significantly and present full results in \autoref{app:full-results-sim-calib}. We use this version of the simulation throughout the rest of this work.

\subsection{Rubik's Cube}

Behind the apparent simplicity of a cube-like exterior, a Rubik's cube hides a high degree of internal complexity and surprisingly nontrivial interactions between elements. A regular 3x3x3 cube consists of 26 externally facing \emph{cubelets} that are bound together to constitute a larger cubic shape. Six cubelets that reside in the center of each face are connected by axles to the inner core and can only rotate in place with one degree of freedom. In contrast to that, the edge and corner cubelets are not fixed and can move around the cube whenever the larger faces are rotated. To prevent the cube from falling apart, these cubelets have little plastic tabs that extend towards the core and allow each piece to be held in place by its neighbors, which in the end are retained by the center elements. Additionally, most Rubik's cubes are to a certain degree elastic and allow for small deformations from their original shape, constituting additional degrees of freedom.

\begin{figure}[h]
    \centering
    \begin{subfigure}[b]{0.48\textwidth}
        \centering
        \includegraphics[width=\textwidth]{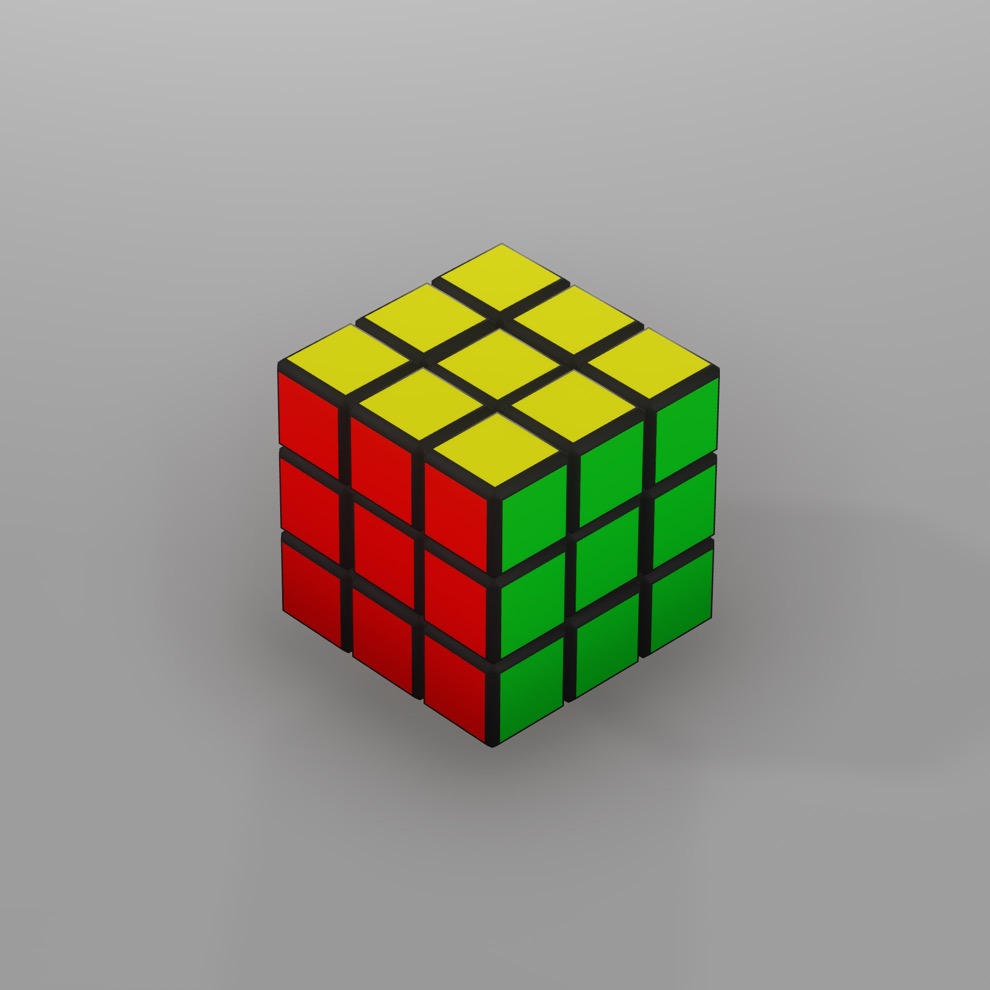}
        \caption{Rendering of the cube model.}
    \end{subfigure}
    \hfill
    \begin{subfigure}[b]{0.48\textwidth}
        \centering
        \includegraphics[width=\textwidth]{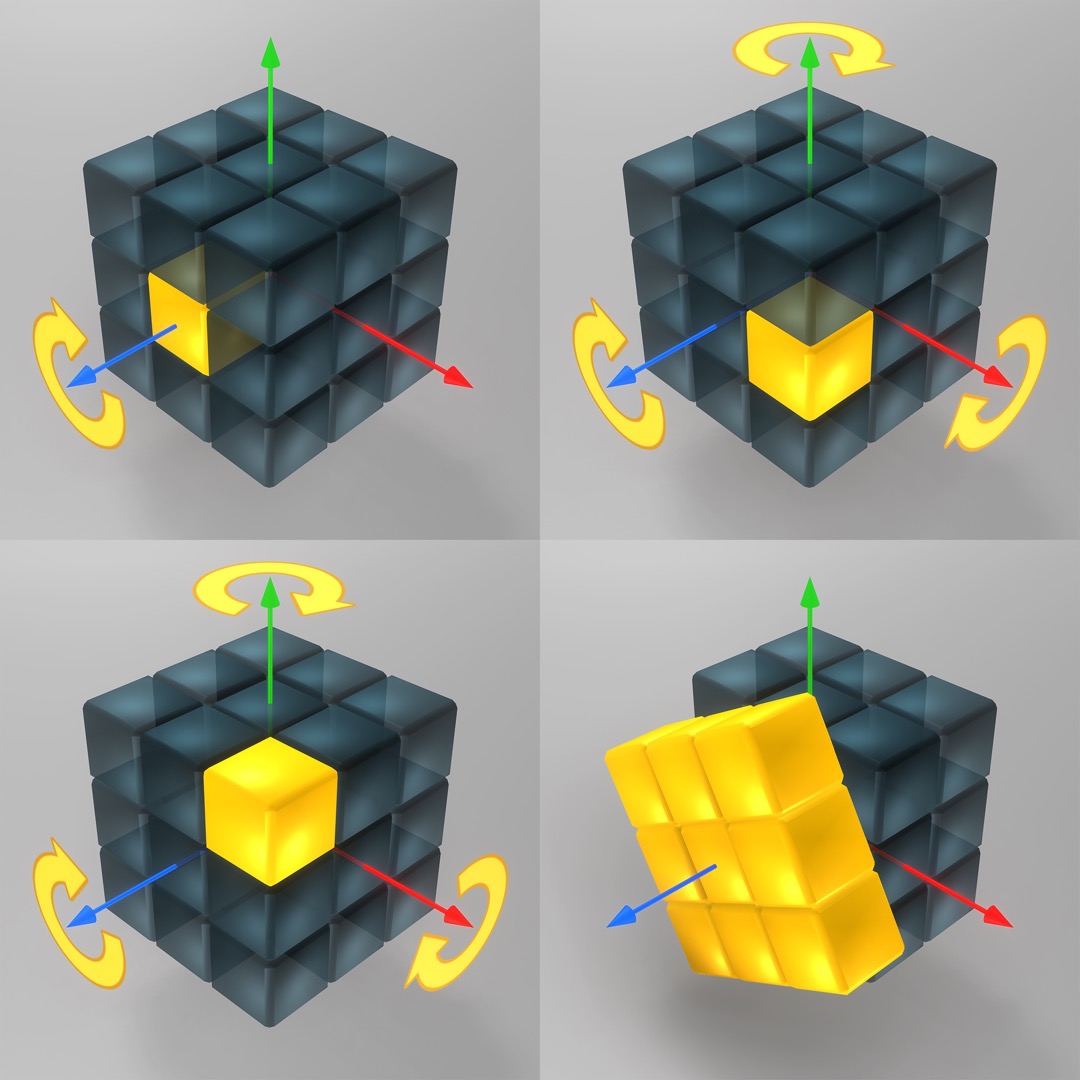}
        \caption{Rendering of the different axis.}
    \end{subfigure}
    \caption{Our MuJoCo model of the Rubik's cube. On the left, we show a rendered version. On the right, we show the individual cublets that make up our model and visualize the different axis and degrees of freedom of our model.}
    \label{fig:simulation-cube}
\end{figure}

The components of the cube constantly exert pressure on each other which results in a certain base level of friction in the system both between the cubelets and in the joints. It is enough to apply force to a single cubelet to rotate a face, as it will be propagated between the neighboring elements via contact forces. Although a cube has six faces that can be rotated, not all of them can be rotated simultaneously -- whenever one face has already been moved by a certain angle, perpendicular faces are in a locked state and prevented from moving. However, if this angle is small enough, the original face often "snaps" back into its nearest aligned state and in that way we can proceed with rotating the perpendicular face. This property is commonly called the "forgiveness" of a Rubik's Cube and its strength varies greatly among models available on the market.

Since we train entirely in simulation and need to successfully transfer to the real world without ever experiencing it, we needed to create a model rich enough to include all of the aforementioned behaviors, while at the same time keeping software complexity and computational costs manageable. We used the MuJoCo~\cite{mujoco} physics engine, which implements a stable and fast numerical solutions for simulating body dynamics with soft contacts.

Inspired by the physical cube, our simulated model consists of 26 rigid body convex cubelets. MuJoCo allows for these shapes to penetrate each other by a small margin when a force is applied. Six central cubelets have a single \emph{hinge joint} representing a single rotational degree of freedom about the axes running through the center of the cube orthogonal to each face. All remaining 20 corner and edge cubelets have three hinge joints corresponding to full Euler angle representation, with rotation axes passing through the center of the cube. In that way, our cube has $6 \times 1 + 20 \times 3 = 66$ degrees of freedom, that allow us to represent effectively not only \emph{43 quintillion} fully aligned cube configurations but also all physically valid intermediate states.

Each cubelet mesh was created on the basis of the cube of size 1.9 cm. Our preliminary experiments have shown that with perfectly cubic shape, the overall Rubik's cube model was highly unforgiving. Therefore, we beveled all the edges of the mesh 1.425 mm inwards, which gave satisfactory results.\footnote{Real Rubik's cubes also have cubelets with rounded corners, for the same reason.}. We do not implement any custom physics in our modelling, but rely on the cubelet shapes, contact forces and friction to drive the movement of the cube. We conducted experiments with spring joints which would correspond to additional degrees of freedom for cube deformation, but found they were not necessary and that native MuJoCo soft contacts already exhibit similar dynamics.

We performed a very rudimentary dynamics calibration of the parameters which MuJoCo allows us to specify, in order to roughly match a physical Rubik's cube. Our goal was not to get an exact match, but rather to have a plausible model as a starting point for domain randomization.

\section{Automatic Domain Randomization}
\label{sec:adr}
In~\cite{openai2018learning}, we were able to train a control policy and a vision model in simulation and then transfer both to a real robot through the use of domain randomization~\cite{tobin2017domain, peng2017sim}. However, this required a significant amount of manual tuning and a tight iteration loop between randomization design in simulation and validation on a robot. In this section, we describe how \emph{automatic domain randomization} (ADR) can be used to automate this process and how we apply ADR to both policy and vision training.

Our main hypothesis that motivates ADR is that \textit{training on a maximally diverse distribution over environments leads to transfer via emergent meta-learning}. More concretely, if the model has some form of memory, it can learn to adjust its behavior during deployment to improve performance on the current environment over time, i.e. by implementing a learning algorithm internally. We hypothesize that this happens if the training distribution is so large that the model cannot memorize a special-purpose solution per environment due to its finite capacity. ADR is a first step in this direction of unbounded environmental complexity: it automates and gradually expands the randomization ranges that parameterize a distribution over environments. Related ideas were also discussed in~\cite{DBLP:journals/corr/DuanSCBSA16, botvinick2019reinforcement, wang2019poet, clune2019ai}.

In the remainder of this section, we first describe how ADR works at a high level and then describe the algorithm and our implementation in greater detail.

\subsection{ADR Overview}
We use ADR both to train our vision models (supervised learning) and our policy (reinforcement learning). In each case, we generate a distribution over environments by randomizing certain aspects, e.g. the visual appearance of the cube or the dynamics of the robotic hand. While domain randomization requires us to define the ranges of this distribution manually and keep it fixed throughout model training, in ADR the distribution ranges are defined automatically and allowed to change.

\begin{figure}[h]
    \centering
    \includegraphics[width=0.6\textwidth]{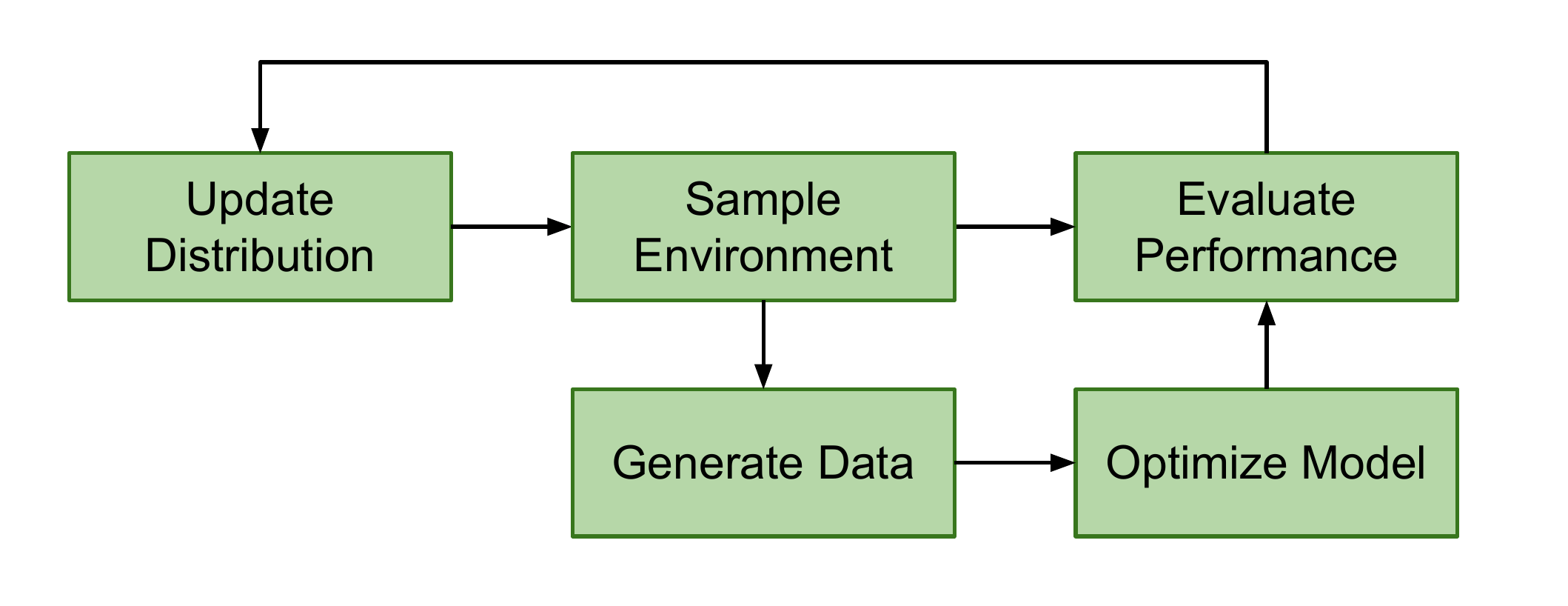}
    \caption{Overview of ADR. ADR controls the distribution over environments. We sample environments from this distribution and use it to generate training data, which is then used to optimize our model (either a policy or a vision state estimator). We further evaluate performance of our model on the current distribution and use this information to update the distribution over environments automatically.}
    \label{fig:adr-architecture-top}
\end{figure}

A top-level diagram of ADR is given in~\autoref{fig:adr-architecture-top}. We give an intuitive overview of ADR below. See \autoref{sec:adr_algorithm} for a formal description of the algorithm.

At its core, ADR realizes a training curriculum that gradually expands a distribution over environments for which the model can perform well. The initial distribution over environments is concentrated on a single environment. For example, in policy training the initial environment is based on calibration values measured from the physical robot.

The distribution over environments is sampled to obtain environments used to generate training data and evaluate model performance. ADR is independent of the algorithm used for model training. It only generates training data. This allows us to use ADR for both policy and vision model training.

As training progresses and model performance improves sufficiently on the initial environment, the distribution is expanded. This expansion continues as long as model performance is considered acceptable. With a sufficiently powerful model architecture and training algorithm, the distribution is expected to expand far beyond manual domain randomization ranges since every improvement in the model's performance results in an increase in randomization.

ADR has two key benefits over manual domain randomization (DR):
\begin{itemize}
    \item Using a curriculum that gradually increases difficulty as training progresses simplifies training, since the problem is first solved on a single environment and additional environments are only added when a minimum level of performance is achieved~\cite{graves2017curriculum, matiisen2017teacherstudent}.
    \item It removes the need to manually tune the randomizations. This is critical, because as more randomization parameters are incorporated, manual adjustment becomes increasingly difficult and non-intuitive.
\end{itemize}

Acceptable performance is defined by \emph{performance thresholds}. For policy training, they are configured as the lower and upper bounds on the number of successes in an episode. For vision training, we first configure target performance thresholds for each output (e.g. position, orientation). During evaluation, we then compute the percentage of samples which achieve these targets for all outputs; if the resulting percentage is above the upper threshold or below the lower threshold, the distribution is adjusted accordingly.

\subsection{Algorithm}\label{sec:adr_algorithm}

Each environment $e_\lambda$ is parameterized by $\lambda \in \mathbb{R}^d$, where $d$ is the number of parameters we can randomize in simulation. In domain randomization (DR), the environment parameter $\lambda$ comes from a \emph{fixed} distribution $P_\phi$ parameterized by $\phi \in \mathbb{R}^{d'}$. However, in automatic domain randomization (ADR), $\phi$ is \emph{changing} dynamically with training progress. The sampling process in \autoref{fig:adr-architecture-top} works out as $\lambda \sim P_\phi$, resulting in one randomized environment instance $e_\lambda$.

To quantify the amount of ADR expansion, we define \emph{ADR entropy} as $\mathcal{H}(P_\phi) = -\frac{1}{d}\int P_{\phi}(\lambda) \log P_{\phi}(\lambda) d\lambda$ 
in units of nats/dimension. The higher the ADR entropy, the broader the randomization sampling distribution. The normalization allows us to compare between different environment parameterizations.

In this work, we use a factorized distribution parameterized by $d' = 2d$ parameters. To simplify notation, let $\phi^L, \phi^H \in \R^d$ be a certain partition of $\phi$. For the $i$-th ADR parameter $\lambda_i$, $i = 1, \dots, d$, the pair $(\phi^L_i,\, \phi^H_i)$ is used to describe a uniform distribution for sampling $\lambda_i$ such that $\lambda_i \sim U(\phi^L_i,\, \phi^H_i)$. Note that the boundary values are inclusive. The overall  distribution is given by
\begin{equation*}
P_{\phi}(\lambda) = \prod_{i=1}^d U(\phi^L_i,\, \phi^H_i)
\end{equation*}
with ADR entropy
\begin{equation*}
\mathcal{H}(P_\phi) = \frac{1}{d}\sum_{i=1}^d \log(\phi^H_i - \phi^L_i).
\end{equation*}

The ADR algorithm is listed in Algorithm~\ref{algorithm:adr}. For the factorized distribution, Algorithm~\ref{algorithm:adr} is applied to $\phi^L$ and $\phi^H$ separately.

At each iteration, the ADR algorithm randomly selects a dimension of the environment $\lambda_i$ to fix to a boundary value $\phi^L_i$ or $\phi^H_i$ (we call this ``boundary sampling''), while the other parameters are sampled as per $P_{\phi}$. Model performance for the sampled environment is then evaluated and appended to the buffer associated with the selected boundary of the selected parameter. Once enough performance data is collected, it is averaged and compared to thresholds. If the average model performance is better than the high threshold $t_H$, the parameter for the chosen dimension is increased. It is decreased if the average model performance is worse than the low threshold $t_L$.

\begin{algorithm}
\caption{ADR}\label{algorithm:adr}
\begin{algorithmic}
\Require $\phi^{0}$ \Comment{Initial parameter values}
\Require $\{D^L_i, D^H_i\}_{i=1}^d$ \Comment{Performance data buffers}
\Require $m$, $t_L$, $t_H$, where $t_L < t_H$ \Comment{Thresholds}
\Require $\Delta$ \Comment{Update step size}
\State $\phi \leftarrow \phi^{0}$
\Repeat
    \State $\lambda \sim P_{\phi}$
    \State $i \sim U\{1, \dots, d\}$, $x \sim U(0, 1)$
    \If{$x < 0.5$}
        \State $D_i \gets D^L_i$, $\lambda_i \gets \phi^L_i$ \Comment{Select the lower bound in ``boundary sampling''}
    \Else{}
        \State $D_i \gets D^H_i$, $\lambda_i \gets \phi^H_i$ \Comment{Select the higher bound in ``boundary sampling''}
    \EndIf
    \State $p \gets$ \Call{EvaluatePerformance}{$\lambda$}  \Comment{Collect model performance on environment parameterized by $\lambda$}
    \State $D_i \gets D_i \cup \{p\}$  \Comment{Add performance to buffer for $\lambda_i$, which was boundary sampled}
    \If{\Call{Length}{$D_i$} $\geq m$}
        \State $\bar{p} \gets$ \Call{Average}{$D_i$}
        \State \Call{Clear}{$D_i$}
        \If{$\bar{p} \geq t_H$}
            \State $\phi_i \gets \phi_i + \Delta$
        \ElsIf{$\bar{p} \leq t_L$}
            \State $\phi_i \gets \phi_i - \Delta$
        \EndIf
    \EndIf
\Until{training is complete}
\end{algorithmic}
\end{algorithm}

As described, the ADR algorithm modifies $P_{\phi}$ by always fixing one environment parameter to a boundary value. To generate model training data, we use Algorithm~\ref{algorithm:generate_data} in conjunction with ADR. The algorithm samples $\lambda$ from $P_{\phi}$ and runs the model in the sampled environment to generate training data.

To combine ADR and training data generation, at every iteration we execute Algorithm~\ref{algorithm:adr} with probability $p_b$ and Algorithm~\ref{algorithm:generate_data} with probability $1 - p_b$. We refer to $p_b$ as the \emph{boundary sampling probability}.

\begin{algorithm}
\caption{Training Data Generation}\label{algorithm:generate_data}
\begin{algorithmic}
\Require $\phi$ \Comment{ADR distribution parameters}
\Repeat
    \State $\lambda \sim P_{\phi}$
    \State \Call{GenerateData}{$\lambda$}
\Until{training is complete}
\end{algorithmic}
\end{algorithm}

\subsection{Distributed Implementation}

\begin{figure}[h]
    \centering
    \begin{subfigure}[b]{0.48\textwidth}
        \centering
        \includegraphics[width=\textwidth]{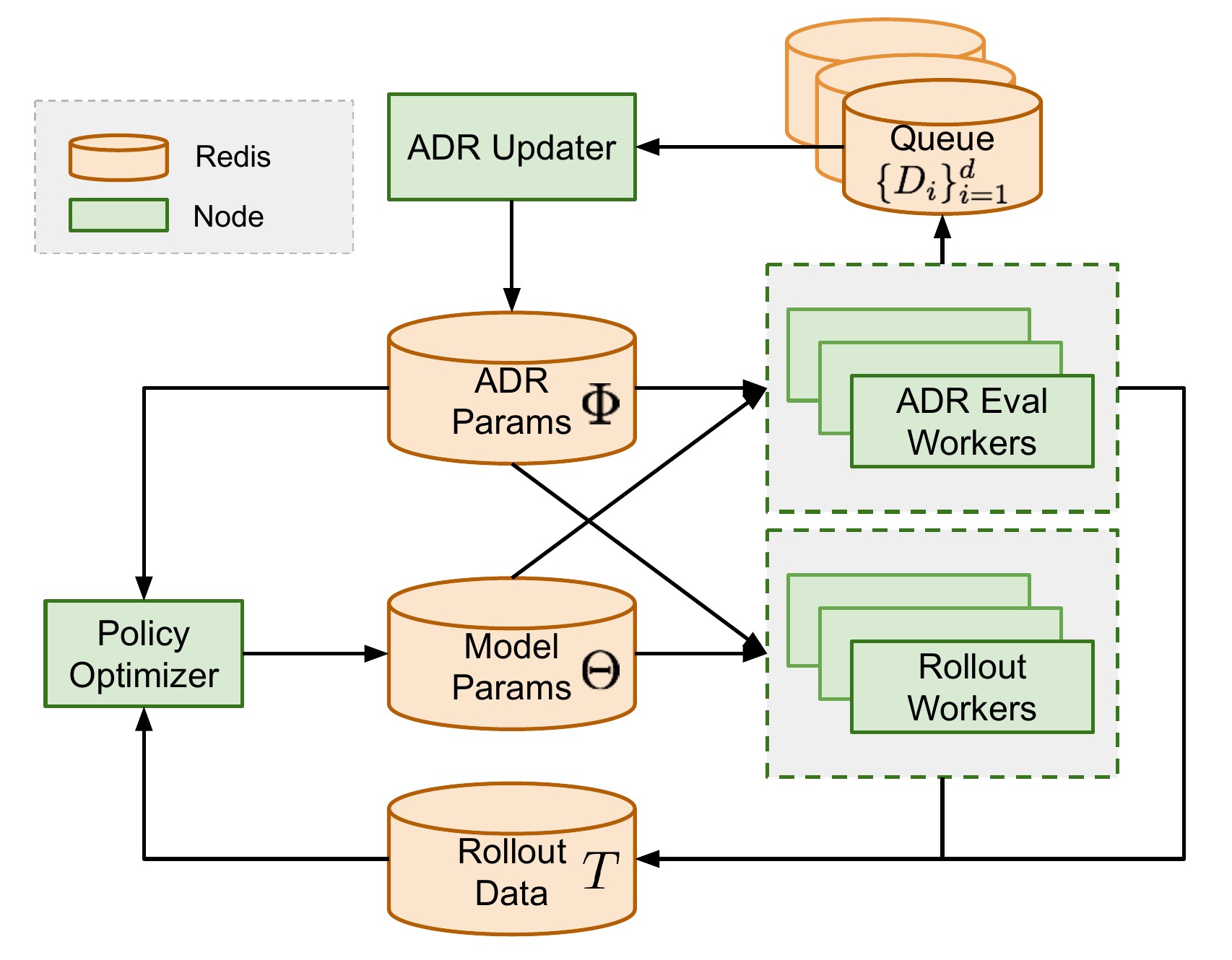}
        \caption{Policy training architecture.}
        \label{fig:adr-architecture-policy}
    \end{subfigure}
    \hfill
    \begin{subfigure}[b]{0.48\textwidth}
        \centering
        \includegraphics[width=\textwidth]{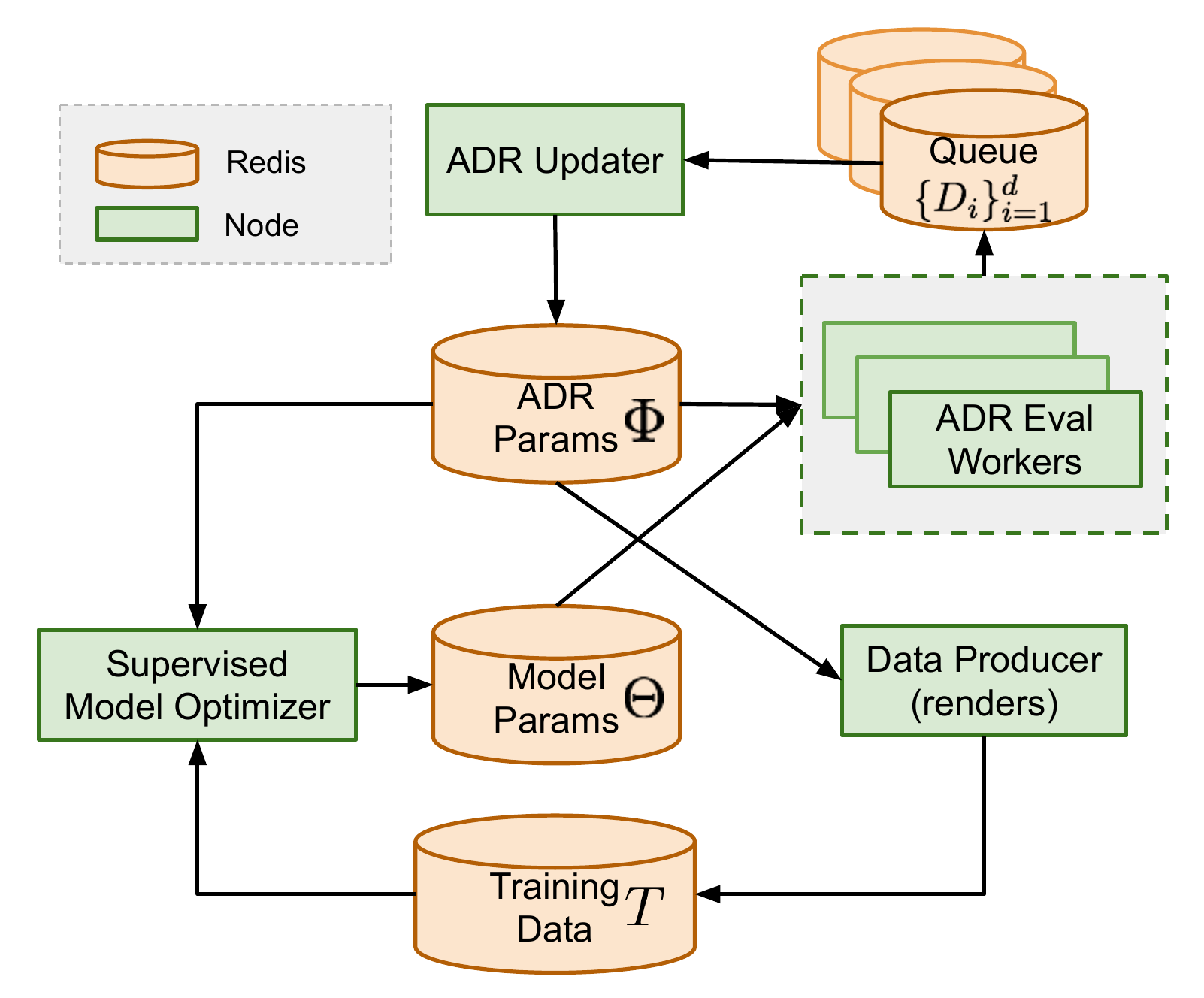}
        \caption{Vision training architecture.}
        \label{fig:adr-architecture-vision}
    \end{subfigure}
    \caption{The distributed ADR architecture for policy (left) and vision (right). In both cases, we use Redis for centralized storage of ADR parameters ($\Phi$), model parameters ($\Theta$), and training data ($T$). ADR eval workers run Algorithm~\ref{algorithm:adr} to estimate performance using boundary sampling and report results using performance buffers ($\{D_i\}_{i=1}^d$). The ADR updater uses those buffers to obtain average performance and increases or decreases boundaries accordingly. Rollout workers (for the policy) and data producers (for vision) produce data by sampling an environment as parameterized by the current set of ADR parameters (see Algorithm~\ref{algorithm:generate_data}). This data is then used by the optimizer to improve the policy and vision model, respectively.}
    \label{fig:adr-architecture}
\end{figure}

We used a distributed version of ADR in this work. The system architecture is illustrated in \autoref{fig:adr-architecture} for both our policy and vision training setup. We describe policy training in greater detail in \autoref{sec:policy} and vision training in \autoref{sec:vision}. Here we focus on ADR.

The highly-parallel and asynchronous implementation depends on several centralized storage of (policy or vision) model parameters $\Theta$,  ADR parameters $\Phi$, training data $T$, and performance data buffers $\{D_i\}_{i=1}^d$. We use Redis to implement them.

By using centralized storage, the ADR algorithm is decoupled from model optimization. However, to train a good policy or vision model using ADR, it is necessary to have a concurrent optimizer that consumes the training data in $T$ and pushes updated model parameters to $\Theta$.

We use $W$ parallel worker threads instead of the sequential while-loop. For training the policy, each worker pulls the latest distribution and model parameters from $\Phi$ and $\Theta$ and executes Algorithm~\ref{algorithm:adr} with probability $p_b$ (denoted as ``ADR Eval Worker'' in \autoref{fig:adr-architecture-policy}). Otherwise, it executes Algorithm~\ref{algorithm:generate_data} and pushes the generated data to $T$ (denoted as ``Rollout Worker'' in \autoref{fig:adr-architecture-policy}). To avoid wasting a large amount of data for only ADR, we also use this data to train the policy. The setup for vision is similar. Instead of rolling out a policy, we use the ADR parameters to render images and use those to train the supervised vision state estimator. Since data is cheaper to generate, we do not use the ADR evaluator data to train the model in this case but only used the data produced by the ``Data Producer'' (compare \autoref{fig:adr-architecture-vision}).

In the policy model, $\phi^{0}$ is set based on a calibrated environment parameter according to $\phi^{0, L}_i = \phi^{0, H}_i = \lambda_{i}^\text{calib}$ for all $i = 1, \dots, d$. In the vision model, the initial randomizations are set to zero, i.e. $\phi^{0, L}_i = \phi^{0, H}_i = 0$. The distribution parameters are pushed to $\Phi$ to be used by all workers at the beginning of the algorithm.

\subsection{Randomizations}

Here, we describe the categories of randomizations used in this work. The vast majority of randomizations are for a scalar environment parameter $\lambda_i$ and are parameterized in ADR by two boundary parameters $(\phi^L_i, \phi^H_i)$. For a full listing of randomizations used in policy and vision training, see~\autoref{app:randomizations}.

A few randomizations, such as observation noise, are controlled by more than one environment parameter and are parameterized by a larger set of boundary parameters. For full details on these randomizations and their ADR parameterization, see~\autoref{app:randomizations}.

\paragraph{Simulator physics.} We randomize simulator physics parameters such as geometry, friction, gravity, etc. See \autoref{sec:physics_randomizations} for details of their ADR parameterization.

\paragraph{Custom physics.} We model additional physical robot effects that are not modelled by the simulator, for example, action latency or motor backlash. See~\citep[Appendix~C.2]{openai2018learning} for implementation details of these models. We randomize the parameters in these models in a similar way to simulator physics randomizations.

\paragraph{Adversarial.} We use an adversarial approach similar to~\cite{pinto2017supervision, pinto2017robust} to capture any remaining unmodeled physical effects in the target domain. However, we use random networks instead of a trained adversary. See \autoref{sec:random_network_adversary} for details on implementation and ADR parameterization.

\paragraph{Observation.} We add Gaussian noise to policy observations to better approximate observation conditions in reality. We apply both correlated noise, which is sampled once at the start of an episode and uncorrelated noise, which is sampled at each time step. We randomize the parameters of the added noise. See \autoref{sec:observation_randomizations} for details of their ADR parameterization.

\paragraph{Vision.} We randomize several aspects in ORRB~\cite{chociej2019orrb} to control the rendered scene, including lighting conditions, camera positions and angles, materials and appearances of all the objects, the texture of the background, and the post-processing effects on the rendered images. See \autoref{app:visual_randomizations} for details.

\section{Policy Training in Simulation}
\label{sec:policy}
In this section we describe how we train control policies using Proximal Policy Optimization~\cite{ppo} and reinforcement learning. Our setup is similar to~\cite{openai2018learning}. However, we use ADR as described in \autoref{sec:adr} to train on a large distribution over randomized environments.

\subsection{Actions, Rewards, and Goals}
Our setup for the action space and rewards is unchanged from~\cite{openai2018learning} so we only briefly recap them here. We use a discretized action space with $11$~bins per actuated joint (of which there are $20$). We use a multi-categorical distribution. Actions are relative changes in generalized joint position coordinates.

There are three types of rewards we provide to our agent during training:
\begin{enumerate*}[label=(\alph*)]
    \item The difference between the previous and the current distance of the system state from the goal state,
    \item an additional reward of $5$ whenever a goal is achieved,
    \item and a penalty of $-20$ whenever a cube/block is dropped.
\end{enumerate*}

We generate random goals during training. For the block, the target rotation is randomly sampled but constrained such that any face points directly upwards. For the Rubik's cube the task generation is slightly more convoluted as it depends on the state of the cube at the time when the goal is generated. If the cube faces are not aligned, we make sure to align them and additionally rotate the whole cube according to a sampled random orientation just like with the block (called a flip). Alternatively, if the faces \emph{are} aligned, we rotate the top cube face with 50\% probability either clockwise or counter-clockwise. Otherwise we again perform a flip. Detailed listings of the goal generation algorithms can be found in the \autoref{app:goal-generation}.

We consider a training episode to be finished whenever one of the following conditions is satisfied:
\begin{enumerate*}[label=(\alph*)]
    \item the agent achieves 50 consecutive successes (of reaching a goal within the required threshold),
    \item the agent drops the cube,
    \item or the agent times out when trying to reach the next goal. Time out limits are 400 timesteps for block reorientation and 800 timesteps\footnote{We use $1600$~timesteps when training from scratch.} for the Rubik's Cube.
\end{enumerate*}

\subsection{Policy Architecture}

We base our policy architecture on~\citep{openai2018learning} but extend it in a few important ways. The policy is still recurrent since only a policy with access to some form of memory can perform meta-learning. We still use a single feed-forward layer with a ReLU activation~\citep{relu} followed by a single LSTM layer~\citep{lstm}. However, we increase the capacity of the network by doubling the number of units: the feed-forward layer now has $2048$~units and the LSTM layer has $1024$~units.

The value network is separate from the policy network (but uses the same architecture) and we project the output of the LSTM onto a scalar value. We also add L2 regularization with a coefficient of $10^{-6}$ to avoid ever-growing weight norms for long-running experiments.

\begin{figure}[h]
    \centering
    \includegraphics[width=\textwidth]{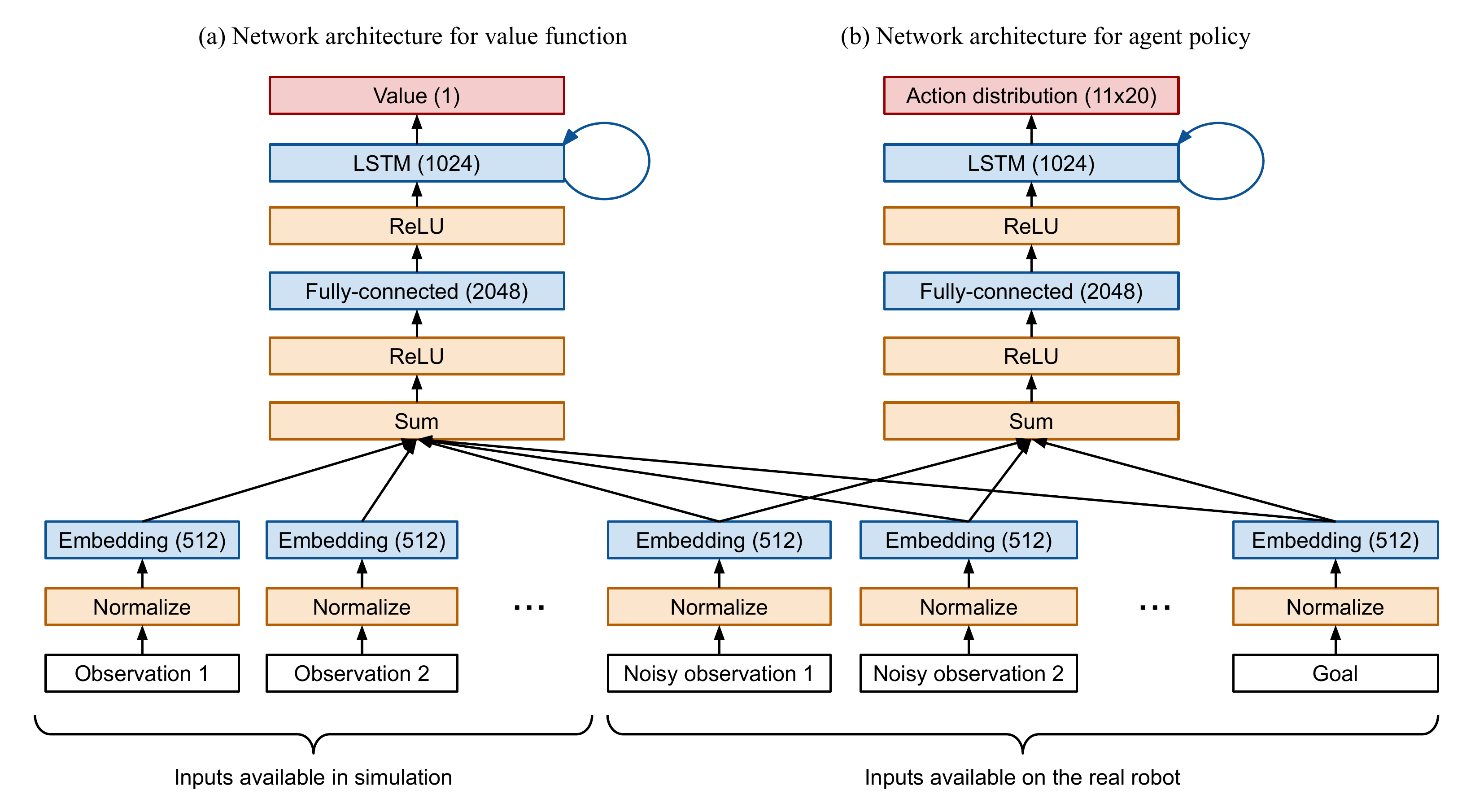}
    \caption{Neural network architecture for (a) value network and (b) policy network.}
    \label{fig:policy-arch}
\end{figure}

\begin{table}[h!]
    \footnotesize
    \centering
    \caption{Inputs for the Rubik's cube task of the policy and value networks, respectively.}
    \renewcommand{\arraystretch}{1.3}
    \begin{tabular}{@{}llcc@{}}
        \toprule
        \textbf{Input} & \textbf{Dimensionality} & \textbf{Policy network} & \textbf{Value network} \\
        \midrule
        Fingertip positions & 15D & $\times$ & \checkmark \\
        Noisy fingertip positions & 15D & \checkmark & \checkmark \\
        Cube position & 3D & $\times$ & \checkmark \\
        Noisy cube position & 3D & \checkmark & \checkmark \\
        Cube orientation & 4D (quaternion) & $\times$ & \checkmark \\
        Noisy cube orientation & 4D (quaternion) & \checkmark & \checkmark \\
        Goal orientation & 4D (quaternion) & \checkmark & \checkmark \\
        Relative goal orientation & 4D (quaternion) & $\times$ & \checkmark \\
        Noisy relative goal orientation & 4D (quaternion) & \checkmark & \checkmark \\
        Goal face angles & 12D\footnote{\label{footnote:angles2}Angles are encoded as $\sin$ and $\cos$, i.e. this doubles the dimensionality of the underlying angle.} & \checkmark & \checkmark \\
        Relative goal face angles & 12D\footref{footnote:angles2} & $\times$ & \checkmark \\
        Noisy relative goal face angles & 12D\footref{footnote:angles2} & \checkmark & \checkmark \\
        Hand joint angles & 48D\footref{footnote:angles2} & $\times$ & \checkmark \\
        All simulation positions \& orientations (\texttt{qpos}) & 170D & $\times$ & \checkmark \\
        All simulation velocities (\texttt{qvel}) & 168D & $\times$ & \checkmark \\
        \bottomrule
    \end{tabular}
\label{table:policy-inputs-full2}
\end{table}

An important difference between our architecture and the architecture used in~\citep{openai2018learning} is how inputs are handled. In \citep{openai2018learning}, the inputs for the policy and value networks consisted of different observations (e.g. fingertip positions, block pose,~\ldots) in noisy and non-noisy versions. For each network, all observation fields were concatenated into a single vector. Noisy observations were provided to the policy network while the value network had access to non-noisy observations (since the value network is not needed when rolling out the policy on the robot and can thus use privileged information, as described in \citep{pinto2017asymmetric}). We still use the same Asymmetric Actor-Critic architecture~\citep{pinto2017asymmetric} but replace the concatenation with what we call an ``embed-and-add'' approach. More concretely, we first embed each type of observation separately (without any weight sharing) into a latent space of dimensionality $512$. We then combine all inputs by adding the latent representation of each and applying a ReLU non-linearity after.
The main motivation behind this change was to easily add new observations to an existing policy and to share embeddings between value and policy network for inputs that feed into both. The network architecture of our control policy is illustrated in \autoref{fig:policy-arch}. More details of what inputs are fed into the networks can be found in \autoref{app:hyperparameters} and \autoref{table:policy-inputs-full2}. We list the inputs for the block reorientation task in \autoref{app:hyper-neural-net}.

\subsection{Distributed Training with Rapid}
We use our own internal distributed training framework, Rapid. Rapid was previously used to train OpenAI Five~\citep{five} and was also used in~\cite{openai2018learning}.

For the block reorientation task, we use $4 \times 8 = 32$ NVIDIA V100 GPUs and $4 \times 100 = 400$ worker machines with $32$ CPU cores each. For the Rubik's cube task, we use $8\times8=64$ NVIDIA V100 GPUs and $8\times115 = 920$ worker machines with $32$ CPU cores each. We've been training the Rubik's Cube policy continuously for several months  at this scale while concurrently improving the simulation fidelity, ADR algorithm, tuning hyperparameters, and even changing the network architecture. The cumulative amount of experience over that period used for training on the Rubik's cube is roughly $13$~thousand years, which is on the same order of magnitude as the $40$~thousand years used by OpenAI Five~\cite{five}.

The hyperparameters that we used and more details on optimization can be found in \autoref{app:hyperparameters}.

\subsection{Policy Cloning}
With ADR, we found that training the same policy for a very long time is helpful since ADR allows us to always have a challenging training distribution. We therefore rarely trained experiments from scratch but instead updated existing experiments and initialized from previous checkpoints for both the ADR and policy parameters. 
Our new "embed-and-add" approach in \autoref{fig:policy-arch} makes it easier to change the observation space of the agent, but doesn't allow us to experiment with changes to the policy architecture, e.g. modify the number of units in each layer or add a second LSTM layer. 
Restarting training from an uninitialized model would have caused us to lose weeks or months of training progress, making such changes prohibitively expensive. 
Therefore, we successfully implemented behavioral cloning in the spirit of the DAGGER~\citep{ross2010dagger} algorithm (sometimes also called policy distillation~\citep{Czarnecki2019DistillingPD}) to efficiently initialize new policies with a level of performance very close to the teacher policy.

Our setup for cloning closely mimics reinforcement learning, except that we now have both teacher and student policies loaded in memory. During a rollout, we use the student actions to interact with the environment, while minimizing the difference between the student and the teacher's action distributions (by minimizing KL divergence) and value predictions (by minimizing L2 loss). This has worked surprisingly well, allowing us to iterate on the policy architecture quickly without losing the accumulated training progress. Our cloning approach works with arbitrary policy architecture changes as long as the action space remains unchanged.

The best ADR policies used in this work were obtained using this approach. We trained them for multiple months while making multiple changes to the model architecture, training environment, and hyperparameters.

\section{State Estimation from Vision}
\label{sec:vision}
As in ~\cite{openai2018learning}, the control policy described in \autoref{sec:policy} receives object state estimates from a vision system consisting of three cameras and a neural network predictor. In this work, the policy requires estimates for all six face angles in addition to the position and orientation of the cube.

Note that the absolute rotation of each face angle in $[-\pi, \pi]$ radians is required by the policy. Due to the rotational symmetry of the stickers on a standard Rubik's cube, it is not possible to predict these absolute face angles from a single camera frame; the system must either have some ability to track state temporally\footnote{We experimented with a recurrent vision model but found it very difficult to train to the necessary performance level. Due to the project's time constraints, we could not investigate this approach further.} or the cube has to be modified.

\begin{figure}[h]
    \centering
    \begin{subfigure}[b]{0.48\textwidth}
        \centering
        \includegraphics[width=\textwidth]{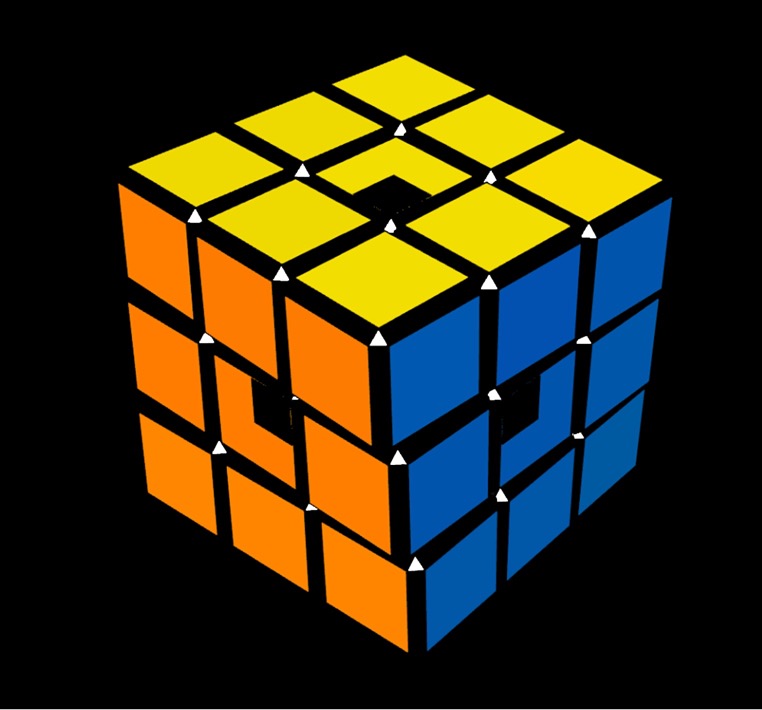}
        \caption{Simulated cube.}
    \end{subfigure}
    \hfill
    \begin{subfigure}[b]{0.48\textwidth}
        \centering
        \includegraphics[width=\textwidth]{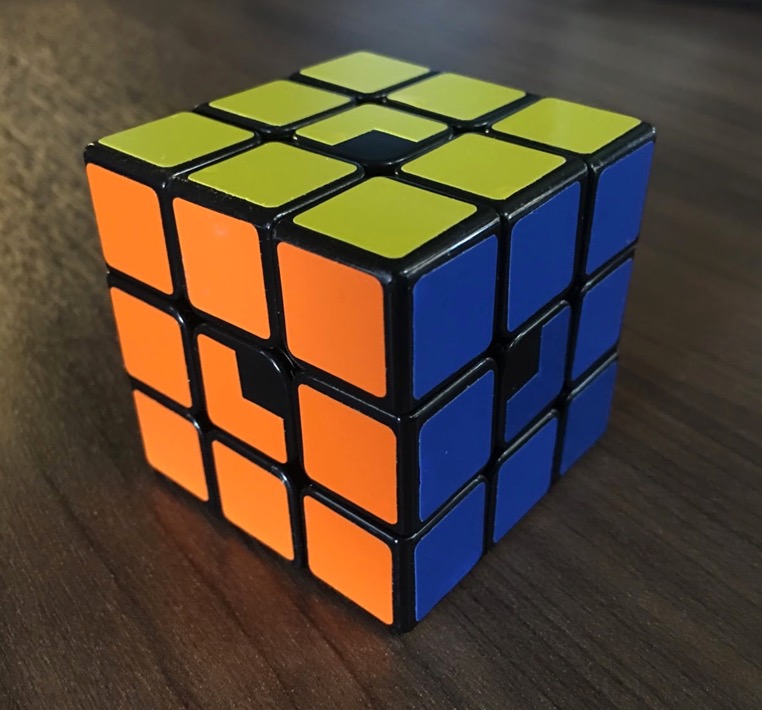}
        \caption{Real cube.}
    \end{subfigure}
    \caption{The Rubik's cube with a corner cut out of each center sticker (a) in simulation and (b) in reality. We used this cube instead of the Giiker cube for some vision state estimation experiments and for evaluating the performance of the policy for solving the Rubik's cube from vision only.}
    \label{fig:cube-cutout-stickers}
\end{figure}

We therefore use two different options for the state estimation of the Rubik's cube throughout this work:
\begin{enumerate}
    \item \textbf{Vision only via asymmetric center stickers.} In this case, the vision model is used to produce the \emph{cube position, rotation, and six face angles}. We cut out one corner of each center sticker on the cube (see \autoref{fig:cube-cutout-stickers}), thus breaking rotational symmetry and allowing our model to determine absolute face angles from a single frame. No further customizations were made to the Rubik's cube. We use this model to estimate final performance of a vision only solution to solving the Rubik's cube.
    \item \textbf{Vision for pose and Giiker cube for face angles.} In this case, the vision model is used to produce the \emph{cube position and rotation}. For the face angles, we use the previously described customized Giiker cube (see \autoref{sec:physical-setup}) with built-in sensors. We use this model for most experiments in order to not compound errors of the challenging face angle estimation from vision only with errors of the policy.
\end{enumerate}

Since our long-term goal is to build robots that can interact in the real world with arbitrary objects, ideally we would like to fully solve this problem from vision alone using a standard Rubik's cube (i.e. without any special stickers). We believe this is possible, though it may require either more extensive work on a recurrent model or moving to an end-to-end training setup (i.e. where the vision model is learned jointly with the policy). This remains an active area of research for us.

\subsection{Vision Model}

\begin{figure}[h]
    \centering
    \includegraphics[width=0.7\textwidth]{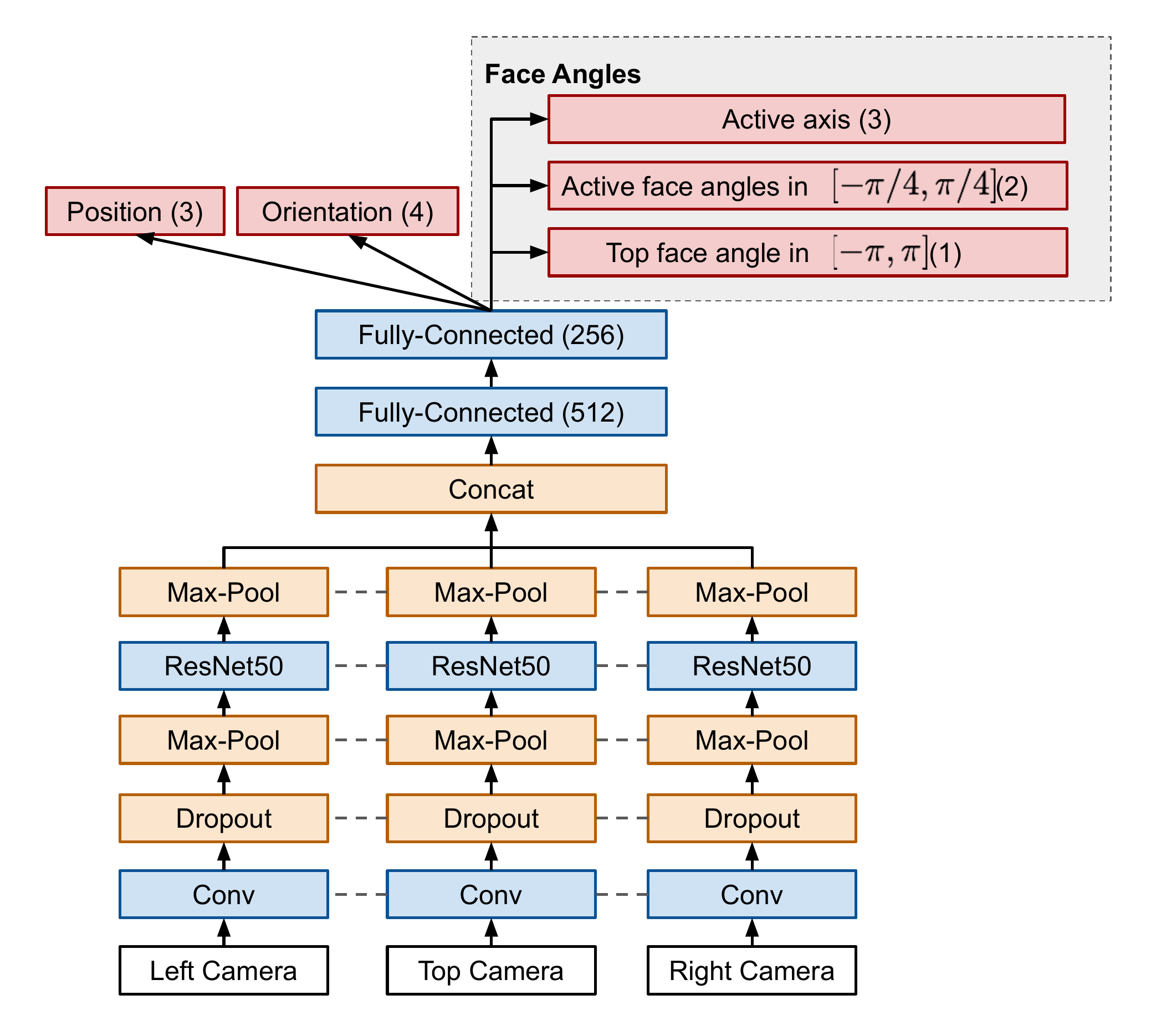}
    \caption{Vision model architecture, which is largely built upon a ResNet50~\cite{He2016DeepRL} backbone. Network weights are shared across the three camera frames, as indicated by the dashed line. Our model produces the position, orientation, and a specific representation of the six face angles of the Rubik's cube. We specify ranges with $[\ldots]$ and dimensionality with $(\ldots)$.}
    \label{fig:vision-arch}
\end{figure}

Our vision model has a similar setup as in~\cite{openai2018learning}, taking as input an image from each of three RGB Basler cameras located at the left, right, and top of the cage (see \autoref{fig:setup-overview}(a)). The full model architecture is illustrated in \autoref{fig:vision-arch}. We produce a feature map for each image by processing it through identically parameterized ResNet50~\cite{He2016DeepRL} networks (i.e. using common weights). These three feature maps are then flattened, concatenated, and fed into a stack of fully-connected layers which ultimately produce predictions sufficient for tracking the full state of the cube, including the position, orientation, and face angles. 

While predicting position and orientation directly works well, we found predicting all six face angles directly to be much more challenging due to heavy occlusion, even when using a cube with asymmetric center stickers. To work around this, we decomposed face angle prediction into several distinct predictions:
\begin{enumerate}
    \item \textbf{Active axis:} We make a slight simplifying assumption that only one of the three axes of a cube can be "active" (i.e. be in an non-aligned state), and have the model predict which of the three axes is currently active.
    \item \textbf{Active face angles:} We predict the angles of the two faces relevant for the active axis \textit{modulo $\pi/2$ radians} (i.e. in $[-\pi/4, \pi/4]$). It is hard to predict the absolute angles in $[-\pi, \pi]$ radians directly due to heavy occlusion (e.g. when a face is on the bottom and hidden by the palm). Predicting these modulo $\pi/2$ angles only requires recognizing the shape and the relative positions of cube edges, and therefore it is an easier task. 
    \item \textbf{Top face angle:} The last piece to predict is the absolute angle in $[-\pi, \pi]$ radians of the "top" face, that is the face visible from a camera mounted directly above the hand. Note that this angle is only possible to predict from frames at a single timestamp because of the asymmetric center stickers (See \autoref{fig:cube-cutout-stickers}). We configure the model to make a prediction only for the top face because the top face's center cubelet is rarely occluded. This gives us a stateless estimate of each face's absolute angle of rotation whenever that face is placed on top.
\end{enumerate}

These decomposed face angle predictions are then fed into post-processing logic (See \autoref{app:hyper} Algorithm~\ref{algorithm:vision_post_processing}) to track the rotation of all face angles, which are in turn passed along to the policy. 
The top face angle prediction is especially important, as it allows us to correct the tracked absolute face angle state mid-trial. For example, if the tracking of a face angle becomes off by some number of rotations (i.e. a multiple of $\pi/2$ radians), we are still able to correct it with a stateless absolute angle prediction from the model whenever this face is placed on top after a flip.
Predictions (1) and (2) are primarily important because the policy is unable to rotate a non-active face if the active face angles are too large (in which case the cube becomes interlocked along non-active axes). 

For all angle predictions, we found that discretizing angles into 90 bins per $\pi$ radians yielded better performance than directly predicting angles via regression; see \autoref{table:vision-ablations} for details.

In the meantime, domain randomization in the rendering process remains a critical role in the sim2real transfer. As shown in \autoref{table:vision-ablations}, a model trained without domain randomization can achieve perfectly low errors in simulation but fails dramatically on real world data.

\begin{center}
\begin{table}[h]
    \caption{Ablation experiments for the vision model. For each experiment, we ran training with 3 different seeds and report the best performance here. Orientation error is computed as rotational distance over a quaternion representation. Position error is the euclidean distance in 3D space, in millimeters. Face angle error is measured in degrees ($^{\circ}$). "Real" errors are computed using data collected over multiple physical trials, where the position and orientation ground truths are from PhaseSpace (\autoref{sec:physical-setup}) and all face angle ground truths are from the Giiker cube. The full evaluation results, including errors on active axis and active face angles, are reported in \autoref{app:full-results}~\autoref{table:vision-ablations-full}.}
    \label{table:vision-ablations}
    \centering
    \renewcommand{\arraystretch}{1.3}
    \begin{tabular}{l c c c | c c c}
        \toprule
        \multirow{2}{*}{\textbf{Experiment}} 
        & \multicolumn{3}{c}{\textbf{Errors (Sim)}} 
        & \multicolumn{3}{c}{\textbf{Errors (Real)}} \\
        & \textbf{Orientation} & \textbf{Position} & \textbf{Top Face} & \textbf{Orientation} & \textbf{Position} & \textbf{Top face} \\
        \midrule
        Full Model & $6.52^\circ$ & $\mathbf{2.63}$ mm & $11.95^\circ$ & $\mathbf{7.81^\circ}$ & $\mathbf{6.47}$ \textbf{mm} & $\mathbf{15.92^\circ}$ \\
        \midrule
        No Domain Randomization & $\mathbf{3.95^\circ}$ & $2.97$ mm & $\mathbf{8.56^\circ}$ & $128.83^\circ$ & $69.40$ mm & $85.33^\circ$  \\
        No Focal Loss & $15.94^\circ$ & $5.02$ mm & $10.17^\circ$ & $19.10^\circ$ & $9.416$ mm & $17.54^\circ$ \\
        Non-discrete Angles & $9.02^\circ$ & $3.78$ mm & $42.46^\circ$ & $10.40^\circ$ & $7.97$ mm & $35.27^\circ$  \\
        \bottomrule
    \end{tabular}
\end{table}
\end{center}

\subsection{Distributed Training with Rapid}

As in control policy training (\autoref{sec:policy}), the vision model is trained entirely from synthetic data, without any images from the real world. This necessarily entails a more complicated training setup, wherein the synthetic image generation must be coupled with optimization. To manage this complexity, we leverage the same Rapid framework~\citep{five} which is used in policy training for distributed training.

\autoref{fig:adr-architecture-vision} gives an overview of the setup for a typical vision experiment. In the case of vision training, the ``data workers'' are standalone Unity renderers, responsible for rendering simulated images using OpenAI Remote Rendering Backend (ORRB)~\cite{chociej2019orrb}. These images are rendered according to ADR parameters pulled from the ADR subsystem (see \autoref{sec:adr}). A list of randomization parameters is available in \autoref{app:visual_randomizations} \autoref{table:vision_randomizations}. Each rendering node uses 1 NVIDIA V100 GPU and 8 CPU cores, and the size of the rendering pool is tuned such that rendering is not a bottleneck in training. The data from these rendering nodes is then propagated to a cluster of Redis nodes where it is stored in separate queues for training and evaluation. The training data is then read by a pool of optimizer nodes, each of which uses 8 NVIDIA V100 GPUs and 64 CPU cores, in order to perform optimization in a data-parallel fashion. Meanwhile, the evaluation data is read by the ``ADR eval workers'' in order to provide feedback on ADR parameters, per \autoref{sec:adr}.

As noted above, the vision model produces several distinct predictions, each of which has its own loss function to be optimized: mean squared error for both position and orientation, and cross entropy for each of the decomposed face angle predictions. To balance these many losses, which lie on different scales, we use focal loss weighting as described in~\cite{guo2018dynamic} to dynamically and automatically assign loss weights. One modification we made in order to better fit our multiple regression tasks is that we define a low target error for each prediction and then use the percentage of samples that obtain errors below the target as the probability $p$ in the focal loss, i.e. $\text{FL}(p; \gamma) = - (1-p)^\gamma \log(p)$, where $\gamma=1$ in all our experiments. This both removes the need to manually tune loss weights and improves optimization, as it allows loss weights to change dynamically during training (see \autoref{table:vision-ablations} for performance details). 

Optimization is then performed against this aggregate loss using the LARS optimizer~\cite{LARS}. We found LARS to be more stable than the Adam optimizer~\cite{Kingma2014AdamAM} when using larger batches and higher learning rates (we use at most a batch size of $1024$ with a peak learning rate of $6.0$). See \autoref{app:hyperparameters} for further hyperparameter details.

\section{Results}
\label{sec:exp-adr}
In this section, we investigate the effect ADR has on transfer (\autoref{sec:result-adr-transfer}), empirically show the importance of having a curriculum for policy training (\autoref{sec:result-adr-curr}), quantify vision performance (\autoref{sec:result-vision}), and finally present our results that push the limits of what is possible by solving a Rubik's cube on the real Shadow hand (\autoref{sec:result-rubik}).

\subsection{Effect of ADR on Policy Transfer}
\label{sec:result-adr-transfer}

To understand the transfer performance impact of training policies with ADR, we study the problem on the simpler block reorientation task previously introduced in~\cite{openai2018learning}. We use this task since it is computationally more tractable and because baseline performance has been established. As in~\cite{openai2018learning}, we measure performance in terms of the number of consecutive successes. We terminate an episode if the block is either dropped or if $50$ consecutive successes are achieved. An optimal policy would therefore be one that achieves a mean of $50$ successes.

\subsubsection{Sim2Sim}

\begin{figure}[h]
    \centering
    \begin{subfigure}[b]{0.48\textwidth}
        \includegraphics[width=\textwidth]{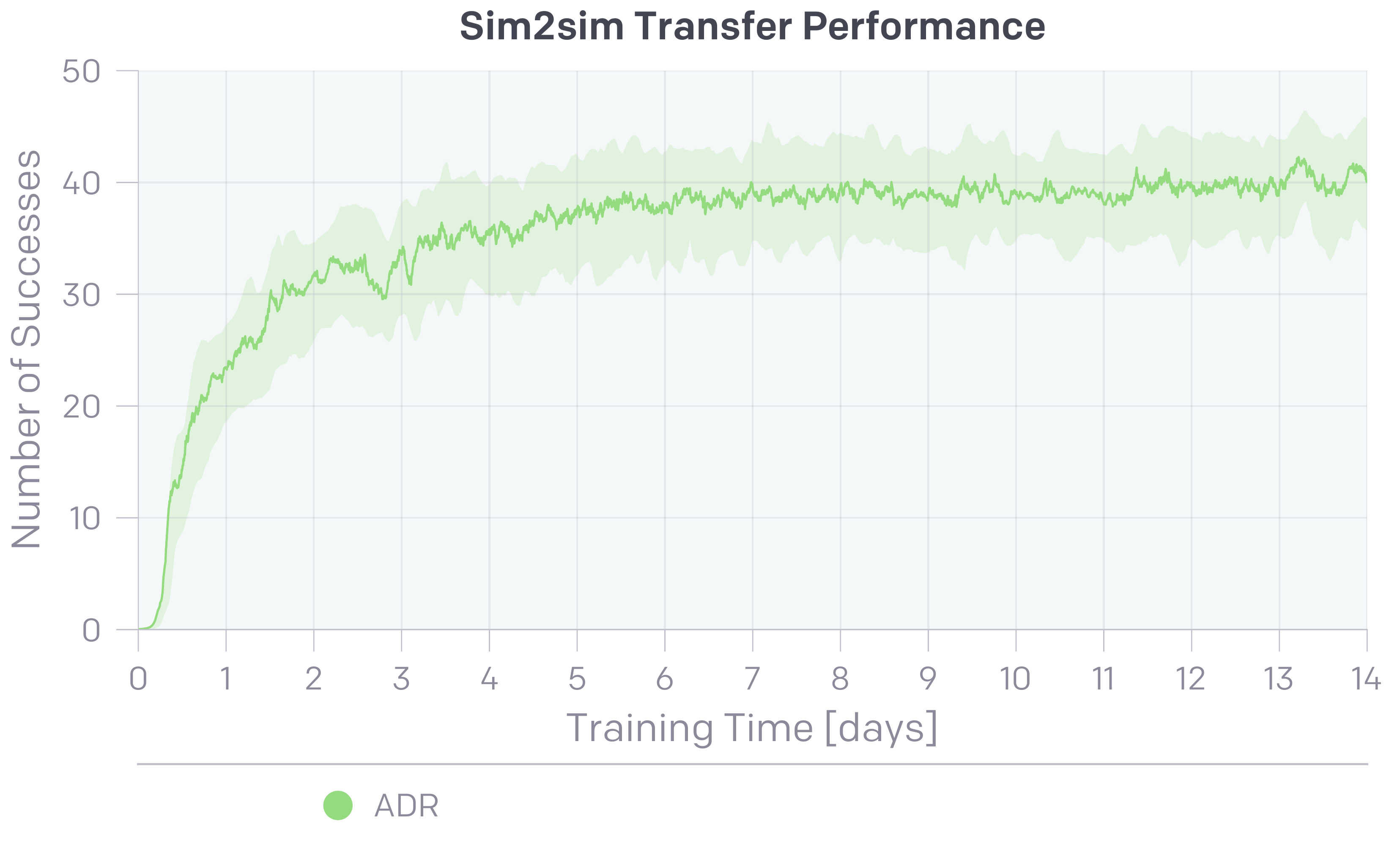}
        \label{fig:adr-sim2sim-perf}
    \end{subfigure}
    \hfill
    \begin{subfigure}[b]{0.48\textwidth}
        \includegraphics[width=\textwidth]{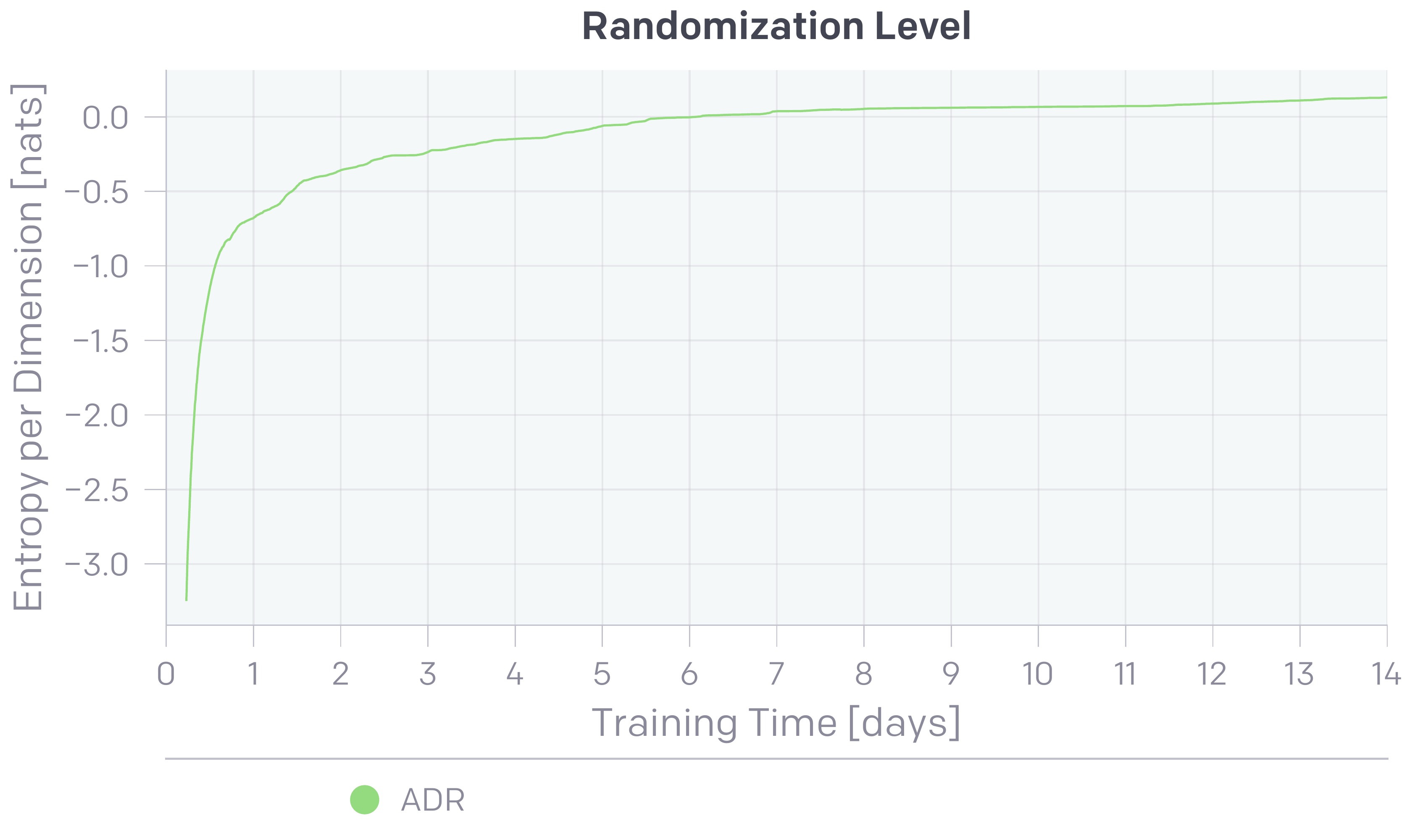}
        \label{fig:adr-sim2sim-volume}
    \end{subfigure}
    \caption{
    Sim2sim performance (left) and ADR entropy (right) over the course of training. We can see that a policy trained with ADR has better sim2sim transfer as ADR increases the randomization level over time.}
    \label{fig:adr-sim2sim}
\end{figure}

We first consider the sim2sim case. More concretely, we train a policy with ADR and continuously benchmark its performance on an distribution of environments with manually tuned randomizations, very similar to the ones we used in~\cite{openai2018learning}. Note that no ADR experiment has ever been trained on this distribution directly. Instead we use ADR to decide what distribution to train on, making the manually designed distribution over environments a test set for sim2sim transfer. We report our results in \autoref{fig:adr-sim2sim}.

As seen in \autoref{fig:adr-sim2sim}, the policy trained with ADR transfers to the manually randomized distribution. Furthermore, the sim2sim transfer performance increases as ADR increases the randomization entropy.

\subsubsection{Sim2Real}
Next, we evaluate the sim2real transfer capabilities of our policies. Since rollouts on the robot are expensive, we limit ourselves to 7 different policies that we evaluate. For each of them, we collect a total of $10$ trials on the robot and measure the number of consecutive successes. As before, a trial ends when we either achieve $50$ successes, the robot times out or the block is dropped. For each policy we deploy, we also report simulation performance by measuring the number of successes across $500$ trials each for reference. As before, we use the manually designed randomizations as described in~\cite{openai2018learning} for sim evaluations. We summarize our results in \autoref{table:adr-xyz-transfer} and report detailed results in \autoref{app:full-results}.

\begin{table}[h]
    \caption{Performance of different policies on the block reorientation task. We evaluate each policy in simulation (N=500 trials) and on the real robot (N=10 trials) and report the mean $\pm$ standard error and median number of successes. For ADR policies, we report the entropy in nats per dimension (npd). For ``Manual DR'', we obtain an upper bound on its ADR entropy by running ADR with the policy fixed and report the entropy once the distribution stops changing (marked with an ``*'').}
    \label{table:adr-xyz-transfer}
    \centering
    \renewcommand{\arraystretch}{1.3}
        \begin{tabular}{@{}lrr|rr|rr@{}}
            \toprule
            \multirow{2}{*}{\textbf{Policy}} & \multirow{2}{*}{\textbf{Training Time}} & \multirow{2}{*}{\textbf{ADR Entropy}} & \multicolumn{2}{c}{\textbf{Successes (Sim)}} & \multicolumn{2}{c}{\textbf{Successes (Real)}} \\
            & & & \textbf{Mean} & \textbf{Median} & \textbf{Mean} & \textbf{Median} \\
            \midrule
            Baseline (data from \cite{openai2018learning}) & --- & --- & $43.4 \pm 0.6$ & $\mathbf{50}$ & $18.8 \pm 5.4$ & $13.0$ \\
            Baseline (re-run of \cite{openai2018learning}) & --- & --- & $33.8 \pm 0.9$ & $\mathbf{50}$ & $4.0 \pm 1.7$ & $2.0$ \\
            \midrule
            Manual DR & $13.78$~days & $-0.348$\textsuperscript{*}~npd& $42.5 \pm 0.7$ & $\mathbf{50}$ & $2.7 \pm 1.1$ & $1.0$ \\
            \midrule
            ADR (Small) & $0.64$~days & $-0.881$~npd & $21.0 \pm 0.8$ & $15$ & $1.4 \pm 0.9$ & $0.5$ \\
            ADR (Medium) & $4.37$~days & $-0.135$~npd & $34.4 \pm 0.9$ & $\mathbf{50}$ & $3.2 \pm 1.2$ & $2.0$ \\
            ADR (Large) & $13.76$~days & $0.126$~npd & $40.5 \pm 0.7$ & $\mathbf{50}$ & $13.3 \pm 3.6$ & $11.5$ \\
            \midrule
            ADR (XL) & --- & $0.305$~npd & $45.0 \pm 0.6$ & $\mathbf{50}$ & $16.0 \pm 4.0$ & $12.5$ \\
            ADR (XXL) & --- & $\mathbf{0.393}$~\textbf{npd} & $\mathbf{46.7 \pm 0.5}$ & $\mathbf{50}$ & $\mathbf{32.0 \pm 6.4}$ & $\mathbf{42.0}$ \\
            \bottomrule
        \end{tabular}
\end{table}

The first two rows connect our results in this work to the previous results reported in~\cite{openai2018learning}. For convenience, we repeat the numbers reported in~\cite{openai2018learning} in the first row. We also re-deploy the exact same policy we used back then on our setup today. We find that the same policy performs much worse today, presumably because both our physical setup and simulation have changed since (as described in \autoref{sec:physical-setup} and \autoref{sec:sim}, respectively).

The next section of the table compares a policy trained with ADR and a policy trained with manual domain randomization (denoted as ``Manual DR''). Note that ``Manual DR'' uses the same randomizations as the baseline from~\cite{openai2018learning} but is trained on our current setup with the same model architecture and hyperparameters as the ADR policy. For the ADR policy, we select snapshots at different points during training at varying levels of entropy and denote them as small, medium, and large.\footnote{Note that this is one experiment, not multiple different experiments, taken at different points in time during training.} We can clearly see a pattern: increased ADR entropy corresponds to increased sim2sim and sim2real transfer. The policy trained with manual domain randomization achieves high performance in simulation. However, when deployed on the robot, it fails. This is because, in contrast to our results obtained in~\cite{openai2018learning}, we did not tune our simulation and randomization setup by hand to match changes in hardware. Our ADR policies transfer because ADR automates this process and results in training distributions that are vastly broader than our manually tuned distribution was in the past. Also note that ``ADR (Large)'' and ``Manual DR'' were trained for the same amount of wall-clock time and share all training hyperparameters except for the environment distribution, i.e. they are fully comparable. Due to compute constraints, we train those policies at $\nicefrac{1}{4}$th of our usual scale in terms of compute (compare \autoref{sec:policy}).

The last block of the table lists results that we obtained when scaling ADR up. We report results for ``ADR (XL)'' and ``ADR (XXL)'', referring to two long-running experiments that were continuously trained for extended periods of time and at larger scale. We can see that they exhibit the best sim2sim and sim2real transfer and that, again, an increase in ADR entropy corresponds to vastly improved sim2real transfer. Our best result significantly beat the baseline reported in~\cite{openai2018learning} even though we did not tune the simulation and robot setup for peak performance on the block reorientation task: we increase mean performance by almost $2\times$ and median performance by more than $3\times$. As a side note, we also see that policies trained with ADR eventually achieve near-perfect performance for sim2sim transfer as well.

In summary, ADR clearly leads to improved transfer with much less need for hand-engineered randomizations. We significantly outperformed our previous best results, which were the result of multiple months of iterative manual tuning.

\subsection{Effect of Curriculum on Policy Training}
\label{sec:result-adr-curr}
We designed ADR to expand the complexity of the training distribution gradually. This makes intuitive sense: start with a single environment and then grow the distribution over environments as the agent progresses. The resulting curriculum should make it possible to eventually master a highly diverse set of environments. However, it is not clear if this curriculum property is important or if we can train with a fixed set of domain randomization parameters once they have been found.

\begin{figure}[h]
    \centering
    \begin{subfigure}[b]{0.48\textwidth}
        \includegraphics[width=\textwidth]{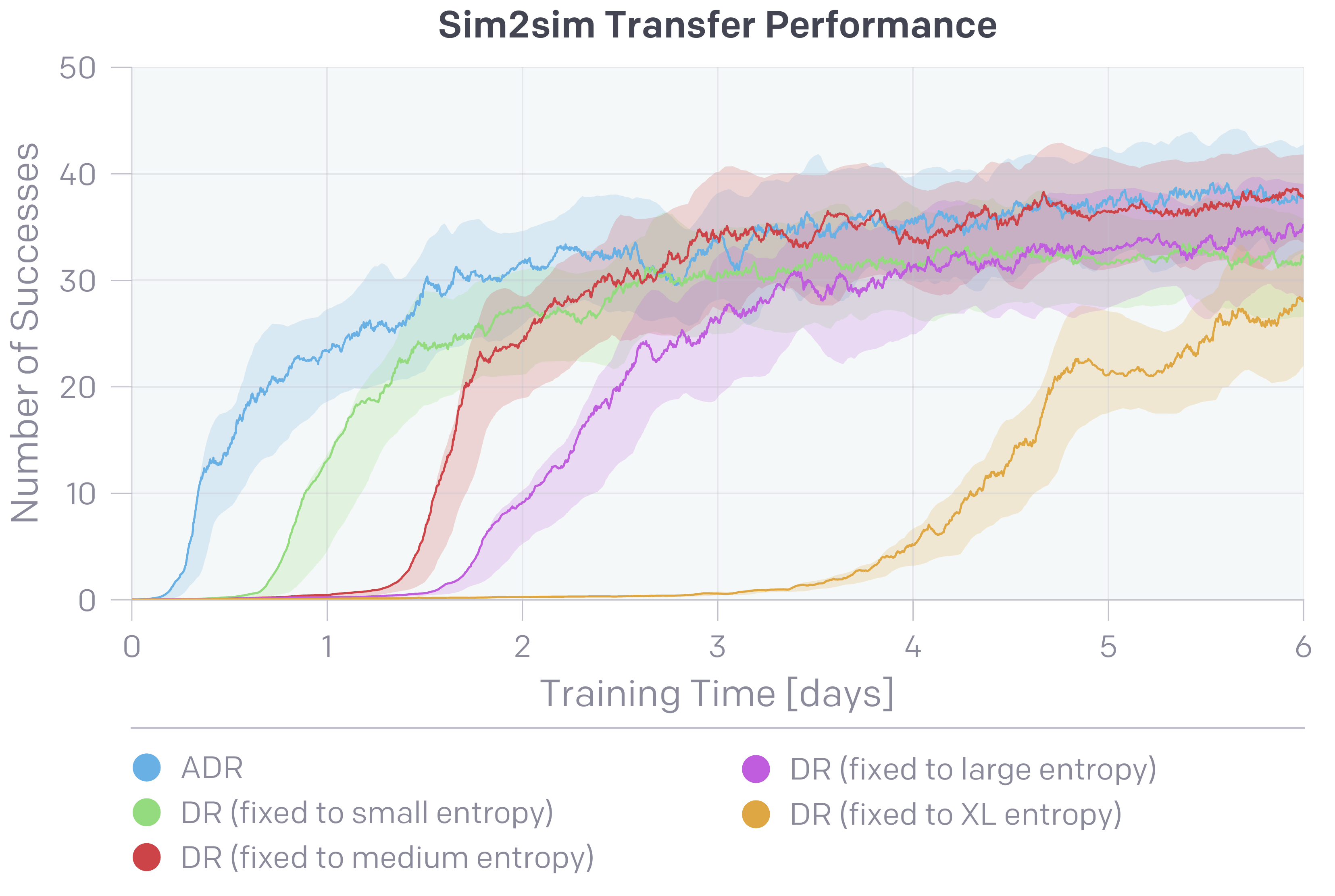}
    \end{subfigure}
    \hfill
    \begin{subfigure}[b]{0.48\textwidth}
        \includegraphics[width=\textwidth]{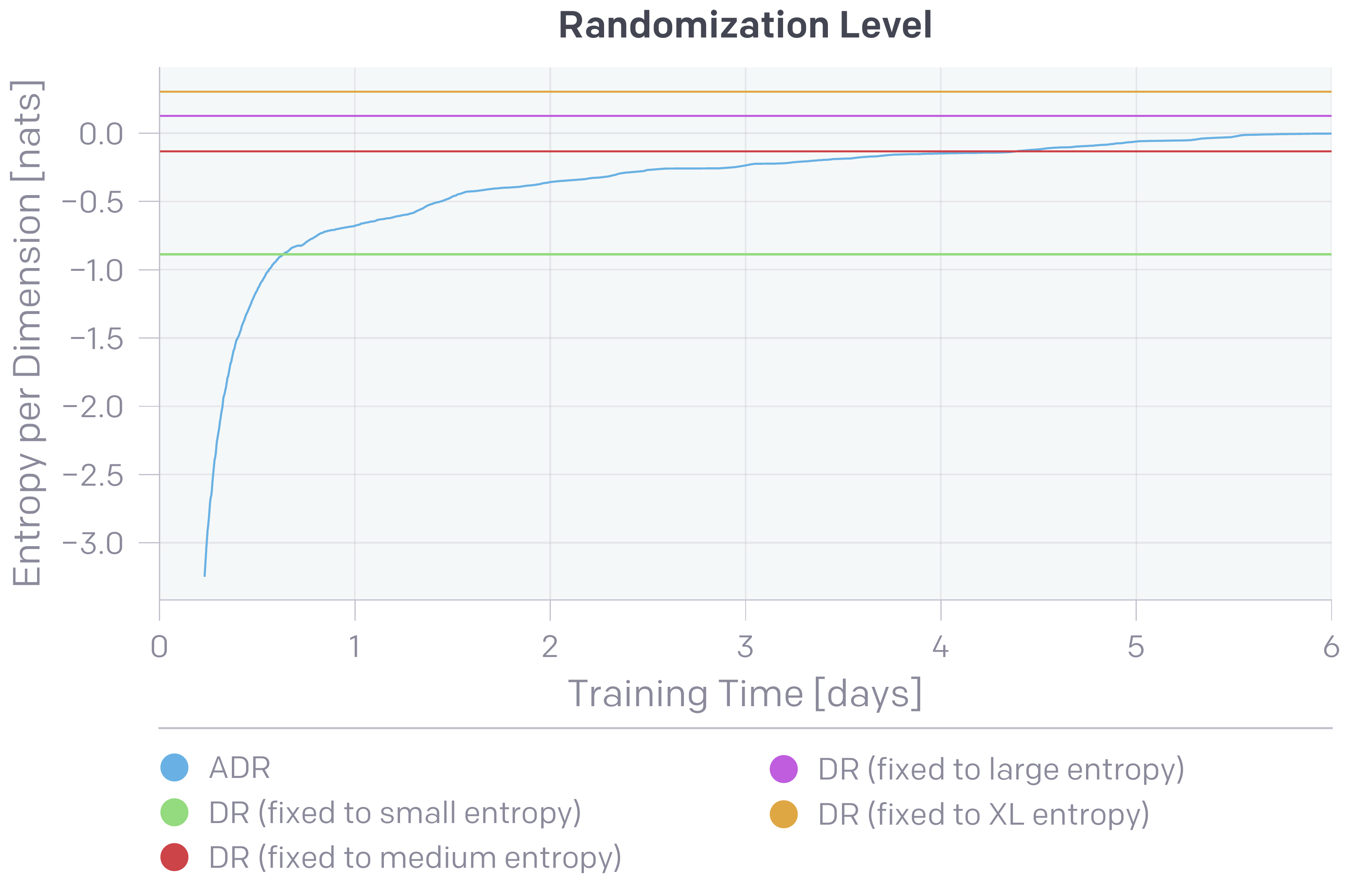}
    \end{subfigure}
    \caption{Sim2sim performance (left) and ADR entropy (right) over the course of training. \emph{ADR} refers to a regular training run, i.e. we start with zero randomization and let ADR gradually expand the randomization level. We compare ADR against runs with domain randomization (DR) fixed at different levels and train policies on each of those environment distributions. We can see that ADR makes progress much faster due to its curriculum property.}
    \label{fig:adr-curr}
\end{figure}

To test for this, we conduct the following experiment. We train one policy with ADR on the block reorientation task and compare it against multiple policies with different fixed randomizations. We use 4 different fixed levels: small, medium, large, and XL. They correspond to the ADR parameters of the policies from the previous section (compare \autoref{table:adr-xyz-transfer}). However, note that we only use the ADR parameters, \emph{not} the policies from \autoref{table:adr-xyz-transfer}. Instead, we train new policies from scratch using these parameters and train all of them for the same amount of wall-clock time. We evaluate performance of all policies continuously on the same manually randomized distribution from~\cite{openai2018learning}, i.e. we test for sim2sim transfer in all cases. We depict our results in \autoref{fig:adr-curr}. Note that for all DR runs the randomization entropy is constant; only the one for ADR gradually increases.

Our results in \autoref{fig:adr-curr} clearly demonstrate that adaptively increasing the randomization entropy is important: the ADR run achieves high sim2sim transfer much more quickly than all other runs with fixed randomization entropy.\footnotetext{The attentive reader will notice that the sim2sim performance reported in \autoref{fig:adr-curr} is different from the sim2sim performance reported in \autoref{table:adr-xyz-transfer}. This is because here we train the policies for \emph{longer} but on a \emph{fixed} ADR entropy whereas in \autoref{table:adr-xyz-transfer} we had a single ADR run and took snapshots at different points over the course of training.} There is also a clear pattern: the larger the fixed randomization entropy, the longer it takes to train from scratch. We hypothesize that for a sufficiently difficult task and randomization entropy, training from scratch becomes infeasible altogether. More concretely, we believe that for too complex environments the policy would never learn due to the task being so hard that there is no sufficient reinforcement learning signal.

\subsection{Effect of ADR on Vision Model Performance}
\label{sec:result-vision}

When training vision models, ADR controls both the ranges of randomization in ORRB (i.e. light distance, material metallic and glossiness) and TensorFlow distortion operations (i.e. adding Gaussian noise and channel noise). A full list of vision ADR randomization parameters are available in \autoref{app:visual_randomizations}~\autoref{table:vision_randomizations}. We train ADR-enhanced vision models to do state estimation for both the block reorientation~\cite{openai2018learning} and Rubik's cube task. 
As shown in \autoref{table:vision_adr_locked}, we are able to reduce the prediction errors on both block orientation and position further than our manual domain randomization results in the previous work~\cite{openai2018learning}.\footnote{Note that this model has the same manual DR as in~\cite{openai2018learning} but is evaluated on a newly collected real image set, so the numbers are slightly different from~\cite{openai2018learning}.} We can also see that increased ADR entropy again corresponds to better sim2real transfer.

\begin{table}[h]
    \caption{Performance of vision models at different ADR entropy levels for the block reorientation state estimation task. Note that the baseline model here uses the same manual domain randomization configuration as in~\cite{openai2018learning} but is evaluated on a newly collected real image dataset (the same real dataset described in \autoref{table:vision-ablations})}
    \centering
    \renewcommand{\arraystretch}{1.3}
    \begin{tabular}{@{}lrr|rr|rr@{}}
        \toprule
        \multirow{2}{*}{\textbf{Model}} & \multirow{2}{*}{\textbf{Training Time}} & \multirow{2}{*}{\textbf{ADR Entropy}} & \multicolumn{2}{c}{\textbf{Errors (Sim)}} & \multicolumn{2}{c}{\textbf{Errors (Real)}} \\
        & & & \textbf{Orientation} & \textbf{Position} & \textbf{Orientation} & \textbf{Position} \\
        \midrule
        Manual DR &  13.62 hrs & --- & $1.99^\circ$ & $4.03$ mm & $5.19^\circ$ & $8.53$ mm \\
        \midrule
        ADR (Small) & 2.5 hrs & $0.922$ npd & $2.81^\circ$ & $4.21$ mm & $6.99^\circ$ & $8.13$ mm \\
        ADR (Middle) & 3.87 hrs & $1.151$ npd & $2.73^\circ$ & $4.11$ mm & $6.66^\circ$ & $8.14$ mm \\
        ADR (Large) & 12.76 hrs & $\mathbf{1.420}$ npd & $1.85^\circ$ & $5.18$ mm & $\mathbf{5.09}^\circ$ & $\mathbf{7.85}$ mm \\
        \bottomrule
    \end{tabular}
    \label{table:vision_adr_locked}
\end{table}

Predicting the full state of a Rubik's cube is a more difficult task and demands longer training time. A vision model using ADR succeeds to achieve lower errors than the baseline model with manual DR configuration given a similar amount of training time, as demonstrated in \autoref{table:vision_adr_full}. Higher ADR entropy well correlates with lower errors on real images. ADR again outperforms our manually tuned randomizations (i.e. the baseline). Note that errors in simulation increase as ADR generates harder and harder synthetic tasks.

\begin{table}[h]
    \caption{Performance of vision models at different ADR entropy levels for the Rubik's cube prediction task (See \autoref{sec:vision}). The real image datasets used for evaluation are same as in \autoref{table:vision-ablations}. The full evaluation results are reported in \autoref{app:full-results}~\autoref{table:vision-ablations-full}.}
    \centering
    \renewcommand{\arraystretch}{1.3}
        \begin{tabular}{@{}lrr|rrr|rrr@{}}
            \toprule
            \multirow{2}{*}{\textbf{Model}} & \textbf{Training} & \textbf{ADR} & \multicolumn{3}{c}{\textbf{Errors (Sim)}} & \multicolumn{3}{c}{\textbf{Errors (Real)}} \\
            & \textbf{Time} & \textbf{Entropy} 
            & \textbf{Orientation} & \textbf{Position} & \textbf{Top angle} 
            & \textbf{Orientation} & \textbf{Position} & \textbf{Top angle} \\
            \midrule
            Baseline 
            & $76.29$ hrs & --- & $6.52^\circ$ & $2.63$ mm & $11.95^\circ$ 
            & $7.81^\circ$ & $6.47$ mm & $15.92^\circ$ \\
            \midrule
            ADR (Small) 
            & $20.9$ hrs & $-0.565$ npd & $5.02^\circ$ & $3.36$ mm & $9.34^\circ$ 
            & $8.93^\circ$ & $7.61$ mm & $16.57^\circ$ \\
            ADR (Middle) 
            & $30.6$ hrs & $0.511$ npd & $15.68^\circ$ & $3.02$ mm & $20.29^\circ$ 
            & $8.44^\circ$ & $7.30$ mm & $15.81^\circ$\\
            ADR (Large) 
            & $75.1$ hrs & $\mathbf{0.806}$ npd & $15.76^\circ$ & $3.58$ mm & $20.78^\circ$ 
            & $\mathbf{7.48}^\circ$ & $\mathbf{6.24}$ mm & $\mathbf{13.83}^\circ$ \\
            \bottomrule
        \end{tabular}
    \label{table:vision_adr_full}
\end{table}

\subsection{Solving the Rubik's Cube}
\label{sec:result-rubik}

In this section, we push the limits of sim2real transfer by considering a manipulation problem of unprecedented complexity: solving Rubik's cube using the real Shadow hand. This is a daunting task due to the complexity of Rubik's cube and the interactions between it and the hand: in contrast to the block reorientation task, there is no way we can accurately capture the object in simulation. While we model the Rubik's cube (see \autoref{sec:sim}), we make no effort to calibrate its dynamics. Instead, we use ADR to automate the randomization of environments.

We further need to sense the state of the Rubik's cube, which is also much more complicated than for the block reorientation task. We always use vision for the pose estimation of the cube itself. For the $6$~face angles, we experiment with two different setups: the Giiker cube (see \autoref{sec:physical-setup}) and a vision model which predicts face angles (see \autoref{sec:vision}).

We first evaluate performance on this task quantitatively and then highlight some qualitative findings.

\subsubsection{Quantitative Results}
We compare four different policies: a policy trained with manual domain randomization (``Manual DR'') using the randomizations that we used in~\cite{openai2018learning} trained for about 2 weeks, a policy trained with ADR for about 2 weeks, and two policies we continuously trained and updated with ADR over the course of months.

\begin{table}[h]
    \caption{Performance of different policies on the Rubik's cube for a fixed fair scramble goal sequence. We evaluate each policy on the real robot (N=10 trials) and report the mean $\pm$ standard error and median number of successes (meaning the total number of successful rotations and flips). We also report two success rates for applying half of a fair scramble (``half'') and the other one for fully applying it (``full''). For ADR policies, we report the entropy in nats per dimension (npd). For ``Manual DR'', we obtain an upper bound on its ADR entropy by running ADR with the policy fixed and report the entropy once the distribution stops changing (marked with an ``*'').}
    \label{table:adr-full-transfer}
    \centering
    \renewcommand{\arraystretch}{1.3}
    \begin{tabular}{@{}lllr|rrrr@{}}
        \toprule
        \multirow{2}{*}{\textbf{Policy}} & \multicolumn{2}{c}{\textbf{Sensing}} & \multirow{2}{*}{\textbf{ADR Entropy}} & \multicolumn{2}{c}{\textbf{Successes (Real)}} & \multicolumn{2}{c}{\textbf{Success Rate}} \\
        & \textbf{Pose} & \textbf{Face Angles} & & \textbf{Mean} & \textbf{Median} & \textbf{Half} & \textbf{Full} \\
        \midrule
        Manual DR & Vision & Giiker & $-0.569$\textsuperscript{*}~npd & $1.8 \pm 0.4$ & $2.0$ & $0$ \% & $0$ \% \\
        ADR & Vision & Giiker & $-0.084$~npd & $3.8 \pm 1.0$ & $3.0$ & $0$ \% & $0$ \% \\
        \midrule
        ADR (XL) & Vision & Giiker & $0.467$~npd & $17.8 \pm 4.2$ & $12.5$ & $30$ \% & $10$ \% \\
        ADR (XXL) & Vision & Giiker & $\mathbf{0.479}$~\textbf{npd} & $\mathbf{26.8 \pm 4.9}$ & $\mathbf{22.0}$ & $\mathbf{60}$ \textbf{\%} & $\mathbf{20}$ \textbf{\%} \\
        \midrule
        ADR (XXL) & Vision & Vision & $\mathbf{0.479}$~\textbf{npd} & $12.8 \pm 3.4$ & $10.5$ & $20$ \% & $0$ \% \\
        \bottomrule
    \end{tabular}
\end{table}

To evaluate performance, we define a fixed procedure that we repeat 10 times per policy to obtain 10 trials. More concretely, we always start from a solved cube state and ask the hand to move the Rubik's cube into a fair scramble. Since the problem is symmetric, this is equivalent to solving the Rubik's cube starting from a fairly scrambled Rubik's cube. However, it reduces the probability of human error and labor significantly since ensuring a correct initial state is much simpler if the cube is solved. We use the following fixed scrambling sequence for all 10 trials, which we obtained using the ``TNoodle'' application of the World Cube Association\footnote{\url{https://www.worldcubeassociation.org/regulations/scrambles/}} via a random sample (i.e., this was not cherry-picked):
\begin{quote}
    L2 U2 R2 B D2 B2 D2 L2 F' D' R B F L U' F D' L2
\end{quote}
Completing this sequence requires a total of $43$ successes ($26$ face rotations and $17$ cube flips). If the sequence is completed successfully, we continue the trial by reversing the sequence. A trial ends if $50$~successes have been achieved, if the cube is dropped, or if the policy fails to achieve a goal within $1600$~timesteps, which corresponds to $128$~seconds.

For each trial, we measure the number of successfully achieved goals (both flips and rotations). We also define two thresholds for each trial: Applying at least half of the fair scramble successfully (i.e. $22$~successes) and applying at least the full fair scramble successfully (i.e. $43$~successes). We report success rates for both averaged across all $10$~trials and denote them as ``half'' and ``full'', respectively. Achieving the ``full'' threshold is equivalent to solving the Rubik's cube since going from solved to scrambled is as difficult as going from scrambled to solved.\footnote{With the exception of the fully solved configuration being slightly easier for the vision model to track.} We report our summarized results in \autoref{table:adr-full-transfer} and full results in \autoref{app:full-results}.

We see a very similar pattern as before: manual domain randomization fails to transfer. For policies that we trained with ADR, we see that sim2real transfer clearly depends on the entropy per dimension. ``Manual DR'' and ``ADR'' were trained for $14$~days at $\nicefrac{1}{4}$th of our usual scale in terms of compute (see \autoref{sec:policy}) and are fully comparable. Our best policy, which was continuously trained over the course of multiple months at larger scale, achieves $26.80$ successes on average over 10 trials. This corresponds to successfully solving a Rubik's cube that requires $15$ face rotations $60\%$ of the time and to solve a Rubik's cube that requires $26$ face rotations $20\%$ of the time. Note that $26$~quarter face rotations is the worst case for solving a Rubik's cube with only about $3$~Rubik's cube configurations requiring that many~\cite{qtm}. In other words, almost all solution sequences will require less than $26$ face rotations.

\subsubsection{Qualitative Results}
We observe many interesting emergent behaviors on our robot when using our best policy (``ADR (XXL)'') for solving the Rubik's cube. We encourage the reader to watch the uncut video footage we recorded:\footnotetext{This video solves the Rubik's cube from an initial randomly scrambled state. This is different from the quantitative experiments we conducted in the previous section.}
\mbox{\url{https://youtu.be/kVmp0uGtShk}}.

\begin{figure}[h]
    \centering
    \begin{subfigure}[b]{0.32\textwidth}
        \includegraphics[width=\textwidth]{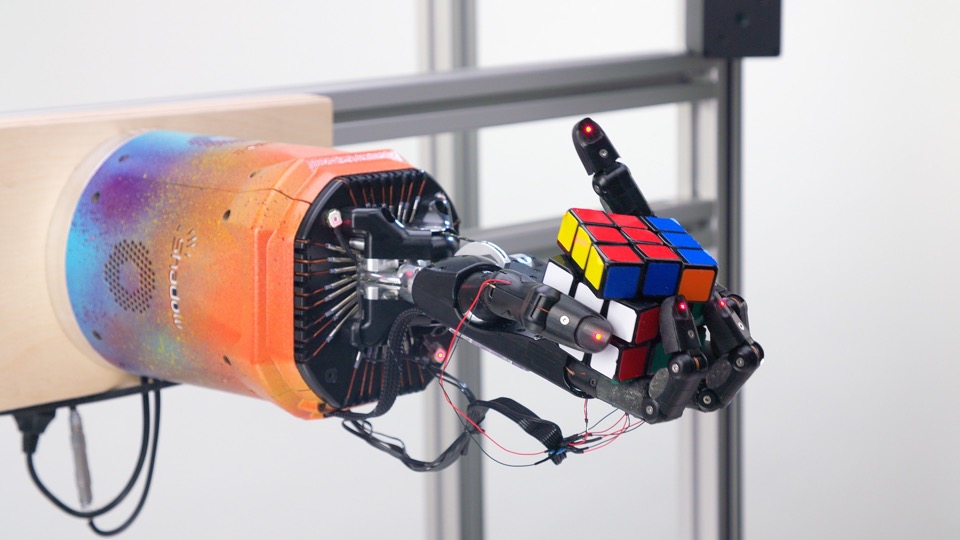}
        \caption{Unperturbed (for reference).}
    \end{subfigure}
    \hfill
    \begin{subfigure}[b]{0.32\textwidth}
        \includegraphics[width=\textwidth]{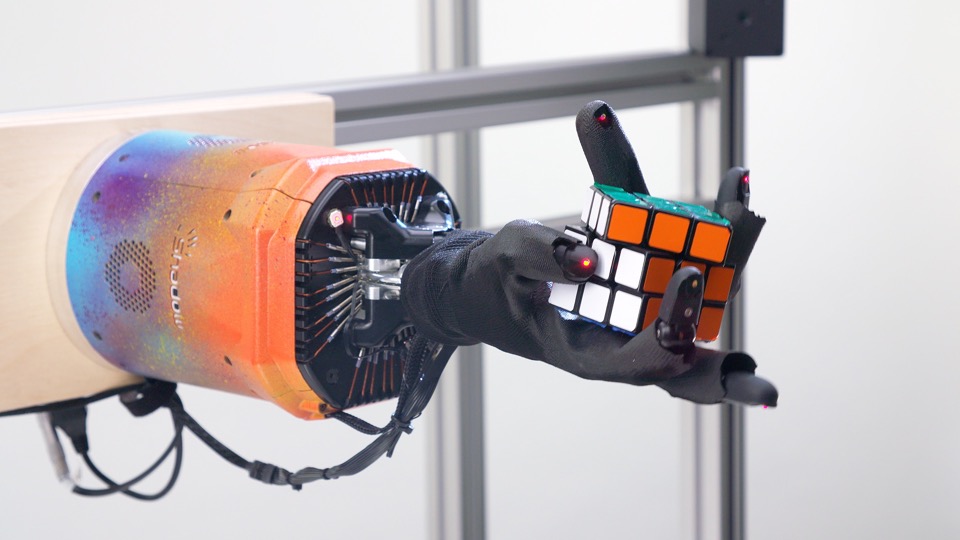}
        \caption{Rubber glove.}
    \end{subfigure}
    \hfill
    \begin{subfigure}[b]{0.32\textwidth}
        \includegraphics[width=\textwidth]{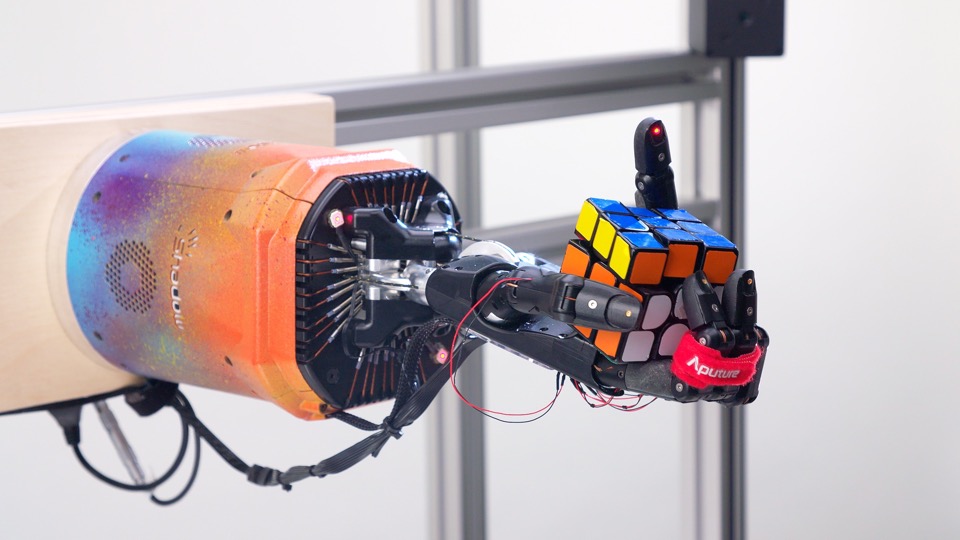}
        \caption{Tied fingers.}
    \end{subfigure}
    \\
    \begin{subfigure}[b]{0.32\textwidth}
        \includegraphics[width=\textwidth]{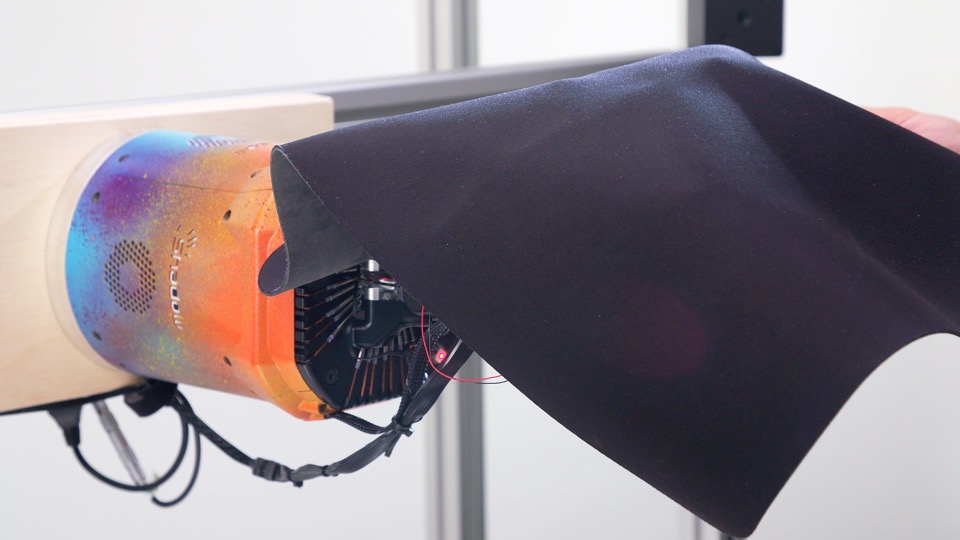}
        \caption{Blanket occlusion and perturbation.}
    \end{subfigure}
    \hfill
    \begin{subfigure}[b]{0.32\textwidth}
        \includegraphics[width=\textwidth]{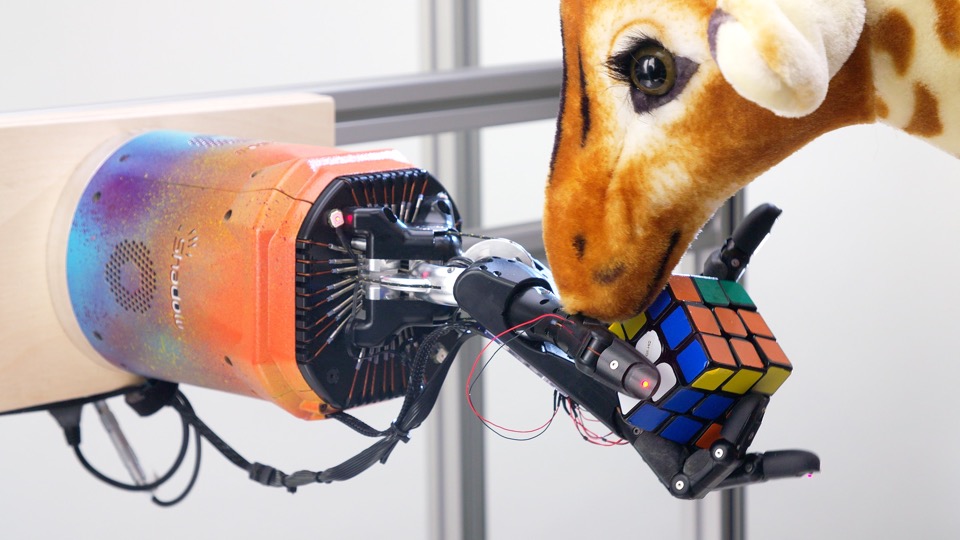}
        \caption{Plush giraffe perturbation.\footnotemark}
    \end{subfigure}
    \hfill
    \begin{subfigure}[b]{0.32\textwidth}
        \includegraphics[width=\textwidth]{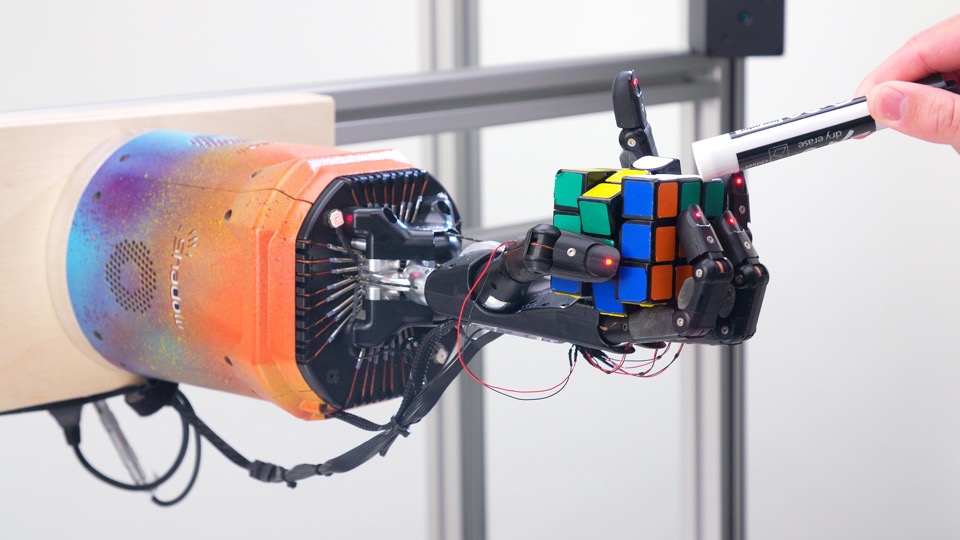}
        \caption{Pen perturbation.}
    \end{subfigure}
    \caption{Example perturbations that we apply to the real robot hand while it solves the Rubik's cube. We did not train the policy to be able to handle those perturbations, yet we observe that it is robust to all of them. A video of is available: \mbox{\url{https://youtu.be/QyJGXc9WeNo}}}
    \label{fig:cube-perturbations}
\end{figure}

For example, we observe that the robot sometimes accidentally rotates an incorrect face. If it does so, our best policies are usually able to recover from this mistake by first rotating the face back and then pursuing the original subgoal without us having to change the subgoal. We also observe that the robot first aligns faces after performing a flip before attempting a rotation to avoid interlocking due to misalignment. Still, rotating a face can be challenging at times and we sometimes observe situations in which the robot is stuck. In this case we often see that the policy eventually adjusts its grasp to attempt the face rotation a different way, thus often succeeding eventually. Other times we observe our policy attempting a face rotation but the cube slips, resulting in a rotation of the entire cube as opposed to a specific face. In this case the policy rearranges its grasp and tries again, usually succeeding eventually.

We also observe that the policy appears more likely to drop the cube after being stuck on a challenging face rotation for a while. We do not quantify this but hypothesize that it might have ``forgotten'' about flips by then since the recurrent state of the policy has only observed a mostly stationary cube for several seconds. For flips, information about the cube's dynamics properties are more important. Similarly, we also observe that the policy appears to be more likely to drop the cube early on, presumably again because the necessary information about the cube's dynamic properties have not yet been captured in the policy's hidden state.

We also experiment with several perturbations. For example, we use a rubber glove to significantly change the friction and surface geometry of the hand. We use straps to tie together multiple fingers. We use a blanket to occlude the hand and Rubik's cube during execution. We use a pen and plush giraffe to poke the cube. While we do not quantify these experiments, we find that our policy still is able to perform multiple face rotations and cube flips under all of these conditions even though it was clearly not trained on them. \autoref{fig:cube-perturbations} shows examples of perturbations we tried. A video showing the behavior of our policy under those perturbations is also available: \mbox{\url{https://youtu.be/QyJGXc9WeNo}}

\footnotetext{Her name is Rubik, for obvious reasons.}

\section{Signs of Meta-Learning}
\label{sec:exp-meta}
We believe that a sufficiently diverse set of environments combined with a memory-augmented policy like an LSTM leads to \emph{emergent meta-learning}.
In this section, we systematically study our policies trained with ADR for signs of meta-learning.

\subsection{Definition of Meta-Learning}
Since we train each policy on only one specific task (i.e. the block reorientation task or solving the Rubik's cube), we define meta-learning in our context as learning about the dynamics of the underlying Markov decision process. More concretely, we are looking for signs where our policy updates its belief about the true transition probability $P(\vec{s}_{t+1} \mid \vec{s}_t, \vec{a}_t)$ as it observes data over time.

In other words, when we say ``meta-learning'', what we really mean is learning to learn about the environment dynamics. Within other communities, this is also called on-line system identification. In our case though, this is an emergent property.

\subsection{Response to Perturbations}
We start by studying the behavior of our policy and how it responds to a variety of perturbations to the dynamics of the environment. We conduct all experiments in simulation and use the Rubik's cube task. In all our experiments, we fix the type of subgoal we consider to be either cube flips or cube rotations. We run the policy until it achieves the $10$\textsuperscript{th} flip (or rotation) and then apply a perturbation. We then continue until the $30$\textsuperscript{th} successful flip (or rotation) and apply another perturbation. We measure the time it took to achieve the $1$\textsuperscript{st}, $2$\textsuperscript{nd}, $\ldots$, $50$\textsuperscript{th} flip (or rotation) for each trial, which we call ``time to completion'' We also measure during which flip (or rotation) the policy failed. By averaging over multiple trials that we all run in simulation, we can compute the average time to completion and failure probability \emph{per flip} (or rotation).

If our policy learns at test time, we'd expect the average time to completion to gradually decrease as the policy learns to identify the dynamics of its concrete environment and becomes more efficient as it accumulates more information over the course of a trial. Once we perturb the system, however, the policy needs to update its belief. We therefore expect to see a spike, i.e. achieving a flip (or rotation) should take longer after a perturbation but should again decrease as the policy readjusts its belief about the perturbed environment. Similarly, we expect to see the failure probability to be higher during the first few flips (or rotations) since the policy has had less time to learn about its environment. We also expect the failure probability to spike after each perturbation.

We experiment with the following perturbations:
\begin{itemize}
    \item \textbf{Resetting the hidden state.} During a trial, we reset the hidden state of the policy. This leaves the environment dynamics unchanged but requires the policy to re-learn them since its memory has been wiped.
    \item \textbf{Re-sampling environment dynamics.} This corresponds to an abrupt change of environment dynamics by resampling the parameters of all randomizations while leaving the simulation state\footnote{The simulation state is the current kinematic configuration of the cube, the hand, and the goal.} and hidden state intact.
    \item \textbf{Breaking a random joint.} This corresponds to us disabling a randomly sampled joint of the robot hand by preventing it from moving. This is a more nuanced experiment since the overall environment dynamics are the same but the way in which the robot can interact with the environment has changed.
\end{itemize}

\begin{figure}[h]
    \centering
    \begin{subfigure}[b]{\textwidth}
        \centering
        \includegraphics[width=0.47\textwidth]{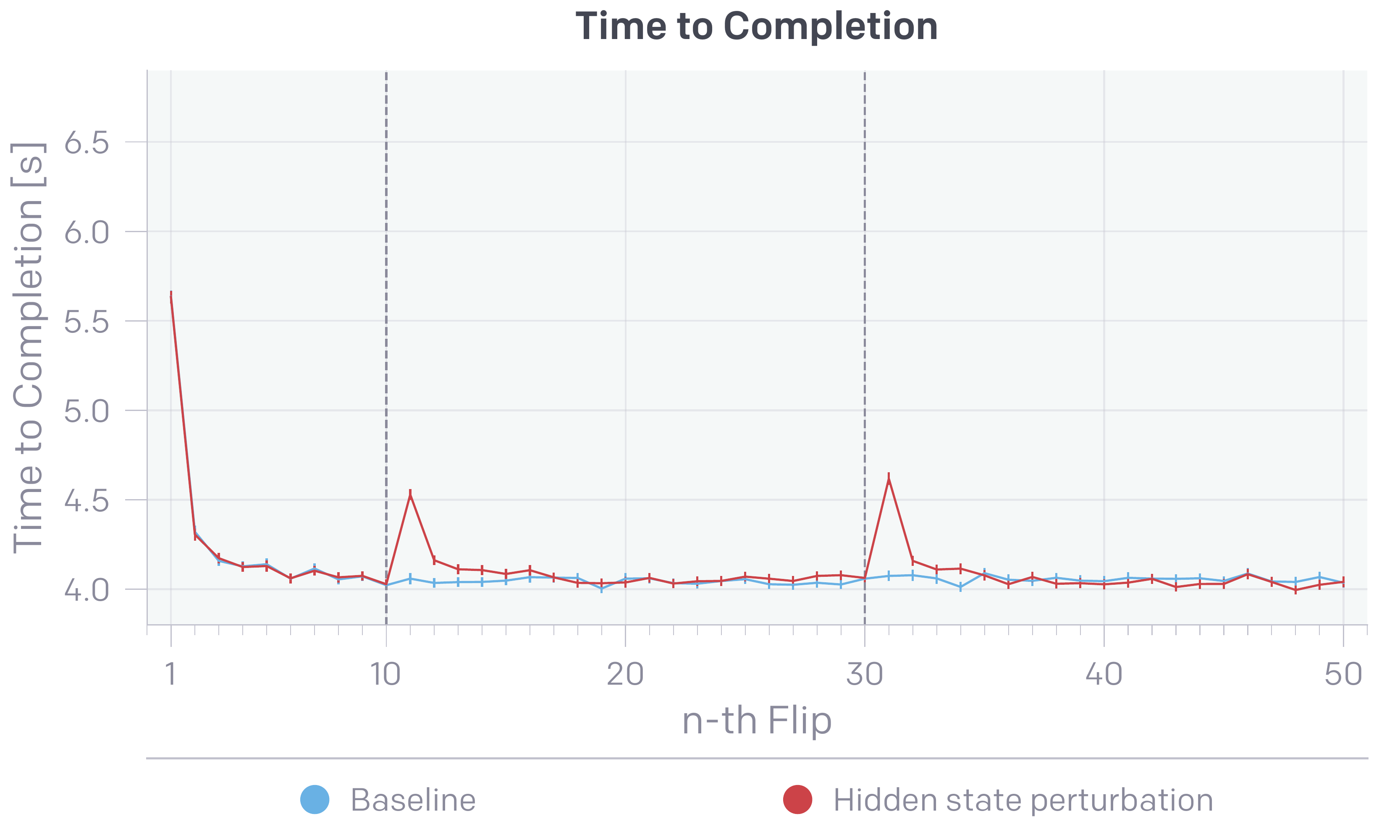}
        \hfill
        \includegraphics[width=0.47\textwidth]{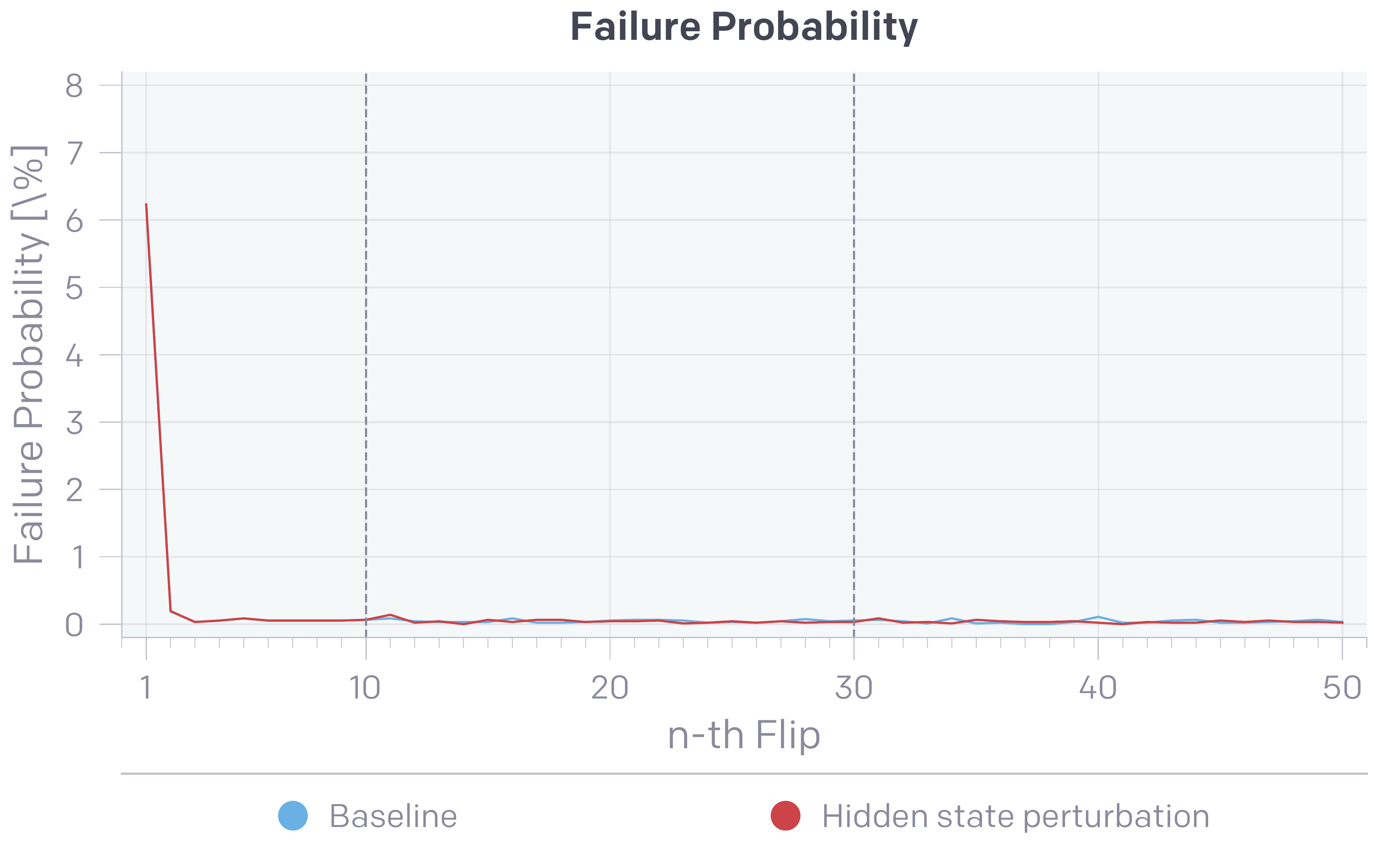}
        \caption{Resetting the hidden state.}
        \label{fig:meta-reset-hidden}
    \end{subfigure}
    \\
    \begin{subfigure}[b]{\textwidth}
        \centering
        \includegraphics[width=0.47\textwidth]{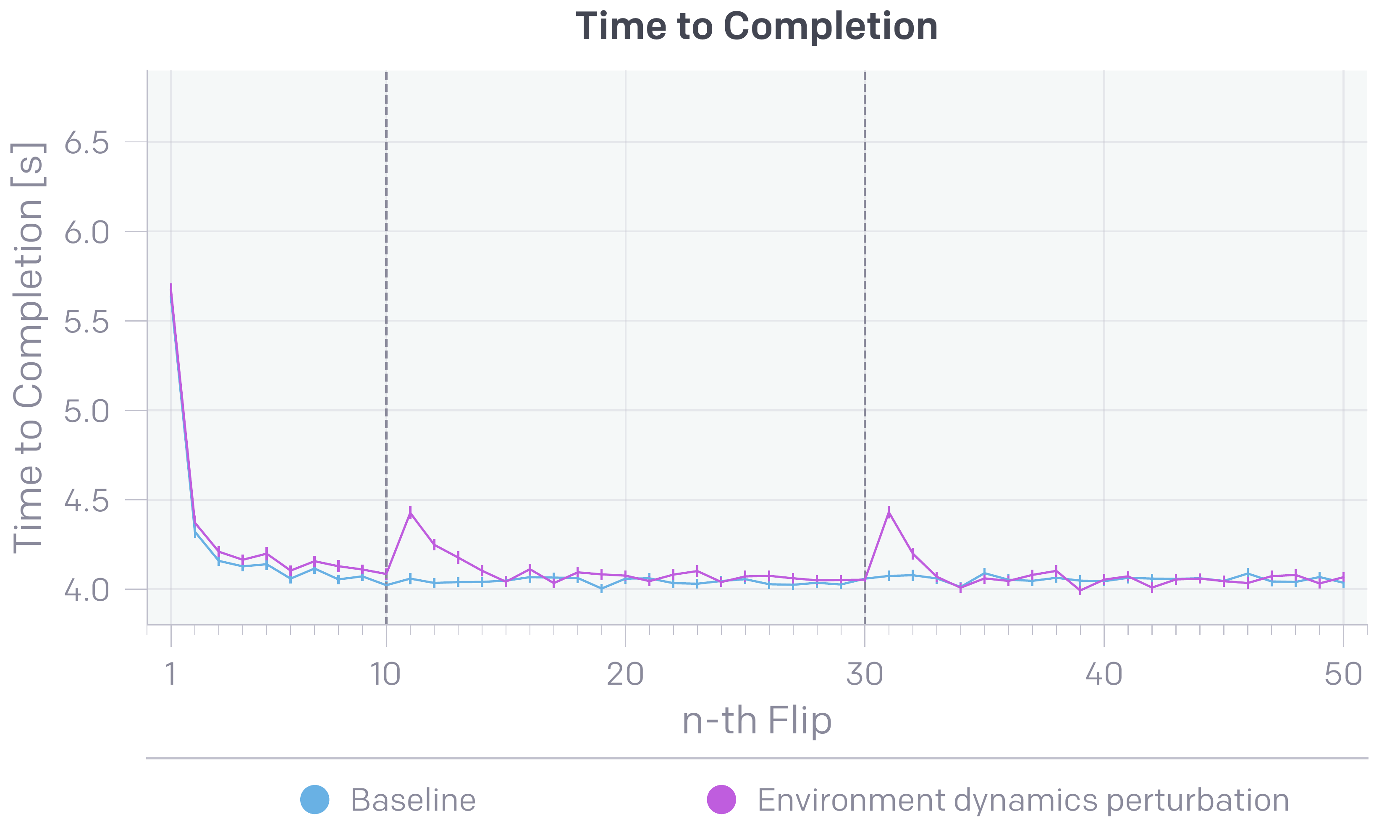}
        \hfill
        \includegraphics[width=0.47\textwidth]{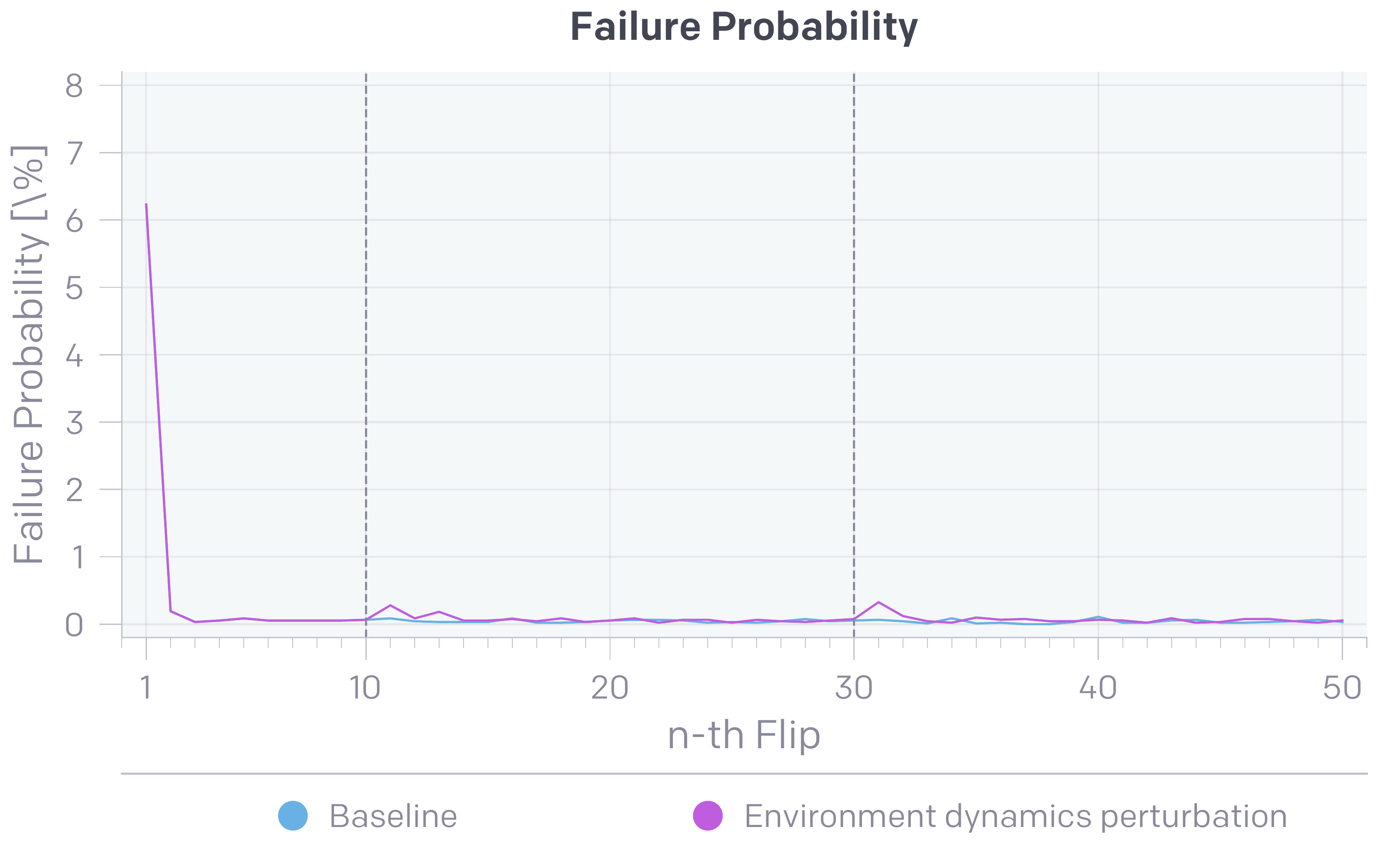}
        \caption{Re-sampling environment dynamics.}
        \label{fig:meta-reset-env}
    \end{subfigure}
    \\
    \begin{subfigure}[b]{\textwidth}
        \centering
        \includegraphics[width=0.47\textwidth]{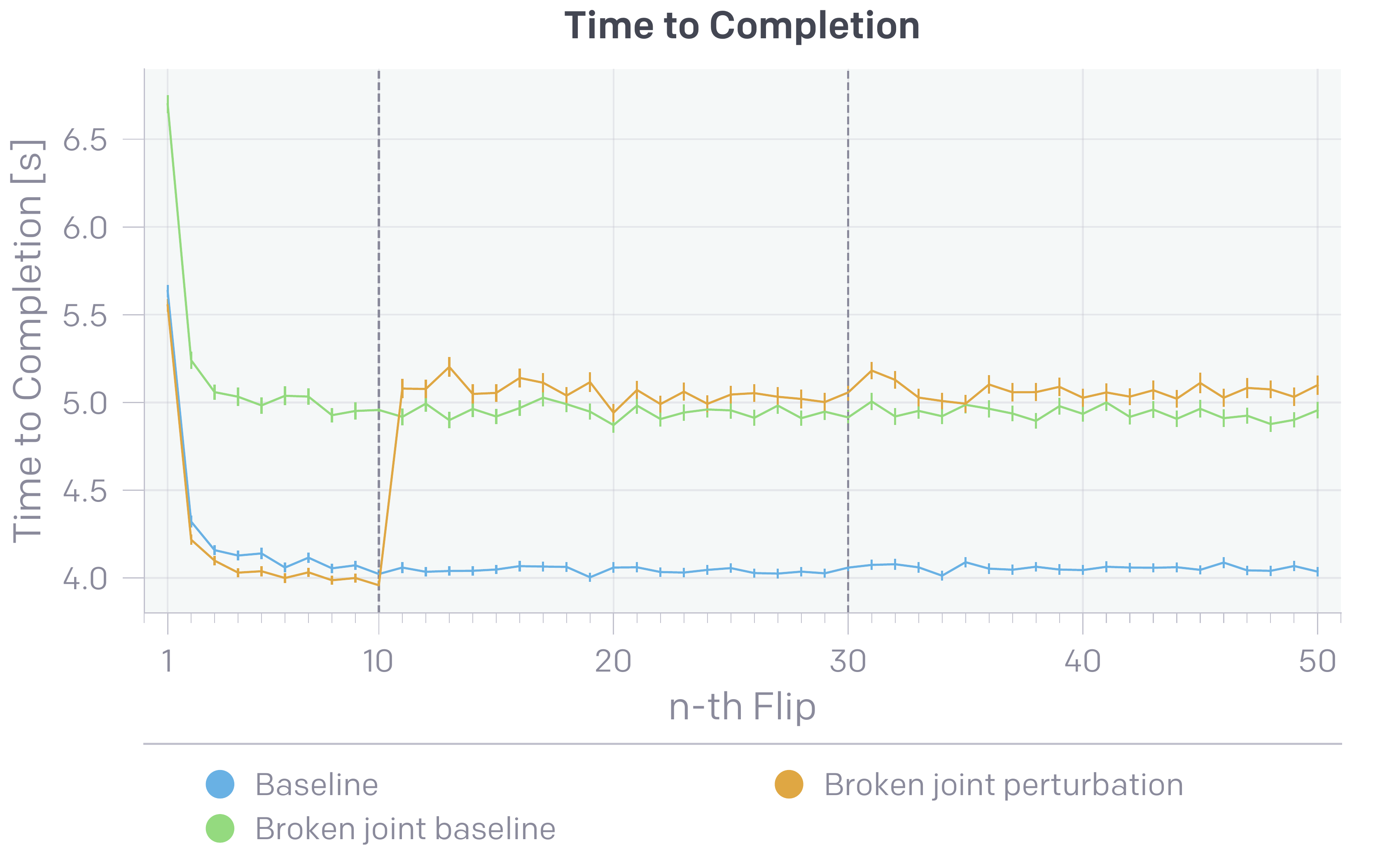}
        \hfill
        \includegraphics[width=0.47\textwidth]{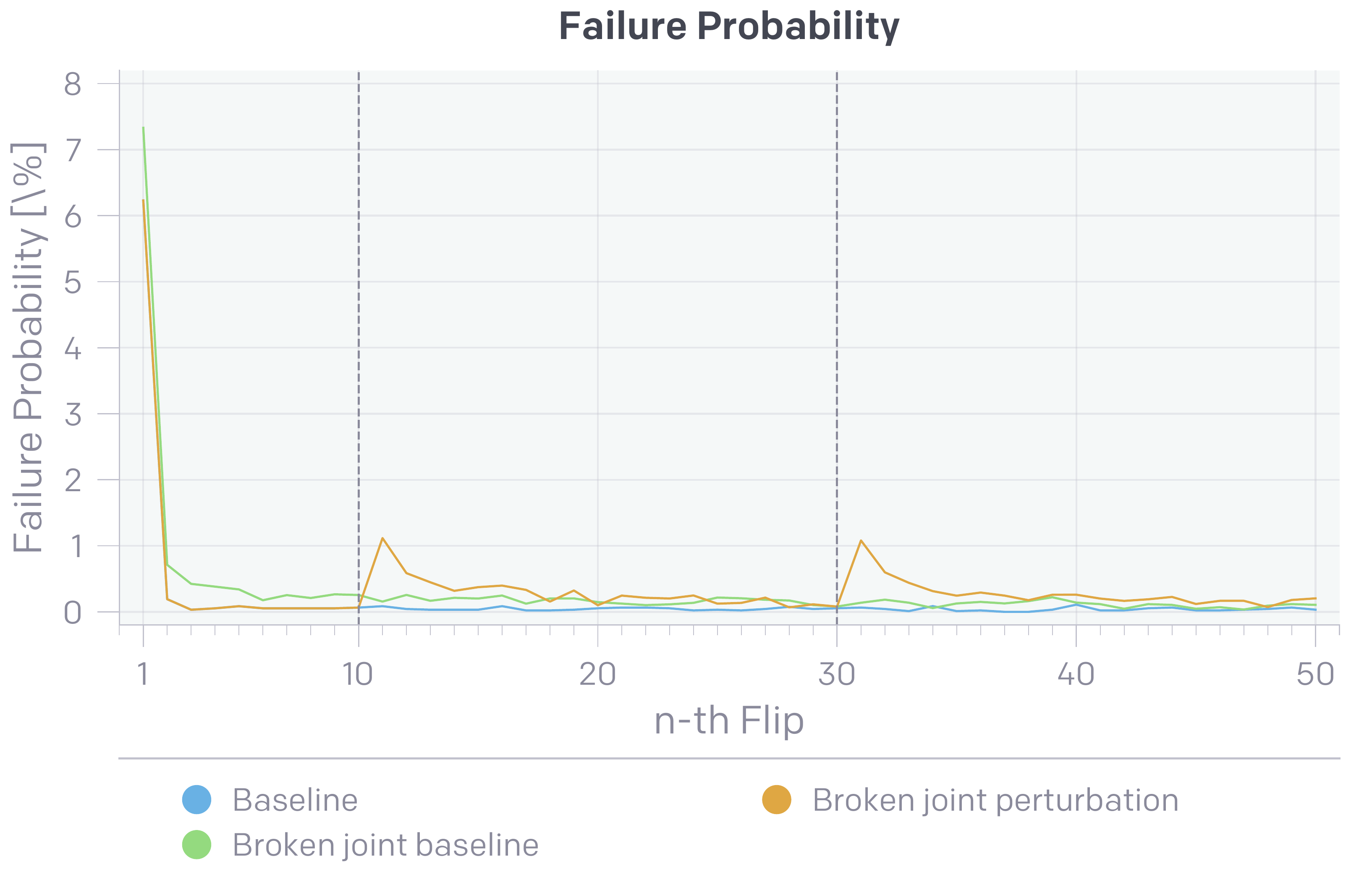}
        \caption{Breaking a random joint.}
        \label{fig:meta-reset-broken}
    \end{subfigure}
    \caption{We run $10\,000$ simulated trials with only cube flips until $50$ flips have been achieved. For each of the cube flips (i.e. the $1$\textsuperscript{st}, $2$\textsuperscript{nd}, $\ldots$, $50$\textsuperscript{th}), we measure the average time to completion (in seconds) and average failure probability over those $10$k trials. Error bars indicate the estimated standard error.  ``Baseline'' refers to a run without any perturbations applied. ``Broken joint baseline'' refers to trials where a joint was randomly disabled from the very beginning. We then compare against trials that start without any perturbations but are perturbed at the marked points after the $10$\textsuperscript{th} and $30$\textsuperscript{th} flip by (a) resetting the policy hidden state, (b) re-sampling environment dynamics, or (c) breaking a random joint.}
    \label{fig:meta-exp-1}
\end{figure}

We show the results of our experiments for cube flips in \autoref{fig:meta-exp-1}. The same plot is available for cube face rotations in \autoref{app:full-results-meta}. For the average time to completion, we only include trials that achieved $50$~flips to avoid inflating our results.\footnote{To explain further: By only include trials with $50$~successes, we ensure that we measure performance over a \emph{fixed} distribution over environments, and thus only measure how the performance of the policy changes across this fixed set. If we would not restrict this, harder environments would be more likely to lead to failures within the first few successes after a perturbation and then would ``drop out''. The remaining ones would be easier and thus even a policy without the ability to adapt would appear to improve in performance. Failures are still important, of course, which is why we report them in the failure probability plots.}

Our results show a very clear trend: for all runs, we observe a clear adjustment period over the first few cube flips. Achieving the first one takes the longest with subsequent flips being achieved more and more quickly. Eventually the policy converges to approximately $4$~seconds per flip on average, which is an improvement of roughly $1.6$~seconds compared to the first flip. This was exactly what we predicted: if the policy truly learns at test time by updating its recurrent state, we would expect it to become gradually more efficient. The same holds true for failure probabilities: the policy is much more likely to fail early.

When we reset the hidden state of our policy (compare \autoref{fig:meta-reset-hidden}), we can see the time to completion spike up significantly immediately after. This is because the policy again needs to identify the environment since all its memory has been wiped. Note that the spike in time to completion is much less than the initial time to completion. This is the case because we randomize the initial cube position and configuration. In contrast, when we reset the hidden state, the hand had manipulated the cube before so it is in a more beneficial position for the hand to continue after the hidden state is reset. This is also visible in the failure probability, which is close to zero even after the perturbation is applied, again because the cube is already in a beneficial position which makes it less likely to be dropped.

The second experiment perturbs the environment by resetting its environment dynamics but keeping the simulation state itself intact (see \autoref{fig:meta-reset-env}). We see a similar effect as before: after the perturbation is applied, the time to completion spikes up and then decreases again as the policy adjusts. Interestingly this time the policy is more likely to fail compared to resetting the hidden state. This is likely the case because the policy is ``surprised'' by the sudden change and performed actions that would have been appropriate for the old environment dynamics but led to failure in the new ones.

An especially interesting experiment is the broken joint one (see \autoref{fig:meta-reset-broken}): for the ``broken joint baseline'', we can see how the policy adjusts over long time horizons, with improvements clearly visible over the course of all $50$~flips in both time to completion and failure probability. In contrast, ``broken joint perturbation'' starts with all joints intact. After the perturbation, which breaks a random joint, we see a significant jump in the failure probability, which then gradually decreases again as the policy learns about this limitation. We also find that the ``broken joint perturbation'' performance never quite catches up to the ``Broken joint baseline''. We hypothesize that this could be because the policy has already ``locked-in'' some information in its recurrent state and therefore is not as adjustable anymore. Alternatively, maybe it just has not accumulated enough information yet and the baseline policy is in the lead because it has an ``information advantage'' of at least 10 achieved flips. We found it very interesting that our policy can learn to adapt internally to broken joints. This is in contrast to prior work that explicitly searched over a set of policies until it found one that works for a broken robot~\cite{cully2015robots}. Note however that the ``Broken joint baseline'' never fully matches the performance of the ``Baseline'', suggesting that the policy can not fully recover performance.

In summary, we find clear evidence of our policy learning about environment dynamics and adjusting its behavior accordingly to become more efficient at \emph{test time}. All of this learning is emergent and only happens by updating the policy's recurrent state.

\subsection{Recurrent State Analysis}

We conducted experiments to study whether the policy has learned to infer and store useful information about the environment in its recurrent state. We consider such information to be strong evidence of meta-learning, since no explicit information regarding the environment parameters was provided during training time.

The main method that we use to probe the amount of useful environment information is to predict environment parameters, such as cube size or the gravitational constant, from the policy LSTM hidden and cell states. Given a policy with an LSTM with hidden states $\vec{h}$ and cell states $\vec{c}$, we use $\vec{z} = \vec{h} + \vec{c}$ as the prediction model input. 
For environment parameter $p$, we trained a simple prediction model $f_p(\vec{z})$ containing one hidden layer with $64$~units and ReLU activation, followed by a sigmoid output. The outputs correspond to the probability that the value of $p$ for the environment is greater or smaller than the average randomized value.

The prediction model was trained on hidden states collected from policy rollouts at time step $t$ in a set of environments $\mathcal{E}_t$. Each environment $e \in \mathcal{E}_t$ contains a different value of the parameter $p$, sampled according to its randomization range. To observe the change in the stored information over time, we collected data from times steps $t \in \{1, 30, 60, 120\}$ where each time step is equal to $\Delta t = \SI{0.08}{\second}$ in simulation. We used the cross-entropy loss and trained the model until convergence. The model was tested on a new set of environments $\mathcal{F}_t$ with newly sampled values of $p$.
\subsubsection{Prediction Accuracy over Time}

We studied four different environment parameters, listed in \autoref{table:prediction_parameters} along with their randomization ranges. The randomization ranges were used to generate the training and test environment sets $\mathcal{E}_{t, p}$, $\mathcal{F}_{t, p}$. Each parameter is sampled using a uniform distribution over the given range. Randomization ranges were taken from the end of ADR training.

\begin{table}[h]
    \caption{Prediction environment parameters and their randomization ranges in physical units. We predict whether or not the parameter is larger or smaller than the given average.}
    \centering
    \renewcommand{\arraystretch}{1.3}
    \begin{tabular}{@{}lrrrr@{}}
        \toprule
        \multirow{2}{*}{\textbf{Parameter}} & \multicolumn{2}{c}{\textbf{Block Reorientation}} & \multicolumn{2}{c}{\textbf{Rubik's Cube}} \\
        & \multicolumn{1}{c}{\textbf{Average}}
        & \multicolumn{1}{c}{\textbf{Range}}
        & \multicolumn{1}{c}{\textbf{Average}}
        & \multicolumn{1}{c}{\textbf{Range}} \\
        \midrule
        Cube size [\si{\meter}] & $0.055$ & $[0.046, 0.064]$ & $0.057$ & $[0.049, 0.066]$ \\
        Time step [\si{\second}] & $0.1$ & $[0.05, 0.15]$ & $0.09$ & $[0.05, 0.13]$ \\
        Gravity [\si{\meter\per\square\second}] & $9.80$ & $[6.00, 14.0]$ & $9.80$ & $[6.00, 14.0]$ \\
        Cube mass [\si{\kilogram}] & $0.0780$ & $[0.0230, 0.179]$ & $0.0902$ & $[0.0360, 0.158]$\\
        \bottomrule
    \end{tabular}
    \label{table:prediction_parameters}
\end{table}

\begin{figure}[h]
    \centering
    \begin{subfigure}[b]{0.48\textwidth}
        \includegraphics[width=\textwidth]{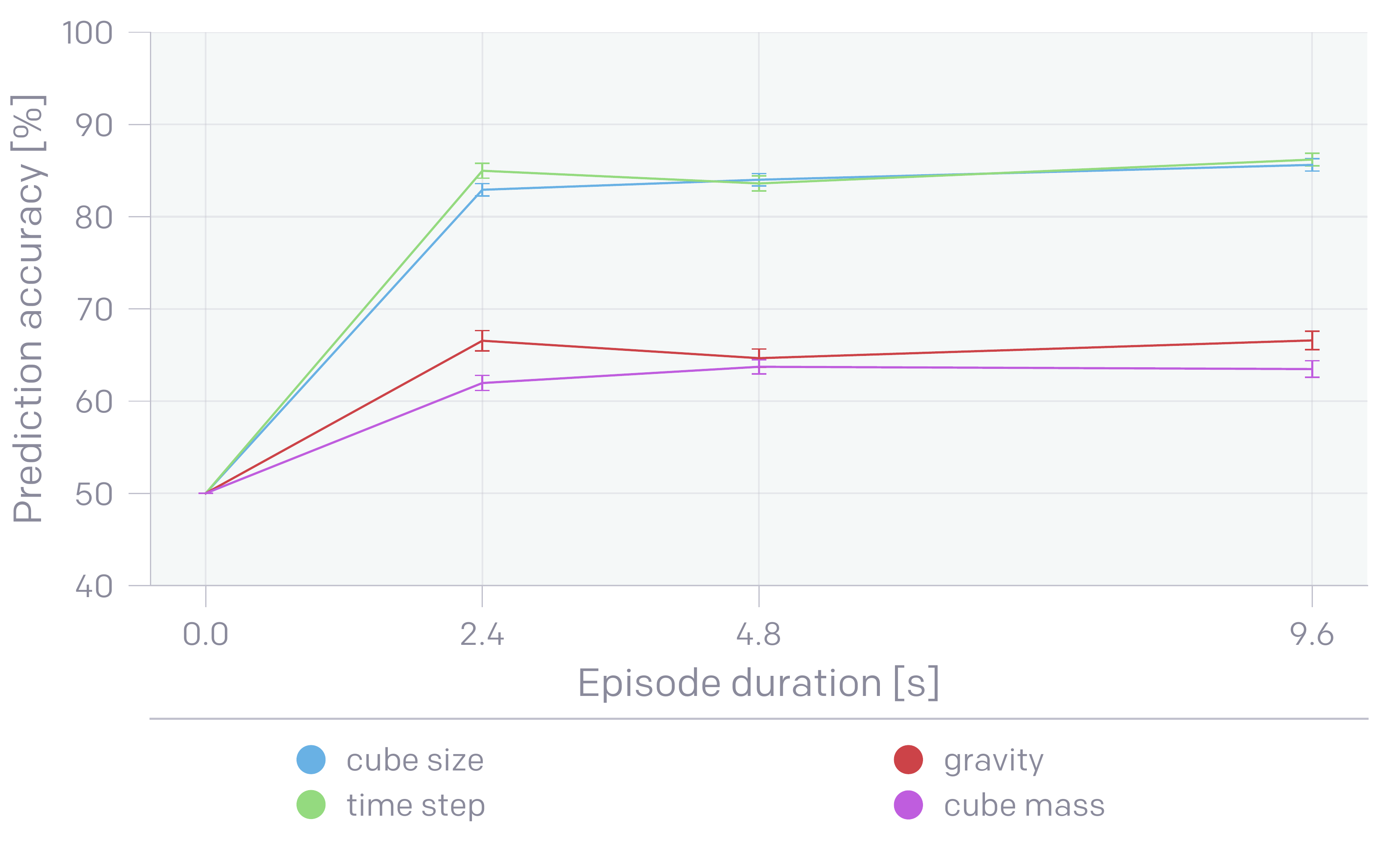}
        \caption{Block reorientation policy}
        \label{fig:meta-hidden-locked}
    \end{subfigure}
    \hfill
    \begin{subfigure}[b]{0.48\textwidth}
        \includegraphics[width=\textwidth]{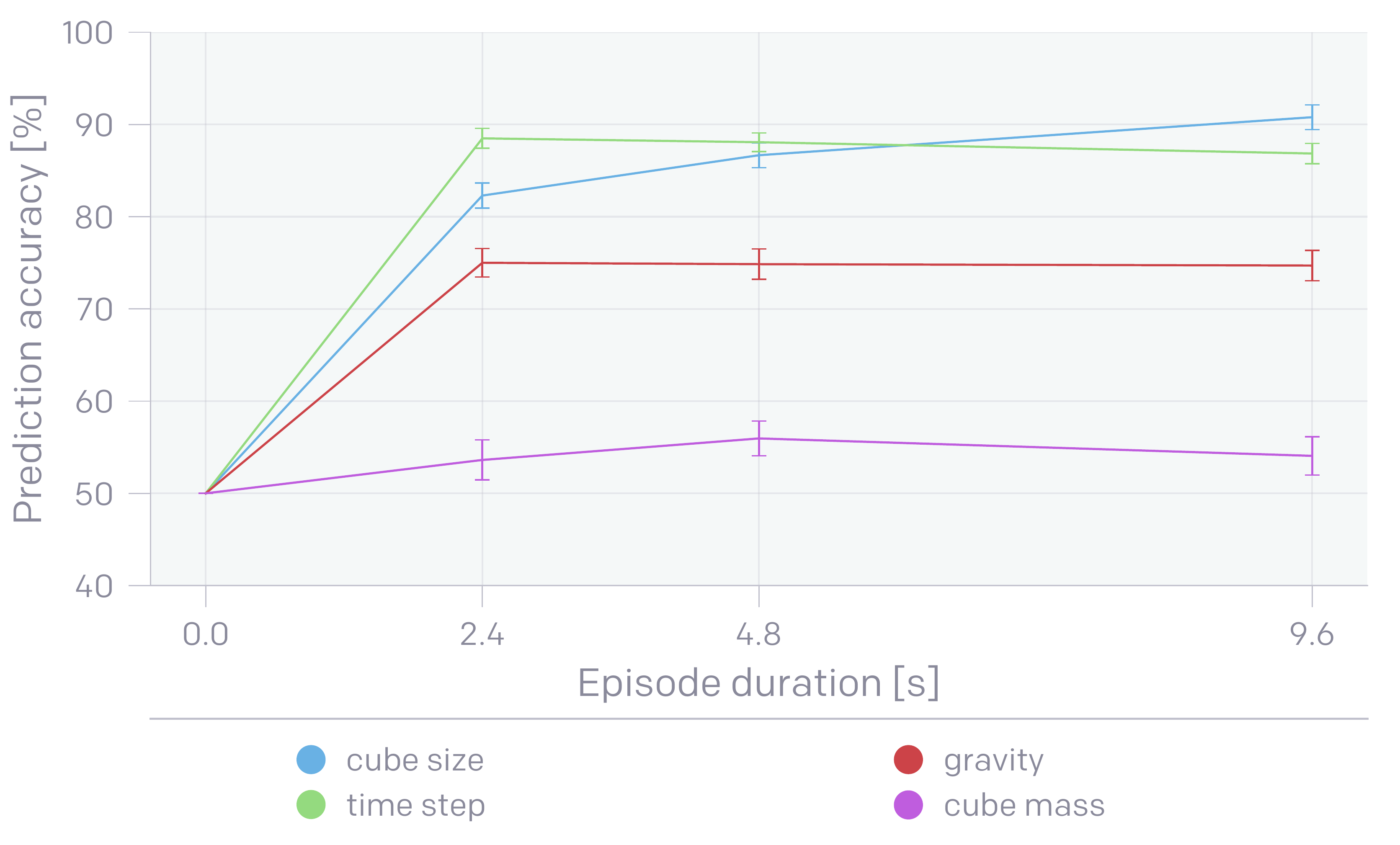}
        \caption{Rubik's cube policy}
        \label{fig:meta-hidden-full}
    \end{subfigure}
    \caption{Test accuracy of environment parameter prediction model based on the hidden states of (a) block reorientation and (b) Rubik's cube policies. Error bars denote the standard error.}
    \label{fig:meta-hidden}
\end{figure}

\autoref{fig:meta-hidden} shows the test accuracy of the trained prediction models for block reorientation and Rubik's cube policies. Observe that prediction accuracy at the start of a rollout is near random guessing, since no useful information is stored in the hidden states. 

As rollouts progress further and the policy interacts with the environment, prediction accuracy rapidly improves to over $80\%$ for certain parameters. This is evidence that the policy is successfully inferring and storing useful information regarding the environment parameters in its LSTM hidden and cell states. Note that we do not train the policy explicitly to store information about those semantically meaningful physical parameters.

There exists some variability in the prediction accuracy over different parameters and between block reorientation  and Rubik's cube policies. For example, note that the prediction accuracy for cube size (over $80\%$) is consistently higher than that of cube mass ($50-60\%$). This may be due to the relative importance of cube size and mass to policy performance; i.e., a heavier cube changes the difficulty of Rubik's cube face rotation less than a larger cube. There also exist differences for the same parameter between tasks: For the cube mass, the block reorientation policy stores more information about it then the Rubik's cube policy. We hypothesize that this is because the block reorientation policy uses a dynamic approach that tosses the block around to flip it. In contrast, the Rubik's cube policy flips the cube much more deliberately in order to avoid unintentional misalignments of the cube faces. For the former, knowing the cube mass is therefore more important since the policy needs to be careful to not apply too much force. We believe the variations in prediction accuracy for the other parameters and the two policies also reflect the relative importance of each parameter to the policy and the given task.

\subsubsection{Information Gain over Time}

To further study the information contained in a policy hidden state and how it evolves during a rollout, we expanded the prediction model in the above experiments to predict a set of 8 equally-spaced discretized values (``bins'') within a parameter's randomization range. We consider the output probability distribution of the predictor to be an approximate representation of the posterior distribution over the environment parameter, as inferred by the policy. 

\begin{figure}[h]
    \centering
    \includegraphics[width=0.7\textwidth]{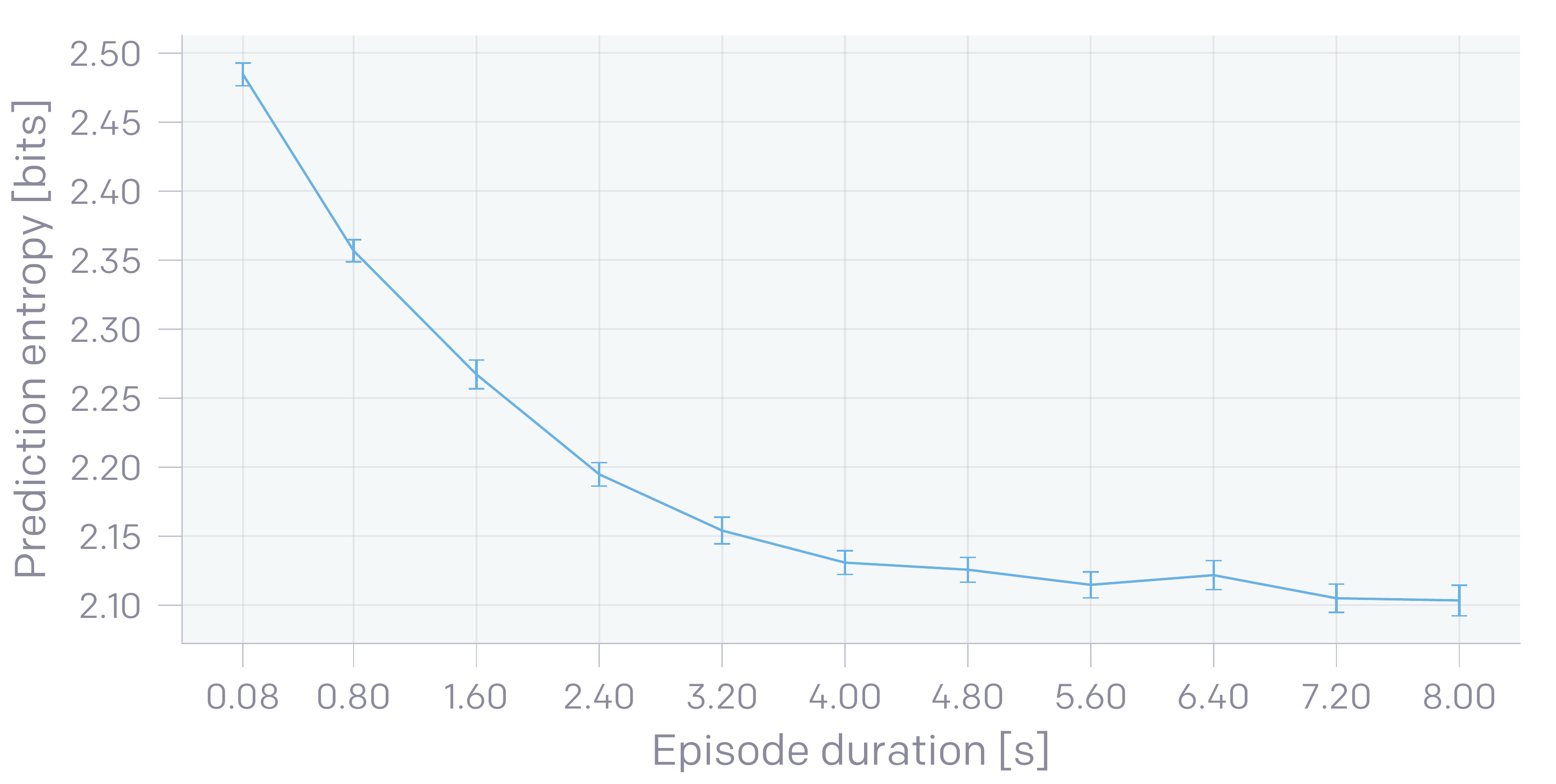}
    \caption{Mean prediction entropy over rollout time for an 8-bin output predictor. Error bars denote the standard error. The predictor was trained at a fixed $4.8$ seconds during rollouts. Parameter being predicted is cube size. Note the information gain of $0.9$ bits in less than $5.0$ seconds. For reference, the entropy for random guessing (i.e. uniform probability mass over all $8$~bins) is $3$~bits.}
    \label{fig:meta-entropy}
\end{figure}

In \autoref{fig:meta-entropy}, we plot the entropy of the predictor output distribution in test environments over rollout time. The parameter being predicted is cube size. The results clearly show that the posterior distribution over the environment parameters converges to a certain final distribution as a rollout progresses. The convergence speed is rather fast at below $5.0$~seconds. Notice that this is consistent with our perturbation experiments (compare \autoref{fig:meta-hidden}): the first flip roughly takes $5-6$ seconds and we see a significant speed-up after. Within this time, the information gain for the cube size parameter is approximately $0.9$~bits. Interestingly, the entropy eventually seems to converge to $2.0$~bits and then stops decreasing. This again highlights that our policies only store the amount of information they need to act optimally.

\subsubsection{Prediction Accuracy and ADR Entropy}

We performed the hidden state prediction experiments for block reorientation policies trained using ADR with different values of ADR entropy. Since we believe that the ability of the policy to infer and store (i.e., meta-learn) useful information regarding environment parameters is correlated with the diversity of the environments used during training, we expect the prediction accuracy to be positively correlated with the policy's ADR entropy. 

Four block reorientation policies corresponding to increasing ADR entropy were used in the following experiments. We seek to predict the cube size parameter at $60$ rollout time steps, which corresponds to $4.8$~seconds of simulated time. The test accuracies are shown in \autoref{table:prediction_volume_locked}. The results indicate that prediction accuracy (hence information stored in hidden states) is strongly correlated with ADR entropy. 

\begin{table}[h]
    \caption{Block reorientation policy hidden state prediction over ADR entropy. The environment parameter predicted is cube size. ADR entropy is defined to be nats per environment parameterization dimension or npd. All predictions were performed at rollout time step $t=60$ (which corresponds to $4.8$~seconds of simulated time). We report mean prediction accuracy $\pm$ standard error.}
    \centering
    \renewcommand{\arraystretch}{1.3}
    \begin{tabular}{@{}lrrc@{}}
        \toprule
        \textbf{Policy} & \textbf{Training Time} & \textbf{ADR Entropy} & \textbf{Prediction Accuracy} \\
	\midrule
	ADR (Small) & 0.64 days & $-0.881$ npd & $0.68 \pm 0.021$ \\
	ADR (Medium) & 4.37 days & $-0.135$ npd & $0.75 \pm 0.027$ \\
	ADR (Large) & 13.76 days & $0.126$ npd & $0.79 \pm 0.022$ \\
	ADR (XL) & --- & $0.305$ npd & $\mathbf{0.83 \pm 0.014}$ \\
        \bottomrule
    \end{tabular}
    \label{table:prediction_volume_locked}
\end{table}

\subsubsection{Recurrent state visualization}
We used neural network interpretability techniques~\cite{carter2016experiments, olah2018the} to visualize policy recurrent states during a rollout. We found distinctive activation patterns that correspond to high-level skills exhibited by the robotic hand. See~\autoref{app:policy_visualizations} for details.

\section{Related Work}
\label{sec:related}
\subsection{Dexterous Manipulation}
Dexterous manipulation has been an active area of research for decades~\citep{DBLP:conf/icra/Fearing86, DBLP:journals/ijrr/Rus99,DBLP:journals/trob/Bicchi00, DBLP:conf/icra/OkamuraSC00, DBLP:conf/icar/MaD11}.
Many different approaches and strategies have been proposed over the years.
This includes rolling~\citep{DBLP:conf/icra/BicchiS95, DBLP:conf/icra/HanGLQT97, DBLP:conf/icra/HanT98, DBLP:journals/trob/CherifG99, DBLP:conf/icra/DoulgeriD13}, sliding~\citep{DBLP:journals/trob/CherifG99, DBLP:journals/trob/ShiWUL17}, finger gaiting~\citep{DBLP:conf/icra/HanT98}, finger tracking~\citep{DBLP:conf/icra/Rus92}, pushing~\citep{DBLP:journals/corr/DafleR17}, and re-grasping~\citep{DBLP:conf/icra/TournassoudLM87, DBLP:conf/icra/DafleRPTSEMLSF14}.
For some hand types, strategies like pivoting~\citep{DBLP:conf/iros/AiyamaII93}, tilting~\citep{DBLP:journals/trob/ErdmannM88}, tumbling~\citep{sawasaki1991tumbling}, tapping~\citep{DBLP:journals/ijrr/HuangM00}, two-point manipulation~\citep{DBLP:conf/iros/AbellE95}, and two-palm manipulation~\citep{DBLP:journals/ijrr/Erdmann98} are also options.
These approaches use planning and therefore require exact models of both the hand and object.
After computing a trajectory, the plan is typically executed open-loop, thus making these methods prone to failure if the model is not accurate.\footnote{Some methods use iterative re-planning to partially mitigate this issue.}

Other approaches take a closed-loop approach to dexterous manipulation and incorporate sensor feedback during execution, e.g. tactile sensing~\citep{DBLP:conf/icra/TaharaAY10, DBLP:conf/iros/LiBKB14, DBLP:conf/icra/LiYTB14, DBLP:journals/ijma/0001MHRB13}.
While those approaches allow for the correction of mistakes at runtime, they still require reasonable models of the robot kinematics and dynamics, which can be challenging to obtain for under-actuated hands with many degrees of freedom.

Deep reinforcement learning has also been used successfully to learn complex manipulation skills on physical robots.
Guided policy search~\citep{DBLP:conf/icml/LevineK13, DBLP:conf/icra/LevineWA15} learns simple local policies directly on the robot and distills them into a global policy represented by a neural network. Soft Actor-Critic\cite{DBLP:journals/corr/abs-1812-05905} has been recently proposed as a state-of-the-art model-free algorithm optimizing concurrently both expected reward and action entropy, that is capable of learning complex behaviors directly in the real world.

Alternative approaches include using many physical robots simultaneously, in order to be able to collect sufficient experience~\citep{DBLP:conf/icra/GuHLL17, DBLP:journals/ijrr/LevinePKIQ18, 2018arXiv180610293K} or leveraging a model-based learning algorithms, which generally possess much more favorable sample complexity characteristics~\cite{DBLP:journals/corr/abs-1808-09105}. Some researchers have successfully  utilized expert human demonstrations in guiding the training process of the agents~\citep{DBLP:journals/corr/abs-1906-11695,DBLP:journals/corr/abs-1903-01973,DBLP:journals/corr/abs-1810-01845,DBLP:journals/corr/abs-1810-06045}.

\subsection{Dexterous In-Hand Manipulation}
Since a very large body of past work on dexterous manipulation exists, we limit the more detailed discussion to setups that are most closely related to our work on dexterous in-hand manipulation.

Mordatch et al.~\citep{DBLP:conf/sca/MordatchPT12} and Bai et al.~\citep{DBLP:conf/icra/BaiL14} propose methods to generate trajectories for complex and dynamic in-hand manipulation, but their results are limited to simulation.
There has also been significant progress in learning complex in-hand dexterous manipulation~\citep{plappert2018multi, DBLP:journals/corr/abs-1804-08617}, tool use~\citep{DBLP:journals/corr/abs-1709-10087} and even solving a smaller model of a Rubik's Cube \cite{2019arXiv190711388L} using deep reinforcement learning, but those approaches were likewise only evaluated in simulation.

In contrast, multiple authors learn policies for dexterous in-hand manipulation directly on the robot.
Hoof et al.~\citep{DBLP:conf/humanoids/HoofHN015} learn in-hand manipulation for a simple 3-fingered gripper whereas Kumar et al.~\citep{DBLP:conf/icra/KumarTL16, DBLP:journals/corr/KumarGTL16} and Falco et al.~\citep{falco2018policy} learn such policies for more complex humanoid hands. In \cite{PDDM}, the authors learn a forward dynamics model and use model predictive control to manipulate two Baoding balls with a Shadow hand.
While learning directly on the robot means that modeling the system is not an issue, it also means that learning has to be performed with limited data.
This is only possible when learning simple (e.g. linear or local) policies that, in turn, do not exhibit sophisticated behaviors.

\subsection{Sim to Real Transfer}

\emph{Domain adaption} methods~\citep{DBLP:journals/corr/TzengDHFPLSD15, DBLP:journals/corr/GuptaDLAL17}, progressive nets~\citep{DBLP:conf/corl/RusuVRHPH17}, and learning inverse dynamics models~\citep{DBLP:journals/corr/ChristianoSMSBT16} were all proposed to help with sim to real transfer.
All of these methods assume access to real data.
An alternative approach is to make the policy itself more adaptive during training in simulation using \emph{domain randomization}.
Domain randomization was used to transfer object pose estimators~\citep{tobin2017domain} and vision policies for fly drones~\citep{DBLP:conf/rss/SadeghiL17}.
This idea has also been extended to dynamics randomization~\citep{peng2017sim, DBLP:journals/corr/AntonovaCSK17, DBLP:journals/corr/abs-1804-10332, DBLP:conf/rss/YuTLT17, openai2018learning} to learn a robust policy that transfers to similar environments but with different dynamics.
Domain randomization was also used to plan robust grasps~\citep{DBLP:conf/rss/MahlerLNLDLOG17, DBLP:journals/corr/abs-1709-06670, DBLP:journals/corr/abs-1710-06425} and to transfer learned locomotion~\citep{DBLP:journals/corr/abs-1804-10332} and grasping~\citep{DBLP:journals/corr/abs-1802-09564} policies for relatively simple robots.
Pinto et al.~\citep{pinto2017robust} propose to use \emph{adversarial training} to obtain more robust policies and show that it also helps with transfer to physical robots~\citep{pinto2017supervision}. Hwangbo et al.~\cite{hwangbo2019learning} used real data to learn an actuation model and combined it with domain randomization to successfully transfer locomotion policies.

A number of recent works have focused on adapting the environment distribution of domain randomization to improve sim-to-real transfer performance. For policy training, one approach viewed the problem as a bi-level optimization~\citep{Vuong2019, Ruiz2018}. Chebotar et al.~\citep{Chebotar2018} used real-world trajectories and a discrepancy metric to guide the distribution search. In~\citep{Mehta2019}, a discriminator was used to guide the distribution in simulation. For vision models, domain randomization has been modified to improve image content diversity~\citep{Kar2019, Prakash2018, Cubuk2018} and to adapt the distribution by using an adversarial network~\citep{Zakharov2019}.

\subsection{Meta-Learning via Reinforcement Learning}

Despite being a very young field, meta-learning in the context of deep reinforcement learning already has a large body of work published. Algorithms such as MAML~\cite{DBLP:journals/corr/FinnAL17} and SNAIL~\cite{DBLP:journals/corr/MishraRCA17} have been developed to improve the sample efficiency of reinforcement learning agents. A common theme in research is to try to exploit a shared structure in a distribution of environments, to quickly identify and adapt to previously unseen cases~\cite{DBLP:journals/corr/abs-1907-04964, Smundsson2018MetaRL,DBLP:journals/corr/abs-1905-06424,Lan2019MetaRL}
There are works that directly treat meta-learning as identification of a dynamics model of the environment~\cite{DBLP:journals/corr/abs-1803-11347} while others tackle the problem of task discovery for training~\cite{DBLP:journals/corr/abs-1806-04640}.
Meta-learning has also been studied in a multi-agent setting~\cite{DBLP:journals/corr/abs-1903-02710}.

The approach we've taken is directly based on $RL^2$~\cite{DBLP:journals/corr/DuanSCBSA16,DBLP:journals/corr/WangKTSLMBKB16} where a general-purpose optimization algorithm trains a model augmented with memory to perform independent learning algorithm in the inner loop. The novelty in our results comes from the combination of automated curriculum generation (ADR), a challenging underlying problem (solving Rubik's Cube) and a completely out-of-distribution test environment (sim2real). 

This approach coincides with what Jeff Clune described as AI-GA~\cite{clune2019ai}, the AI-generating algorithms, whose one of three pillars is generating effective and diverse learning environments. In similar spirit to our ADR, related works include  PowerPlay~\cite{DBLP:journals/corr/abs-1112-5309,DBLP:journals/corr/abs-1210-8385}, POET~\cite{wang2019poet} and different varieties of self-play~\cite{alphago,DBLP:journals/corr/abs-1712-01815,five}.

\subsection{Robotic Systems Solving Rubik's Cube}

While many robots capable of solving a Rubik's Cube exist today~\cite{rubikguinnes,mindcuber,rcr3d,higo2018rubik}, all of them which we are aware of have been built exclusively for this purpose and cannot generalize to other manipulation tasks.

\section{Conclusion}
\label{sec:conclusion}
In this work, we introduce automatic domain randomization (ADR), a powerful algorithm for sim2real transfer. We show that ADR leads to improvements over previously established baselines that use manual domain randomization for both vision and control. We further demonstrate that ADR, when combined with our custom robot platform, allows us to successfully solve a manipulation problem of unprecedented complexity: solve a Rubik's cube using a real humanoid robot, the Shadow Dexterous Hand. By systematically studying the behavior of our learned policies, we find clear signs of emergent meta-learning. Policies trained with ADR are able to adapt at deployment time to the physical reality, which it has never seen during training, via updates to their recurrent state.

\section*{Acknowledgements}
We would like to thank Shadow Robot Company Ltd. for building, maintaining, and improving the Shadow Dexterous Hand, PhaseSpace Inc. for building custom Rubik's cubes and supporting our motion capture setup, JITX for building custom electronics and writing firmware for the Giiker cube, and Kimly Do for designing the cage. We would also like to thank Chris Hallacy, Jacob Hilton, and Ludwig Schubert for volunteering as human Rubik's cube solvers and and everybody at OpenAI for their help and support.

We would also like to thank the following people for providing thoughtful feedback on earlier versions of this
manuscript: Josh Achiam, Nick Cammarata, Jeff Clune, Harri Edwards, David Farhi, Ken Goldberg, Lerrel Pinto, and John Schulman.

\section*{Author Contributions}
This manuscript is the result of the work of the entire OpenAI Robotics team. We list the contributions of every team member here grouped by topic and in alphabetical order.

\begin{itemize}
    \item Marcin Andrychowicz, Matthias Plappert, and Raphael Ribas developed the first version of ADR.
    
    \item Maciek Chociej, Alex Paino, Peter Welinder, Lilian Weng, and Qiming Yuan developed the vision state estimator.
    
    \item Maciek Chociej, Peter Welinder and Lilian Weng applied ADR to vision.
    
    \item Ilge Akkaya, Marcin Andrychowicz, Matthias Plappert, Raphael Ribas, Nikolas Tezak, Jerry Tworek, and Lei Zhang contributed to the RL training setup.
    
    \item Mateusz Litwin, Jerry Tworek, and Lei Zhang improved ADR for RL training.
    
    \item Mateusz Litwin improved the robot model and Jerry Tworek developed the Rubik's cube simulation model.
    
    \item Arthur Petron and Jonas Schneider built the physical robot setup.
    
    \item Ilge Akkaya, Maciek Chociej, Mateusz Litwin, Arthur Petron, Glenn Powell, Jonas Schneider, Peter Welinder, Lilian Weng, and Qiming Yuan contributed software for the robot platform.
    
    \item Mateusz Litwin, Glenn Powell, and Jonas Schneider developed system infrastructure.
    
    \item Ilge Akkaya, Mateusz Litwin, Arthur Petron, Alex Paino, Matthias Plappert, Jerry Tworek, Lilian Weng, and Lei Zhang wrote this manuscript.
    
    \item Marcin Andrychowicz, Bob McGrew, Matthias Plappert, Jonas Schneider, Peter Welinder, and Wojciech Zaremba led aspects of this project and set research directions.
    
    \item Wojciech Zaremba led the team.
\end{itemize}

\bibliographystyle{abbrv}
\bibliography{references}

\newpage
\appendix
\appendixpage

\section{System Monitoring}
\label{app:hardware-monitoring}

Reliability of the physical setup was one of the key challenges mentioned in \cite{openai2018learning}. To address this issue we built new monitoring system which ensures experiments were performed on a fully functional robot and that physical setup was consistent for all experiments. 
New system includes recording, visualization, alerting and persistence of all operations performed on the physical robots. New debugging and investigation capabilities were unlocked and allowed us to detect robot breakage in real time (e.g. breaking tendons), find regressions in the physical setup and identify differences between policies deployed on the physical setup and in the simulation.

We used InfluxDB\footnote{InfluxDB is a time series database designed to handle high write and query loads. See \url{https://www.influxdata.com/products/influxdb-overview} for more information.} database to do real-time tracking and persistence of native resolution sensor data and detailed information about the context in which experiments were conducted.

We used Grafana\footnote{Grafana is an open-source, general purpose dashboard and graph composer, which runs as a web application. See \url{https://grafana.com} for more information.}, with InfluxDB as a data source, to display information on policy performance on the task, to visualize data from robot sensors and to show the health status of robotic components.

\section{Randomizations}
\label{app:randomizations}

Detailed listings of the randomizations used in policy and vision training are given in Tables~\ref{table:manipulation_randomizations} and~\ref{table:vision_randomizations}. Below, we describe how the randomizations are parameterized in ADR.

\begin{table}[h]
    \caption{Randomizations used to train manipulation policies with ADR. For simulator physics randomizations with generic randomizers, values in the parenthesis denote (noise mode, $\alpha$). Generic randomizers without parenthesis used default values (MG, $1.0$).}
    \centering
    \renewcommand{\arraystretch}{1.3}
    \begin{tabular}{@{}l|l|l|l@{}}
        \toprule
        \textbf{Category} & \textbf{All policies} & \textbf{Reorientation policy} & \textbf{Rubik's cube policy} \\
        \midrule
	\multirow{16}{2cm}{Simulator physics (generic)} 	
       			& Actuator force range & Dof armature & Body position (AG, 0.02) \\
       			& Actuator gain prm & Dof damping & Dof armature cube \\
			& Body inertia & Dof friction loss & Dof armature robot \\
			& Geom size robot spatial & Geom friction	 & Dof damping cube	 \\
			& Tendon length spring (M, 0.75) & Geom gap (M, 0.03) & Dof damping robot \\
			& Tendon stiffness (M, 0.75) & & Dof friction loss cube \\
			& & & Dof friction loss robot \\
			& & & Geom gap cube (AU, 0.01) \\
			& & & Geom gap robot (AU, 0.01) \\
			& & & Geom pos cube (AG, 0.002) \\
			& & & Geom pos robot (AG, 0.002) \\
			& & & Geom margin cube (AG, 0.0005) \\
			& & & Geom margin robot (AG, 0.0005) \\
			& & & Geom solimp (M, 1.0) \\
			& & & Geom solref (M, 1.0) \\
			& &  & Joint stiffness robot (UAG, 0.005) \\
	\midrule
	\multirow{3}{2cm}{Simulator physics (custom)}
			& Body mass &  & Friction robot \\
			& Cube size &  & Friction cube	\\
			& Tendon range &  & \\
	\midrule
	\multirow{5}{2cm}{Custom physics}
			& Action latency &  & Action noise \\
			& Backlash &  & Time step variance \\
			& Joint margin &  & \\
			& Joint range &  & \\
			& Time step & & \\
	\midrule
	Adversary & Adversary & \\
	\midrule
	Observation & & & Observation \\
        \bottomrule
    \end{tabular}
    \label{table:manipulation_randomizations}
\end{table}

A simple randomization for environment parameter $\lambda_i$, such as gravity, is parameterized by two ADR parameters corresponding to the boundaries of its randomization range: $\phi_{L, i}$ and $\phi_{H, i}$. Most randomizations that we used were simple. A more complex randomization such as observation noise is parameterized by several ADR parameters.

In the following, we denote default (non-randomized) values by using the subscript $(\cdot)_0$. We use $\lambda_i, \lambda_j, \dots$ to denote randomized environment parameters, which are parameterized in ADR by boundary values $\phi{L, i}, \phi{H, i}$, $\phi{L, j}, \phi{H, j}$, etc. Lastly, define $g(x) \triangleq \exp(x - 1.0)$.

\subsection{Simulator Physics Randomizations}\label{sec:physics_randomizations}

Simulator physics parameters are randomized by using either \emph{generic} or \emph{custom} randomizers.

\subsubsection{Generic randomizers}
A generic randomizer is specified by a \emph{noise mode} and a scaling factor $\alpha$. The noise modes used in this work are listed in Table~\ref{table:noise_modes}. The variable $x$ denotes a simulator physics parameter being randomized.

\begin{table}[h]
	\caption{Generic randomizer noise modes and their abbreviations. $x$ denotes a simulator parameter being randomized and $x_0$ its default value.}
	\centering
	\renewcommand{\arraystretch}{1.3}
    	\begin{tabular}{@{}lll@{}}
        \toprule
        \textbf{Noise Mode} & \textbf{Abbreviation} & \textbf{Expression} \\
    \midrule
	Additive Gaussian & AG & $x_0 + |N| \textrm{ for } N \sim \mathcal{N}(g(\alpha\lambda_i), g(|\alpha\lambda_j|)^2)$ \\
	Unbiased Additive Gaussian & UAG & $x_0 + N \textrm{ for } N \sim \mathcal{N}(0, g(|\alpha\lambda_i|)^2)$ \\
	Multiplicative & M & $x_0 e^{N} \textrm{ for } N \sim \mathcal{N}(\alpha\lambda_i, |\alpha\lambda_j|^2)$ \\
	\bottomrule
	\end{tabular}
	\label{table:noise_modes}
\end{table}

\subsubsection{Custom randomizers}
We also used several custom randomizers that perform slightly different randomization methods than generic randomizers. Again, $x$ denotes a simulator physics parameter being randomized.

\paragraph{Cube and robot friction.} The slide, spin, or roll friction of a cube or robot body is randomized by
\[
x = x_0 e^{w\lambda_i},
\]
where $w$ is a fixed weight specific for the type of friction. The weights for robot friction are 1.0 for all types. The weights for cube friction is $w=1.0$ for slide and $w=2.0$ for spin and roll.

\paragraph{Cube size.} Cube size is randomized by
\[
x = x_0 e^{0.15\lambda_i}
\]

\paragraph{Joint and tendon limits.} Both the lower and upper limits on joint and tendon ranges are randomized by 
\[
x = x_0 + n \textrm{ for } n \sim \mathcal{N}(0, 0.1g(|\lambda_i|)).
\]

\subsection{Custom Physics Randomizations}\label{sec:custom_physics}
Custom physics models were used to capture physical effects that are not modelled by the simulator. These models range from simple time delays to motor backlash.

\paragraph{Action delay.} Action delay $d$ in milliseconds is modelled and randomized by
\[
d = |\lambda_i| n_0 n_1 \textrm{ for } N_0 \sim \mathcal{N}(1, |\lambda_j|^2), N_1 \sim \mathcal{N}(1, |\lambda_k|^2),
\]
where $n_0$ is sampled once per episode, $n_1$ is sampled once per step.

\paragraph{Action latency.} Action latency $l$ in time steps is randomized by
\[
l \sim C[\lambda_i],
\]
where $C[n]$ is the uniform categorical distribution over $n$ elements.

\paragraph{Action noise.} Action noise $a$ is randomized by 
\[
a = a_0n_0 + n_1 + n_2 \textrm{ for } n_0 \sim \mathcal{N}(1, g(|\lambda_i|)^2), n_1 \sim \mathcal{N}(0, g(|\lambda_j|)^2), n_2 \sim \mathcal{N}(0, g(|\lambda_k|)^2),
\]
where $n_0,\, n_1$ are sampled once per episode, $n_2$ is sampled once per step.

\paragraph{Motor Backlash.} As described in ~\citep[C.2]{openai2018learning} our motor backlash implementation contains two parameters $\delta_{\pm 1}$. They are randomized by 
\[
\delta_{\pm1} = e^{n\delta_{\pm1, 0}} \textrm{ for } n \sim \mathcal{N}(1, \lambda_i^2).
\]

\paragraph{Gravity.} Gravity vector $g$ is randomized by
\[
g = g_0 + ug(\lambda_i)
\]
for $u \in \R^3$ a random vector uniformly distributed over the unit sphere.

\paragraph{Joint margin.} Joint margin $m$ is randomized by
\[
m = m_0n \textrm{ for } n \sim U[0, 0.15g(\lambda_i)].
\]

\paragraph{Time step.} Simulation time step $t$ is randomized by
\[
t = e^{0.6 \lambda_i}(t_0 + ne^{\lambda_j}) \textrm{ for } n \sim \textrm{Exp}(1/\kappa), \kappa \sim U[1250, 10000].
\]
The \emph{time step variance} randomization is equivalent to the time step randomization with $\lambda_i=0$.

\subsection{Random Network Adversary (RNA)}\label{sec:random_network_adversary}
ADR allows us to automatically choose randomization ranges and allows us to train on an ever-growing distribution over environments. However, one problem that ADR faces is that there might be unmodeled effects in the target domain. Since they are not modeled, ADR cannot randomize them.

To resolve this, we use an adversarial approach similar to~\cite{pinto2017supervision, pinto2017robust}. However, instead of training the adversary in a zero-sum fashion, we use networks with randomly sampled weights. We re-sample these weights at the beginning of every episode and keep them fixed throughout the episode. Our adversarial approach has two different means of perturbing the policy:
\begin{itemize}
    \item \emph{Action perturbations}. Let $\pi_{\text{adv}}$ denote an adversarial policy with the same action space as the policy controlling the robot (usually denoted just as $\pi$ but denoted as $\pi_{\text{robot}}$ in this section for clarity). We then use the following simple convex combination to obtain the action we execute in simulation: $\vec{a}_t = (1-\alpha) \vec{a}_{\text{robot}} + \alpha  \vec{a}_{\text{adv}}$, where $\alpha \in [0,1]$, $\vec{a}_{\text{robot}} \sim \pi_{\text{robot}}(\vec{o}_t)$ and $\vec{a}_{\text{adv}} \sim \pi_{\text{adv}}(\vec{o}_t)$. $\alpha$ controls the influence of the adversary, with $\alpha=0$ meaning no adversarial perturbation and $\alpha=1$ meaning only adversarial perturbations. Given a sufficiently powerful adversarial policy, action perturbations allow us to model arbitrary effects in the actuation system of the robot.
    \item \emph{Force/torque perturbations}. Similar to action perturbations, we apply force/torque perturbations to all bodies of the object being manipulated.\footnote{For the block reorientation task this is just a single body but for the Rubik's cube we apply perturbations to each cublet.}. More concretely, we generate a $6D$ vector for each body where the first $3$ dimensions define a force vector and the last $3$ dimensions define a torque vector for the same body. We scale the force vector by the body's mass and the torques by the body's corresponding inertia (we use a diagonal inertia matrix and scale each element of the torque vector by the corresponding diagonal entry) to normalize the effect each force/torque has. We then scale all force/torque perturbations by a scalar $\beta \in \R_{\geq0}$, thus allowing us to control how much influence force/torque perturbations have.
\end{itemize}
Given this formulation, we can use ADR to automatically adjust $\alpha$ and $\beta$ over time without the need for hand-tuning (which would be exceedingly difficult). Furthermore, we can sample $\alpha$ and $\beta$ iid per coordinate (i.e. per joint and per body, respectively), this having simulations in which some joints or bodies are ``more adversarial'' than others.

However, one question still remains: How do we obtain the perturbations, i.e. how do we implement the adversarial policy? We experimented with sampling random perturbations, using a zero-sum self-play setup where we train an adversarial policy directly using either feed-forward or recurrent networks. Surprisingly we found that a very simple method outperforms them in terms of sim2sim transfer: Random networks.

More concretely, we use a simple feed-forward network with $3$~hidden layers of $265$~units each. After each hidden layer we use a ReLU non-linearity. We use a discretized action space with $31$ bins per dimensions. We do this to ensure that randomly initializing the network leads uniform probability of selecting one of the $31$ bins (if we would use a continuous space and a $\tanh$ non-linearity, actions would always be close to $0$). We re-initialize the adversarial network at the beginning of each episode and keep it fixed after. The inputs of the adversarial network are the noisy fingertip positions as well as cube position, orientation, and face angle configuration. This idea is also related to random network distillation, where a randomly initialized CNN is used to help with exploration~\citep{burda2018exploration}.

We found that this approach works very well, outperforming more sophisticated methods like adversarial training. We believe this is because diversity is what ultimately aids in transfer. Even though a trained adversary is a stronger opponent, random networks provide the maximum amount of diversity.

\subsection{Observation Randomizations}\label{sec:observation_randomizations}
We add both correlated and uncorrelated noise to observations. Correlated noise is sampled once per episode and uncorrelated noise is sampled once per step. Given default noise standard deviations $a_0, b_0, c_0$, fixed for each observation element $o$, we randomize each observation by 
\[
o = o_0 n_0 + n_1 + n_2 \textrm{ for } n_0 \sim \mathcal{N}(1, (a_0e^{\lambda_i})^2), n_1 \sim \mathcal{N}(0, (b_0 e^{\lambda_i})^2), n_2 \sim \mathcal{N}(0, (c_0e^{\lambda_j})^2),
\]
where $n_0, n_1$ are sampled once per episode, $n_2$ is sampled once per step.

\subsection{Visual Randomizations}
\label{app:visual_randomizations}

There are two categories of randomizations for vision training, (a) variations in the scenes that are rendered by ORRB~\citep{chociej2019orrb} and (b) TensorFlow distortion operations applied on the images. ADR controls to what extent each category of randomizations affect the appearance of final images that are fed into the model with different ADR parameters. See the full list in \autoref{table:vision_randomizations}.

\begin{table}[h]
    \caption{Parameter randomizations used in vision model ADR training.}
    \centering
    \renewcommand{\arraystretch}{1.3}
    \begin{tabular}{@{}l|l|l@{}}
        \toprule
        \textbf{Category} & \textbf{Randomizer} & \textbf{Parameter} \\
        \midrule
        \multirow{26}{*}{ORRB~\cite{chociej2019orrb}}
        & \multirow{3}{*}{Camera Randomizer} & Position perturbation distance \\
        & & Rotation perturbation magnitude \\
        & & Field of view (fov) perturbation \\
        \cmidrule{2-3}
        & \multirow{3}{*}{Light Randomizer} & Individual light intensity \\
        & & Total intensity of lights in the scene \\
        & & Spotlight angle \\
        \cmidrule{2-3}
        & \multirow{3}{*}{Lighting Rig} & Number of lights \\
        & & Light distance \\
        & & Light height \\
        \cmidrule{2-3}
        & \multirow{7}{*}{Material Fixed Hue Randomizer} & Hue perturbation radius \\
        & & Saturation perturbation radius \\
        & & Value perturbation radius \\
        & & Range of saturation \\
        & & Range of value \\
        & & Range of glossiness \\
        & & Range of metallic \\
        \cmidrule{2-3}
        & \multirow{11}{*}{Post Processing Randomizer} & Hue shift of all colors \\
        & & Saturation of all colors \\
        & & Contrast of all colors \\
        & & Brightness of all colors \\
        & & Color temperature to set the white balance to \\
        & & Range of tint \\
        & & Strength of bloom filter\footnote{\url{https://docs.unity3d.com/Packages/com.unity.postprocessing@2.1/manual/Bloom.html}} \\
        & & Extent of veiling effects \\
        & & Degree of darkness added by ambient occlusion\footnote{\url{https://docs.unity3d.com/Packages/com.unity.postprocessing@2.1/manual/Ambient-Occlusion.html}} \\
        & & Grain strength \\
        & & Grain article size \\
        \midrule
        \multirow{6}{*}{TensorFlow} 
            & \multirow{6}{*}{---} & \texttt{tf.image.adjust\textunderscore hue} \\
            & & \texttt{tf.image.adjust{\textunderscore}saturation} \\
            & & \texttt{tf.image.adjust{\textunderscore}brightness} \\
            & & \texttt{tf.image.adjust{\textunderscore}contrast} \\
            & & Add random Gaussian noise \\
            & & Add random Gaussian noise that is same across channels \\
        \bottomrule
    \end{tabular}
    \label{table:vision_randomizations}
\end{table}

\FloatBarrier

\FloatBarrier
\section{Implementation Details}
\label{app:hyper}

\subsection{Goal Generation}
\label{app:goal-generation}

Algorithm listings for goal generations for both tasks are given here.
It is important to note that in both cases random orientation is sampled in such way that one face of the block or cube is pointing directly upwards. For reference, as was described earlier, the cube or block orientations are considered aligned if any face is pointing upwards within a tolerance of 0.4 radians. Cube faces are considered aligned if all six are within 0.1 radians from a straight angle configuration.

\begin{algorithm}
\caption{Block reorientation goal generation}
\label{algorithm:block_goal_generation}
\begin{algorithmic}
\State $goal\_orientation \gets$ \Call{RandomUniformUpwardOrientation}{} \Comment{Sample random quaternion}
\State \Call{SetOrientationGoal}{$goal\_orientation$} 
\end{algorithmic}
\end{algorithm}

\begin{algorithm}
\caption{Rubik's cube goal generation}
\label{algorithm:rubik_goal_generation}
\begin{algorithmic}
\Require $cube\_quat$ \Comment{Quaternion describing current cube orientation}
\Require $face\_angles$ \Comment{Current rotation angles of six faces of the cube}

\State $is\_aligned \gets$ \Call{IsOrientationAligned}{$cube\_quat$} \textbf{and} \Call{IsFaceAligned}{$face\_angles$}
\Statex \Comment{Check if cube is in the \emph{aligned} state}

\State $u \gets U(0, 1)$ \Comment{Sample random number uniformly between 0 and 1}
\Statex

\If{$is\_aligned$ \textbf{and} $(u < 0.5)$} \Comment{If cube is aligned then with 50\% probability we perform face rotation}
    \State $side \gets $ \Call{Random}{\{\texttt{CW}, \texttt{CCW}\}} \Comment{Choose clockwise or counterclockwise}
    \State $goal\_face\_angles \gets$ \Call{RotateTopFace}{$side$, $face\_angles$} \Comment{Rotate top face by 90 degrees}
    \State \Call{SetFaceGoal}{$goal\_face\_angles$}
    \State \Call{SetOrientationGoal}{$NULL$}
\Else \Comment{Otherwise, perform cube flip}
    \State $aligned\_face\_angles \gets$ \Call{Align}{$face\_angles$} \Comment{Align face angles to closest straight angle configuration}
    \State $goal\_orientation \gets$ \Call{RandomUniformUpwardOrientation}{} \Comment{Sample random quaternion}
    \State \Call{SetFaceGoal}{$aligned\_face\_angles$}
    \State \Call{SetOrientationGoal}{$goal\_orientation$}
\EndIf
\end{algorithmic}
\end{algorithm}

\FloatBarrier

\subsection{Neural Network Inputs}
\label{app:hyper-neural-net}

\autoref{table:policy-inputs-locked} lists the policy and value function inputs for the block reorientation task, respectively. The number of parameters for the Rubik's cube task network is given in \autoref{table:network-size}.

\begin{table}[h!]
    \footnotesize
    \centering
    \caption{Observations for the block reorientation task of the policy and value networks, respectively.}
    \renewcommand{\arraystretch}{1.3}
    \begin{tabular}{@{}llcc@{}}
        \toprule
        \textbf{Input} & \textbf{Dimensionality} & \textbf{Policy network} & \textbf{Value network} \\
        \midrule
        Fingertip positions & 15D & $\times$ & \checkmark \\
        Noisy fingertip positions & 15D & \checkmark & \checkmark \\
        Cube position & 3D & $\times$ & \checkmark \\
        Noisy block position & 3D & \checkmark & \checkmark \\
        Block orientation & 4D (quaternion) & $\times$ & \checkmark \\
        Noisy block orientation & 4D (quaternion) & \checkmark & \checkmark \\
        Goal orientation & 4D (quaternion) & \checkmark & \checkmark \\
        Relative goal orientation & 4D (quaternion) & $\times$ & \checkmark \\
        Noisy relative goal orientation & 4D (quaternion) & \checkmark & \checkmark \\
        Hand joint angles & 48D\footnote{\label{footnote:angles}Angles are encoded as $\sin$ and $\cos$, i.e. this doubles the dimensionality of the underlying angle.} & $\times$ & \checkmark \\
        All simulation positions \& orientations (\texttt{qpos}) & 38D & $\times$ & \checkmark \\
        All simulation velocities (\texttt{qvel}) & 36D & $\times$ & \checkmark \\
        \bottomrule
    \end{tabular}
\label{table:policy-inputs-locked}
\end{table}

\begin{table}[h!]
    \footnotesize
    \centering
    \caption{Network size measured by the number of parameters for the Rubik's cube network (parameters for the block reorientation task are the same except for the input embeddings).}
    \renewcommand{\arraystretch}{1.3}
    \begin{tabular}{lr}
        \toprule
        \textbf{Type} & \textbf{Number of parameters} \\ 
        \midrule
        Nontrainable parameters & 1539  \\
        Policy network & 13,863,132 \\ 
        Value function network & 13,638,657 \\
        Input embeddings & 267,776 \\
        \bottomrule
    \end{tabular}

    \label{table:network-size}
\end{table}

\FloatBarrier

\subsection{Hyperparameters}
\label{app:hyperparameters}

Hyperparameters used in our PPO policy are listed in \autoref{tbl:ppo}.

\autoref{tbl:adr-hparams} shows hyperparameters used in ADR for policy training.

Vision training hyperparameters are given in \autoref{tbl:vision-hyp} and the details of the model architecture are given in \autoref{tbl:vision-hyper-arch}. The target errors for our loss weight auto-balancer and vision ADR are listed in \autoref{tbl:vision-auto-balancer}. The increase and decrease thresholds in vision ADR are 70\% and 30\% respectively.

We also apply data augmentation for vision training. More specifically, we leave the object pose as is with $20\%$ probability, rotate the object by $90^{\circ}$ around its main axes with $40\%$ probability, and ``jitter'' the object by adding Gaussian noise to both the position and rotation independently with $40\%$ probability.

\begin{table}[h!]
    \footnotesize
    \centering
    \caption{Hyperparameters used for PPO.}
    \renewcommand{\arraystretch}{1.3}
    \begin{tabular}{@{}ll@{}}
        \toprule
        \textbf{Hyperparameter} & \textbf{Value} \\ 
        \midrule
        hardware configuration - block & 32 NVIDIA V100 GPUs + 12'800 CPU cores \\ 
        hardware configuration - Rubik's Cube & 64 NVIDIA V100 GPUs + 29'440 CPU cores \\ 
        action distribution & categorical with $11$ bins for each one of 20 action coordinates \\
        discount factor $\gamma$ & $0.998$ \\
        Generalized Advantage Estimation $\lambda$ & $0.95$ \\
        entropy regularization coefficient & varying $0.01$ -- $0.0025$ \\
        PPO clipping parameter $\epsilon$ & $0.2$ \\
        optimizer & Adam~\citep{Kingma2014AdamAM} \\
        learning rate & varying $3 \times 10^{-4}$ -- $1 \times 10^{-4}$  \\
        batch size (per GPU) & $5120$ chunks x $10$ transitions = $51'200$ frames \\
        Sample reuse (experience replay) & 3 \\
        Value loss weight & $1.0$ \\
        L2 regularization weight & $10^{-6}$ \\
    \bottomrule\end{tabular}
    \label{tbl:ppo}
\end{table}

\begin{table}[h!]
    \footnotesize
    \centering
    \caption{Hyperparameters used for ADR.}
    \renewcommand{\arraystretch}{1.3}
    \begin{tabular}{@{}ll@{}}
        \toprule
        \textbf{Hyperparameter} & \textbf{Value} \\ \midrule
        Maximum value of a $\phi$ & $4.0$ \\
        Boundary sampling probability & $0.5$ \\
        ADR step size $\Delta$ & $0.02$ \\
        ADR increase threshold $t_H$ & $20$ \\
        ADR decrease threshold $t_L$ & $10$ \\
        Performance queue length $m$ & $240$ \\
    \bottomrule\end{tabular}
    \label{tbl:adr-hparams}
\end{table}

\begin{table}[h!]
    \centering
    \footnotesize
    \caption{Hyperparameters used for the vision model training.}
    \renewcommand{\arraystretch}{1.3}
    \begin{tabular}{@{}ll@{}}
        \toprule
        \textbf{Hyperparameter} & \textbf{Value} \\ \midrule
        optimization hardware & 8 NVIDIA V100 GPUs + 64 vCPU cores \\
        rendering hardware & $16 \times 1$ NVIDIA V100 GPU + 8 vCPU core machines \\
        optimizer & LARS~\citep{LARS} \\
        learning rate & $0.5$, halved every $40\,000$ batches \\
        minibatch size &  $256 \times 3 = 768$ RGB images \\
        image size & $200 \times 200$ pixels \\
        weight decay regularization & $0.0001$ \\
        number of training batches & $300\,000$ \\
        network architecture & shown in \autoref{fig:vision-arch} \\
    \bottomrule\end{tabular}
    \label{tbl:vision-hyp}
\end{table}

\begin{table}[h!]
    \centering
    \footnotesize
    \caption{Hyperparameters for the vision model architecture.}
    \renewcommand{\arraystretch}{1.3}
    \begin{tabular}{@{}ll@{}}
        \toprule
        \textbf{Layer} & \textbf{Details} \\ \midrule
        Input RGB Image & $200\times200\times3$ \\
        Batch Norm & \\
        Conv2D & 64 filters, $5\times5$, stride 2, no padding \\
        Dropout & keep probability = 0.9 \\
        Max Pooling & $2\times2$, stride 1, no padding \\
        Batch Norm & \\
        ResNet~\citep{He2016DeepRL} Bottleneck & 3 blocks, 64 filters, $1\times1$, stride 1 \\
        Dropout & keep probability = 0.9 \\
        ResNet~\citep{He2016DeepRL} Bottleneck & 4 blocks, 128 filters, $2\times2$, stride 2 \\
        ResNet~\citep{He2016DeepRL} Bottleneck & 6 blocks, 256 filters, $2\times2$, stride 2 \\
        ResNet~\citep{He2016DeepRL} Bottleneck & 3 blocks, 512 filters, $2\times2$, stride 2 \\
        Max Pooling & $2\times2$, stride 1, no padding \\
        Flatten & \\
        Concatenate & all 3 image towers combined\\ \midrule
        Dropout & keep probability = 0.9 \\
        Fully Connected & 512 units \\
        Batch Norm & \\
        Fully Connected & 256 units \\
        Batch Norm & \\
        Fully Connected & output, $3$ pos. + $4$ quat. + $180$ top face angle + $90$ active angles + $3$ active axis = 280 units \\
    \bottomrule\end{tabular}
    \label{tbl:vision-hyper-arch}
\end{table}

\begin{table}[h!]
    \footnotesize
    \centering
    \caption{Target error thresholds for different prediction labels in the loss weight auto-balancer and ADR of the vision model. The active axis target error is set to 0.5 because the error is binary for each individual sample.}
    \renewcommand{\arraystretch}{1.3}
    \begin{tabular}{@{}lrr@{}}
        \toprule
        \multirow{2}{*}{\textbf{Label}} & \multicolumn{2}{c}{\textbf{Target Errors}} \\
        & Auto Balancer & Vision ADR \\
        \midrule
        Orientation & $3^\circ$ & $5^\circ$ \\
        Position & $3$ mm & $5$ mm \\
        Top face angle & $3^\circ$ & $5^\circ$ \\
        Active axis & $0.5$ & $0.5$ \\
        Active face angles & $3^\circ$ & $5^\circ$ \\
    \bottomrule\end{tabular}
    \label{tbl:vision-auto-balancer}
\end{table}

\FloatBarrier

\subsection{Vision Post-Processing}

Algorithm~\ref{algorithm:vision_post_processing} describes the post-processing logic that we apply when using the vision model for face angle predictions.
 
\begin{algorithm}
\caption{Vision model post-processing logic for tracking the full state of a Rubik's cube on the physical robot.}
\label{algorithm:vision_post_processing}
\begin{algorithmic}
\Require $base\_angles[0..5]$ \Comment{6 angles in total}
\Require $tracked\_angles[0..5]$
\Require $cube\_quat$ \Comment{Predicted cube orientation}
\Require $active\_axis$ \Comment{Predicted active axis, $\in (0, 1, 2)$}
\Require $active\_angles[0,1]$ \Comment{Predicted two active face angles in $[-\pi/4, \pi/4]$}
\Require $top\_angle$ \Comment{Predicted top face angle in $[-\pi, \pi]$}
\State $k \gets$ \Call{ExtractTopFaceId}{$cube\_quat$}
\State $is\_aligned \gets$ \Call{IsOrientationAligned}{$cube\_quat$}
\For{$i:=0 \text{ to } 5$}
    \If{$i = k \textbf{ and } is\_aligned$} \Comment{If this face is placed on top}
        \State $base\_angles[i] \gets top\_angle$
        \State $tracked\_angles[i] \gets top\_angle$
    \Else
        \State {$a, b \gets \text{divmod}(i, 2)$}
        \If{$a = active\_axis$} \Comment{If this is a active face}
            \State $tracked\_angles[i] \gets$ \Call{MoveAngle}{$base\_angles[i]$, $active\_angles[b]$} \Comment{See Algorithm~\ref{algorithm:vision_move_angle}}
        \EndIf
    \EndIf
\EndFor
\end{algorithmic}
\end{algorithm}

\begin{algorithm}
\caption{\protect\Call{MoveAngle}{}: Move the base face angle to the target by a modulo $\pi/2$ delta angle.}
\label{algorithm:vision_move_angle}
\begin{algorithmic}
\Require $a$ \Comment{Base face angle}
\Require $b$ \Comment{Target face angle}
\State $\delta \gets \text{mod}(b - a + \pi/4, \pi/2) - \pi/4$
\State \textbf{return} $a + \delta$
\end{algorithmic}
\end{algorithm}

\FloatBarrier

\section{Full Results}
\label{app:full-results}

\subsection{Simulation Calibration}
\label{app:full-results-sim-calib}

We test how much of an impact simulation calibration has. We use a simplifed version of the Rubik's cube: The face cube. In this version, the Rubik's cube is constrained to only have 2 degrees of freedom with the other degrees disabled. We evaluate a policy trained on the old simulation and on the new simulation (i.e. with coupling and dynamics calibration). We calibrate the following MuJoCo parameters that are related to joint movements: \texttt{dof\_damping}, \texttt{jnt\_range}, \texttt{geom\_size}, \texttt{tendon\_range}, \texttt{tendon\_lengthspring}, \texttt{tendon\_stiffness}, \texttt{actuator\_forcerange}, and \texttt{actuator\_gainprm}. Results are in \autoref{table:sim-calibration-face-cube}.

\begin{table}[h]
    \caption{Performance of two policies trained on two different environments on a simplified version of the Rubik's cube task. We run 10 trials for each policy on the physical setup and report the mean and median number of face rotations and mean number of policy steps per face rotation.}
    \label{table:sim-calibration-face-cube}
    \centering
    \renewcommand{\arraystretch}{1.3}
    \begin{tabular}{@{}lrrr@{}}
        \toprule
        \multirow{2}{*}{\textbf{Policy}} &  \multicolumn{2}{c}{\textbf{Successes}} & \textbf{Steps Per Success} \\
        & \textbf{Mean} & \textbf{Median} & \textbf{Mean} \\
        \midrule
        Original Simulation & $4.8 \pm 2.89$ & $5.00$ & $321.10$ \\
        New Simulation (Coupling Model + Calibration) & $\mathbf{14.30 \pm 6.56}$ & $\mathbf{15.50}$ &  $\mathbf{252.50}$ \\
        \bottomrule
    \end{tabular}
\end{table}

\subsection{Physical Trials}
\label{app:full-results-phy-trials}

The full statistics of our physical trials are available in \autoref{table:adr-xyz-transfer-full} and \autoref{table:adr-full-transfer-full}.

\begin{table}[h]
    \caption{Performance of different policies on the block reorientation task. We evaluate each policy on the real robot for 10 trials. For each individual trial, we report the mean $\pm$ standard error and median number of successes, the time per success in second, as well as the full list of success counts.}
    \label{table:adr-xyz-transfer-full}
    \centering
    \renewcommand{\arraystretch}{1.3}
    \begin{tabular}{l|rrr|l}
        \toprule
        \textbf{Policy}
        & \textbf{Mean}
        & \textbf{Median}
        & \textbf{Time per Success} 
        & \multicolumn{1}{c}{\textbf{Trials (sorted)}} \\

        \midrule
        
        Baseline (from \cite{openai2018learning})
        & $18.8 \pm 5.4$ & $13.0$ & --- 
        & $50,41,29,27,14,12,6,4,4,1$ \\
        
        Baseline (re-run of \cite{openai2018learning})
        & $4.0 \pm 1.7$ & $2.0$ & $17.24$ sec
        & $17,10,3,3,2,2,1,1,1,0$ \\

        \midrule

        Manual DR 
        & $2.7 \pm 1.1$ & $1.0$ & $31.65$ sec
        & $11,5,4,3,1,1,1,1,0,0$ \\
        ADR (Small)
        & $1.4 \pm 0.9$ & $0.5$ & $67.49$ sec
        & $9,2,1,1,1,0,0,0,0,0$ \\
        ADR (Medium)
        & $3.2 \pm 1.2$ & $2.0$ & $29.50$ sec
        & $12,7,4,3,2,2,1,1,0,0$ \\
        ADR (Large)
        & $13.3 \pm 3.6$ & $11.5$ & $4.45$ sec
        & $38,25,17,16,12,11,5,4,3,2$ \\
        
        \midrule
        
        ADR (XL)
        & $16.0 \pm 4.0$ & $12.5$ & $\mathbf{5.10}$ sec
        & $42,28,27,16,13,12,8,7,5,2$ \\
        ADR (XXL)
        & $\mathbf{32.0 \pm 6.4}$ & $\mathbf{42.0}$ & $5.18$ sec
        & $\mathbf{50},\mathbf{50},\mathbf{50},\mathbf{50},43,41,13,12,6,5$ \\

        \bottomrule
    \end{tabular}
\end{table}

\begin{table}[h]
    \caption{Performance of different policies on the Rubik's cube for a fixed fair scramble goal sequence. We evaluate each policy on the real robot for 10 trials. For each individual trial, we report the mean $\pm$ standard error and median number of successes, the time per success in second, as well as the full list of success counts.}
    \label{table:adr-full-transfer-full}
    \centering
    \renewcommand{\arraystretch}{1.3}
    \begin{tabular}{ll|rr|rr|l}
        \toprule
        \multirow{2}{*}{\textbf{Policy}}
        & \multirow{2}{*}{\textbf{Sensing}}
        & \multirow{2}{*}{\textbf{Mean}}
        & \multirow{2}{*}{\textbf{Median}}
        & \multicolumn{2}{|c}{\textbf{Time per Success}} 
        & \multicolumn{1}{|c}{\multirow{2}{*}{\textbf{Trials (sorted)}}} \\
        & & & & (Flip) & (Rotation) & \\

        \midrule
        Manual DR & Giiker + Vision 
        & $1.8 \pm 0.4$ & $2.0$  & $102.92$ sec & $35.60$ sec
        & $4,3,2,2,2,2,2,1,0,0$ \\
        ADR & Giiker + Vision 
        & $3.8 \pm 1.0$ & $3.0$ & $66.97$ sec & $30.65$ sec 
        & $8,8,8,5,4,2,2,1,0,0$ \\
        
        \midrule
        ADR (XL) & Giiker + Vision 
        & $17.8 \pm 4.2$ & $12.5$  & $\mathbf{6.44}$ sec & $\mathbf{10.98}$ sec
        & $44,38,24,17,14,11,9,8,7,6$ \\
        ADR (XXL) & Giiker + Vision 
        & $\mathbf{26.8 \pm 4.9}$ & $\mathbf{22.0}$ & $6.55$ sec & $11.79$ sec 
        & $\mathbf{50},\mathbf{50},42,24,22,22,21,19,13,5$ \\
        
        \midrule
        ADR (XXL) & Vision 
        & $12.8 \pm 3.4$ & $10.5$ & $9.71$ sec & $13.23$ sec 
        & $31,25,21,18,17,4,3,3,3,3$ \\
        \bottomrule
    \end{tabular}
\end{table}

\FloatBarrier

\subsection{Vision Model Performance}
\label{app:full-results-vision}

Our vision model for estimating the full state of a Rubik's cube outputs five labels. The prediction errors are evaluated on both simulated and real data. Orientation error is computed as rotational distance over a quaternion representation. Position error is the euclidean distance in 3D space, in millimeters. Face angle error is measured in degree ($^\circ$). For active axis, we measure the percentage of incorrect predicted label (equivalent to 1.0 - accuracy). Note that the high error on the active axis prediction is mostly due to the fact that the cube is often aligned with very tiny active face angles and hence the active axis is less pronounced.
\begin{itemize}
    \item \textbf{Orientation:} Cube orientation in quaternion.
    \item \textbf{Position:} Cube position in millimetre.
    \item \textbf{Top angle:} The full angle of the top face in $[-\pi, \pi]$.
    \item \textbf{Active axis:} Which one out of three axes is active (i.e., has non-aligned faces).
    \item \textbf{Active angles:} The angles of two faces relevant for the active axis in $[-\pi/4, \pi/4]$.
\end{itemize}

\begin{table}[h]
    \caption{Performance of vision models in all our experiments for the Rubik's cube prediction task, including experiment in the ablation study and models trained at different ADR entropy levels.}
    \label{table:vision-ablations-full}
    \centering
    \renewcommand{\arraystretch}{1.3}
    \begin{tabular}{ll|r|rrr|rrr}
        \toprule
        \multicolumn{2}{c|}{Errors} & Baseline & No DR & No focal loss & Non-discrete angles & ADR (S) & ADR (M) & ADR (L) \\
        \midrule
        \multirow{5}{*}{Sim} 
        & \textbf{Orientation} & $6.52^\circ$ & $3.95^\circ$ & $15.94^\circ$ & $9.02^\circ$ & $5.02^\circ$ & $15.68^\circ$ & $15.76^\circ$ \\
        &  \textbf{Position} & $2.63$ mm & $2.97$ mm & $5.02$ mm & $3.78$ mm & $3.36$ mm & $3.02$ mm & $3.58$ mm \\
        & \textbf{Top angle} & $11.95^\circ$ & $8.56^\circ$ & $10.17^\circ$ & $42.46^\circ$ & $9.34^\circ$ & $20.29^\circ$ & $20.78^\circ$ \\
        & \textbf{Active axis} & $7.24$ \% & $4.3$ \% & $5.6$ \% & $6.3$ \% & $8.10$ \% & $10.97$ \% & $10.67$ \%\\
        & \textbf{Active angles} & $3.40^\circ$ & $2.95^\circ$ & $2.86^\circ$ & $9.6^\circ$ & $3.63^\circ$ & $4.46^\circ$ & $4.31^\circ$ \\
        \midrule
        \multirow{5}{*}{Real} 
        & \textbf{Orientation} & $7.81^\circ$ & $128.83^\circ$ & $19.10^\circ$ & $10.40^\circ$ & $8.93^\circ$ & $8.44^\circ$ & $\mathbf{7.48}^\circ$ \\
        &  \textbf{Position} & $6.47$ mm & $69.40$ mm & $9.42$ mm & $7.97$ mm & $7.61$ mm & $7.30$ mm & $\mathbf{6.24}$ mm \\
        & \textbf{Top angle} & $15.92^\circ$ & $85.33^\circ$ & $17.54^\circ$ & $35.27^\circ$ & $16.57^\circ$ & $15.81^\circ$ & $\mathbf{13.82}^\circ$ \\
        & \textbf{Active angles} & $8.89^\circ$ & $33.04^\circ$ & $9.27^\circ$ & $17.68^\circ$ & $9.16^\circ$ & $8.97^\circ$ & $\mathbf{8.84}^\circ$ \\
        \bottomrule
    \end{tabular}
\end{table}

\subsection{Meta-Learning}
\label{app:full-results-meta}

\begin{figure}[h]
    \centering
    \begin{subfigure}[b]{\textwidth}
        \includegraphics[width=0.47\textwidth]{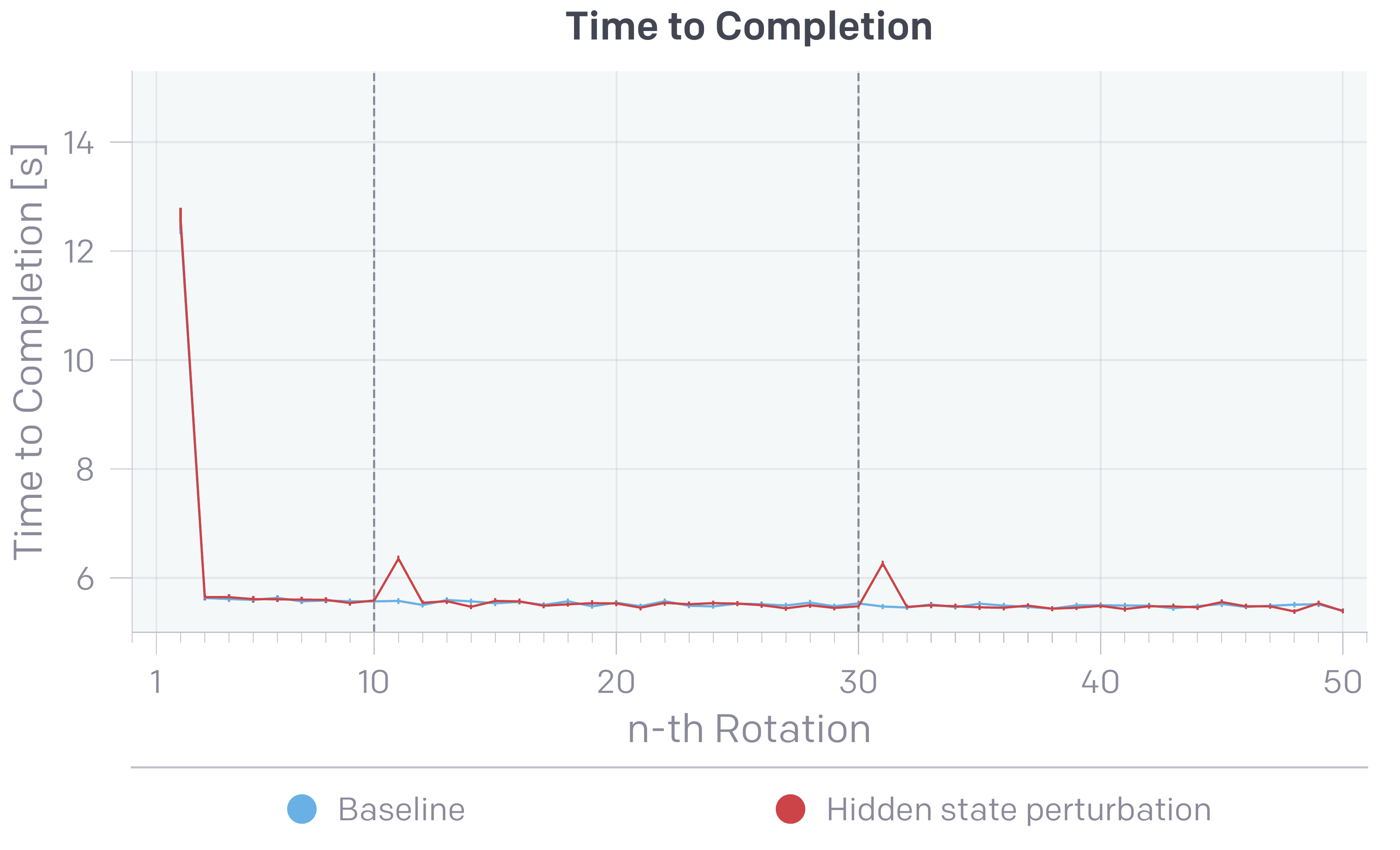}
        \hfill
        \includegraphics[width=0.47\textwidth]{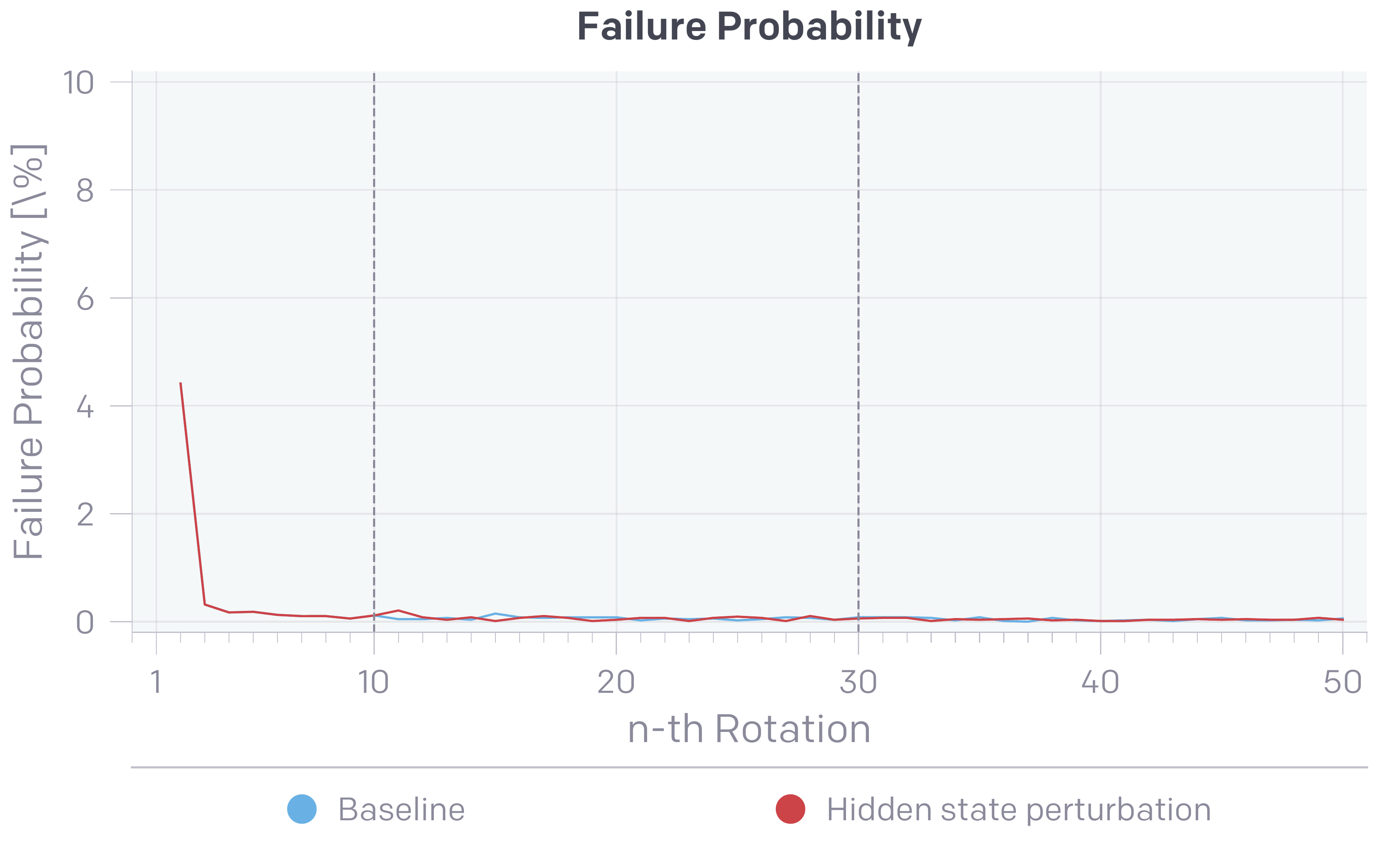}
        \caption{Resetting the hidden state.}
    \end{subfigure}
    \\
    \begin{subfigure}[b]{\textwidth}
        \includegraphics[width=0.47\textwidth]{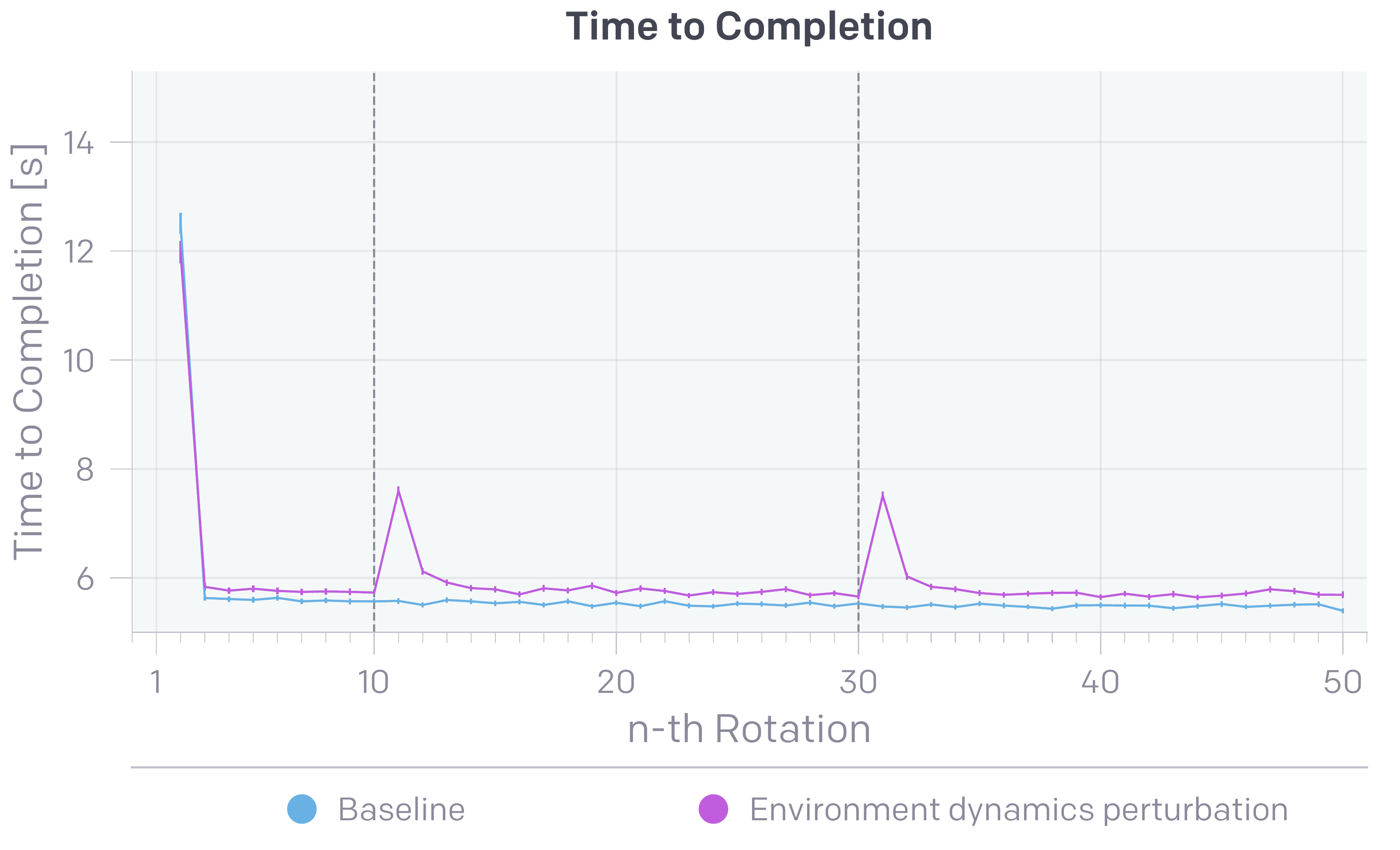}
        \hfill
        \includegraphics[width=0.47\textwidth]{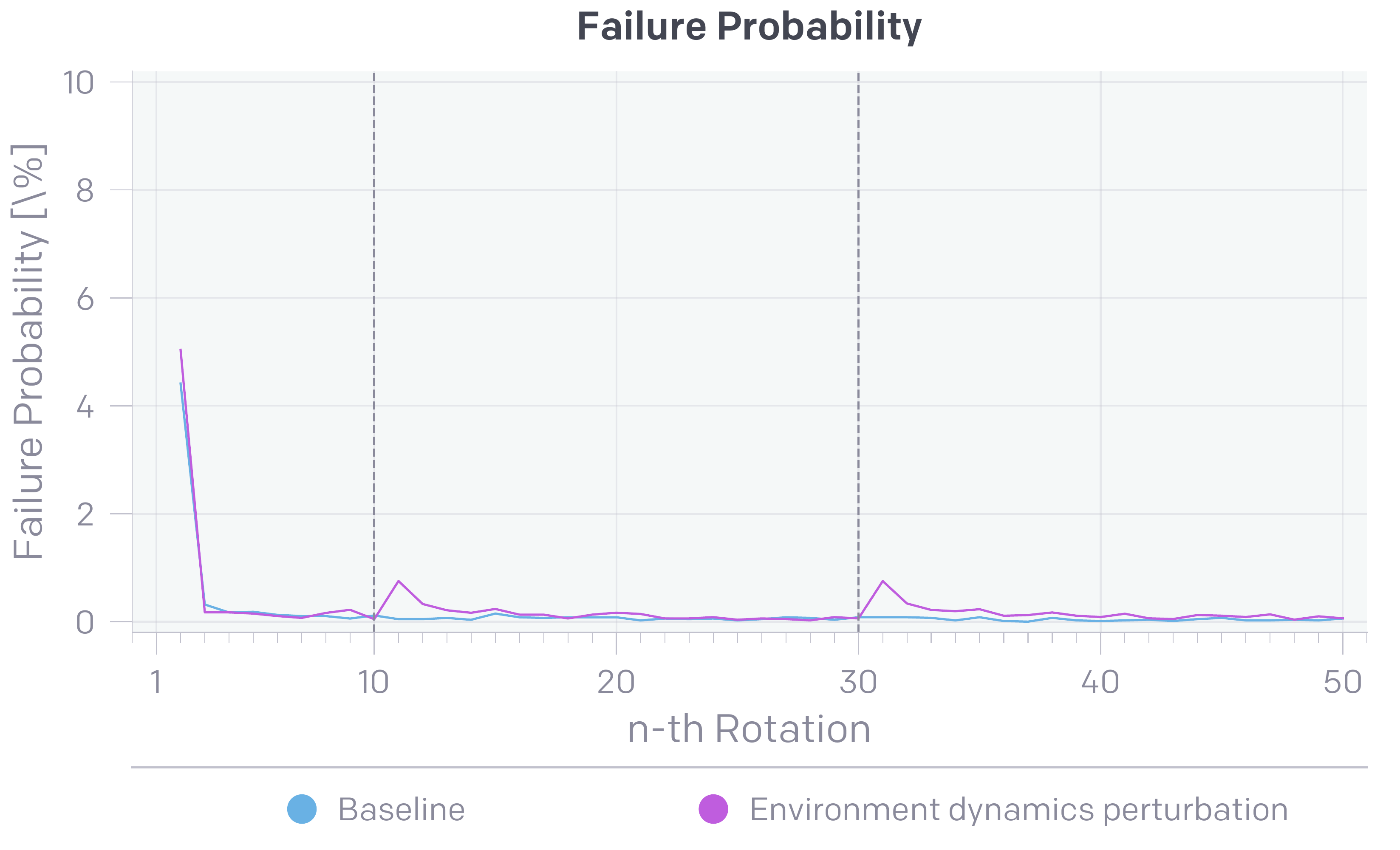}
        \caption{Re-sampling environment dynamics.}
    \end{subfigure}
    \\
    \begin{subfigure}[b]{\textwidth}
        \includegraphics[width=0.47\textwidth]{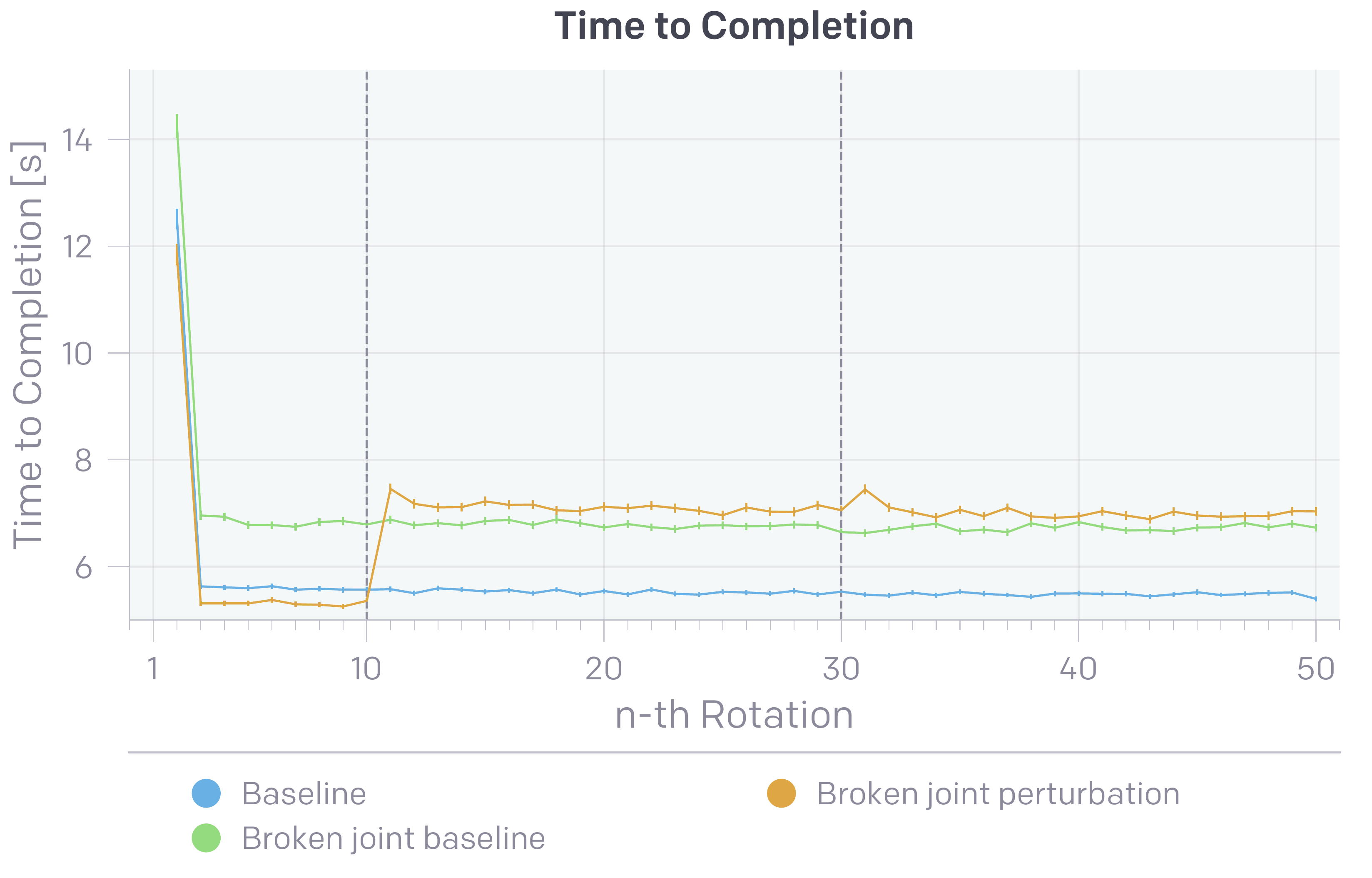}
        \hfill
        \includegraphics[width=0.47\textwidth]{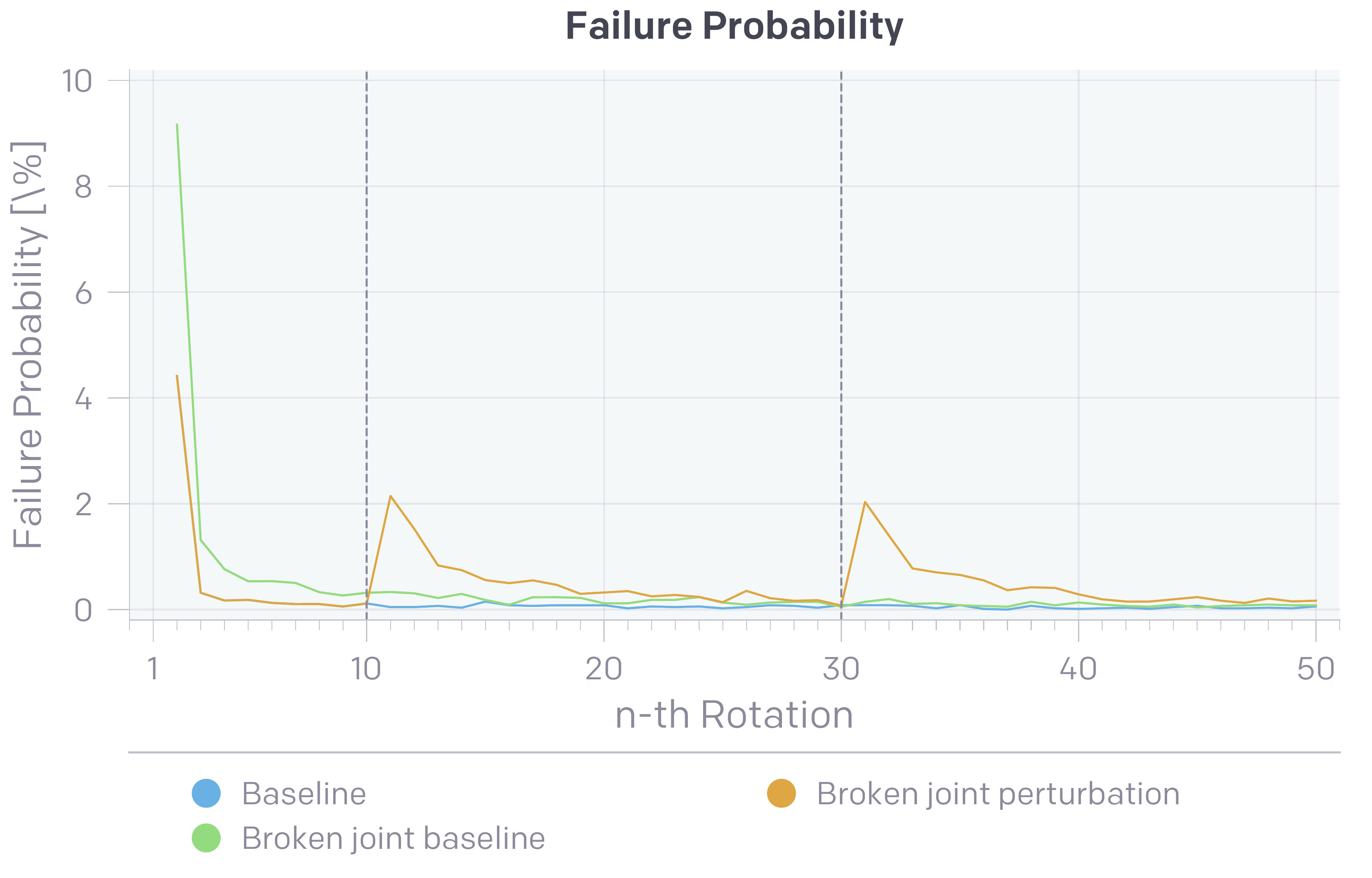}
        \caption{Breaking a random joint.}
    \end{subfigure}
    \caption{We run $10\,000$ simulated trials with only face rotations until $50$ rotations have been achieved. For each of the face rotations (i.e. the $1$\textsuperscript{st}, $2$\textsuperscript{nd}, $\ldots$, $50$\textsuperscript{th}), we measure the average time to completion (in seconds) and average failure probability over those $10$k trials. Error bars indicate the estimated standard error.  ``Baseline'' refers to a run without any perturbations applied. ``Broken joint baseline'' refers to trials where a joint was randomly disabled from the very beginning. We then compare against trials that start without any perturbations but are perturbed at the marked points after the $10$\textsuperscript{th} and $30$\textsuperscript{th} rotation by (a) resetting the policy hidden state, (b) re-sampling environment dynamics, or (c) breaking a random joint.}
    \label{fig:meta-exp-2}
\end{figure}

The meta-learning results for face rotations are available in \autoref{fig:meta-exp-2}. The results are comparable to the cube flips presented in the main body of the paper. We include them in the appendix for completeness.

\FloatBarrier

\section{Visualizations}
\label{app:visualizations}

\subsection{Policy Recurrent States}
\label{app:policy_visualizations}
We visualize the LSTM cell and hidden states of a Rubik's cube policy during rollouts. The techniques we use have been successfully applied to study the interpretability of deep neural networks used for hand-writing generation and vision~\cite{carter2016experiments, olah2018the}.

We collected data from a Rubik's cube policy rollout in simulation. The data spans 1500 time steps (2 minutes) and contains a sequence of successful cube re-orientations and face rotations. We collected the policy LSTM hidden and cell states at each time step, along with policy observations and actions. Each hidden or cell state vector contained 1024 units.

\begin{figure}[h]
    \centering
    \begin{subfigure}[b]{\textwidth}
        \centering
        \includegraphics[trim=24 24 24 24, clip, width=\textwidth]{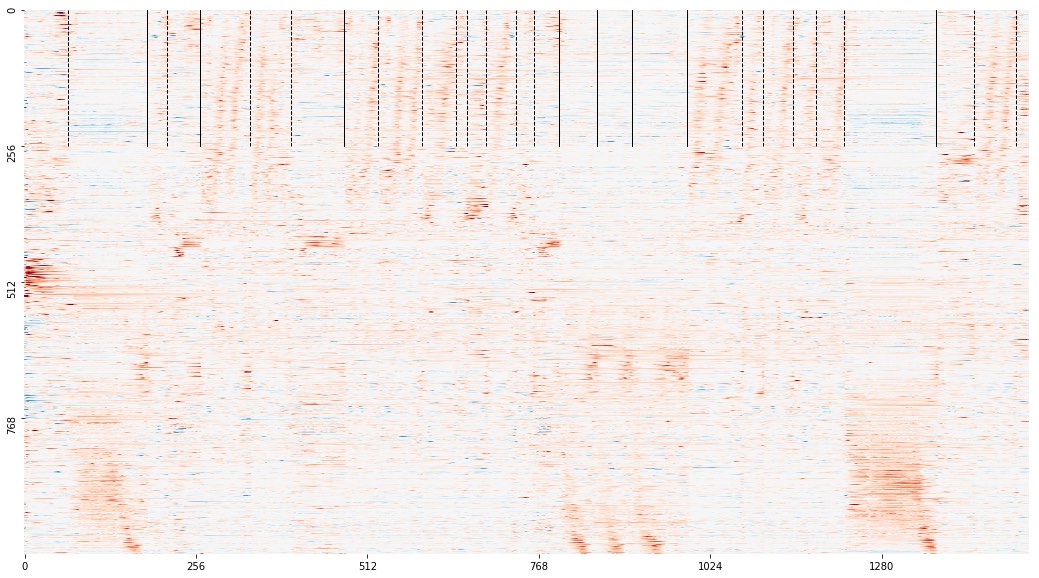}
        \caption{}
    \end{subfigure}
    \\
    \begin{subfigure}[b]{\textwidth}
        \centering
        \includegraphics[width=\textwidth]{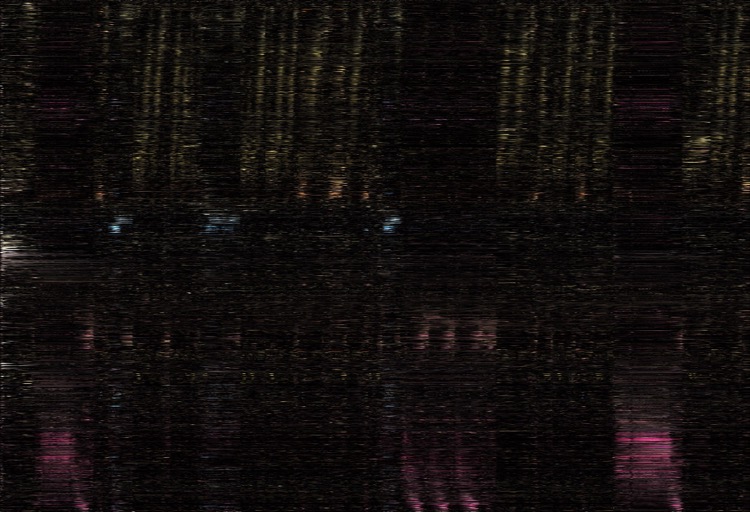}
        \caption{}
    \end{subfigure}
    \caption{Visualization of Rubik's cube policy LSTM cell states. Each row corresponds to the activation values of a cell state (1024 in total) over 1500 time steps. (a) after t-SNE reordering, dashed lines mark a successful block re-orientation, solid lines mark a successful face rotation. (b) after coloring by NMF factors.}
    \label{fig:policy_visualization}
\end{figure}

We arranged the cell (or hidden) state data into a matrix of 1024 rows by 1500 columns. We applied 1D t-SNE~\cite{vanDerMaaten2008} to the rows of this matrix in order to group cell state units (``neurons'') that have similar activation patterns across time. We also applied similar tricks as in~\cite{carter2016experiments} to further enhance the appearance of the grouped neurons. The resulting visualization is shown in~\autoref{fig:policy_visualization}~(a). 

We marked the time steps when the policy successfully completed a block re-orientation (dashed) and a face rotation (solid). We see a clear distinction in activation patterns of groups of neurons in the time before a successful re-orientation vs.~face rotation. We can also clearly see when the policy struggles with face rotations (both near the beginning and end of the rollout) where the same group of neurons are active over an extended time.

We performed Non-negative Matrix Factorization (NMF)~\cite{olah2018the} on the activations matrix to identify factors that can be used to group neurons. We used 6 factors and assigned a distinct color to each factor. We colored each neuron by first projecting the row vector onto each factor then linearly combining the factor colors using the projection weights. The results are shown in~\autoref{fig:policy_visualization}~(b).

With coloring by NMF factors, we observe 4 distinct groups of neuron activations: yellow, orange, blue, and pink. By matching the activation time of these neuron groups with actions, we found that each group of activations corresponds to a sequence of complex joint actuations, in other words, a skill. For example, the yellow activation group corresponds to flipping the cube back towards the palm of the hand\footnote{See \url{https://openai.com/blog/solving-rubiks-cube} for the skills corresponding to the other activation groups.}. It is quite surprising to see such high-level skills being clearly identifiable in the cell state representation.

\end{document}